\documentclass[journal,final,twocolumn,letterpaper,10pt,twoside]{IEEEtran}

\IEEEoverridecommandlockouts

\usepackage{amsmath}
\usepackage{subfig}
\usepackage{graphicx}
\usepackage{amsfonts}
\usepackage{wrapfig}
\usepackage{makeidx}
\usepackage{moreverb}
\usepackage[lined,boxed,commentsnumbered]{algorithm2e}
\usepackage{multirow}
\usepackage{array}
\usepackage{MnSymbol}
\usepackage{bbding}
\usepackage{amsthm}
\usepackage{setspace}
\usepackage{epstopdf}

\DeclareGraphicsExtensions{.jpg,.png,.tif,.eps}
\begin{document}

\title{Machine Learning Methods for Attack Detection in the Smart Grid}

\author{Mete Ozay\IEEEmembership{, Member,~IEEE,} I\~naki Esnaola\IEEEmembership{, Member,~IEEE,} Fatos T. Yarman Vural\IEEEmembership{, Senior Member,~IEEE,} \\ Sanjeev R. Kulkarni\IEEEmembership{, Fellow,~IEEE,} and H. Vincent Poor\IEEEmembership{, Fellow,~IEEE}

%
\thanks{M. Ozay is with the School of Computer Science,  University of Birmingham, B15 2TT, UK (e-mail: m.ozay@cs.bham.ac.uk). I. Esnaola, S. R. Kulkarni and H. V. Poor are with the Department of Electrical Engineering, Princeton University, Princeton, NJ 08544, USA (e-mail: \{jesnaola, kulkarni, poor\}@princeton.edu). I. Esnaola is also with the Department of Automatic Control and Systems Engineering, University of Sheffield, S1
3JD, UK. F. T. Yarman Vural is with the Department of Computer Engineering, Middle East Technical University, Ankara, Turkey (e-mail: vural@ceng.metu.edu.tr).}
\thanks{This research was supported in part by the U. S. National Science Foundation under Grant CMMI-1435778.}}


%


\maketitle

\begin{abstract}
Attack detection problems in the smart grid are posed as statistical learning problems for different attack scenarios in which the measurements are observed in batch or online settings. In this approach, machine learning algorithms are used to classify measurements as being either secure or attacked.  An attack detection framework is provided to exploit any available prior knowledge about the system and surmount constraints arising from the sparse structure of the problem in the proposed approach. Well-known batch and online learning algorithms (supervised and semi-supervised) are employed with decision and feature level fusion to model the attack detection problem. The relationships between statistical and geometric properties of attack vectors employed in the attack scenarios and learning algorithms are analyzed to detect \textit{unobservable attacks} using statistical learning methods. The proposed algorithms are examined on various IEEE test systems. Experimental analyses show that machine learning algorithms can detect attacks with performances higher than the attack detection algorithms which employ state vector estimation methods in the proposed attack detection framework.
\end{abstract}




%
\IEEEpeerreviewmaketitle

\section{Introduction}

Machine learning methods have been widely proposed in the smart grid literature for monitoring and control of power systems \cite{ml1,ml5,ml3,ml4}. Rudin et al. \cite{ml1} suggest an intelligent framework for system design in which machine learning algorithms are employed to predict the failures of system components. Anderson et al. \cite{ml5} employ machine learning algorithms for the energy management of loads and sources in smart grid networks. Malicious activity prediction and intrusion detection problems have been analyzed using machine learning techniques at the network layer of smart grid communication systems \cite{ml3,ml4}. 

In this paper, we focus on the false data injection attack detection problem in the smart grid at the physical layer. We use the Distributed Sparse Attacks model proposed by Ozay et al. \cite{mi_jsac}, where the attacks are directed by injecting false data into the local measurements observed by either local network operators or smart Phasor Measurement Units (PMUs) in a network with a hierarchical structure, i.e. the measurements are grouped into clusters. In addition, network operators who employ statistical learning algorithms for attack detection know the  topology of the network, measurements observed in the clusters and the measurement matrix \cite{mi_jsac}.

In attack detection methods that employ state vector estimation, first the state of the system is estimated from the observed measurements. Then, the residual between the observed and the estimated measurements is computed. If the residual is greater than a given threshold, a data injection attack is declared \cite{mi_jsac,book,s1,kosut}. However, exact recovery of state vectors is a challenge for state vector estimation based methods in sparse networks \cite{mi_jsac,cn,kim}, where the Jacobian measurement matrix is sparse. Sparse reconstruction methods can be employed to solve the problem, but the performance of this approach is limited by the sparsity of the state vectors \cite{mi_jsac,candes,donoho}.  In addition, if false data injected vectors reside in the column space of the Jacobian measurement matrix and satisfy some sparsity conditions (e.g., the number of nonzero elements is at most $\kappa^*$, which is bounded by the size of the Jacobian matrix), then false data injection attacks, called \textit{unobservable} attacks, cannot be detected \cite{s1, kosut}. 


The contributions of this paper are as follows:
\begin{enumerate}

\item  We conduct a detailed analysis of the techniques proposed by Ozay et al. \cite{mi1} who employ supervised learning algorithms to predict false data injection attacks. In addition, we discuss the validity of the fundamental assumptions of statistical learning theory in the smart grid. Then, we propose semi-supervised, online learning, decision and feature level fusion algorithms in a generic attack construction framework, which can be employed in hierarchical and topological networks for different attack scenarios.

\item We analyze the geometric structure of the measurement space defined by measurement vectors, and the effect of false data injection attacks on the distance function of the vectors. This leads to algorithms for \textit{learning} the distance functions, \textit{detecting} unobservable attacks, \textit{estimating} the attack strategies and \textit{predicting} future attacks using a set of observations.


\item We empirically show that the statistical learning algorithms are capable of detecting both observable and unobservable  attacks with performance better than the attack detection algorithms that employ state vector estimation methods. In addition, phase transitions can be observed in the performance of Support Vector Machines (SVM)  at a value of $\kappa^*$ \cite{phase}.

\end{enumerate}

In the next section, the attack detection problem is formulated as a statistical classification problem in a  network according to the model proposed by Ozay et al. \cite{mi_jsac}. In Section~\ref{sec:formulation}, we establish the relationship between  statistical learning methods and attack detection problems in the smart grid. Supervised, semi-supervised, decision and feature level fusion, and online learning algorithms are used to solve the classification problem in Section~\ref{sec:methods}. In Section~\ref{sec:exp}, our approach is numerically evaluated on IEEE test systems. A summary of the results and discussion on future work are given in Section~\ref{sec:conc}.

\section{Problem Formulation}
\label{sec:formulation}

In this section, the attack detection problem is formalized as a machine learning problem. 

\subsection{False Data Injection Attacks}

False Data Injection Attacks are defined in the following model:  
 \begin{equation}
\mathbf{z} = \mathbf{H} \mathbf{x} + \mathbf{n},
\end{equation}
where $\mathbf{x} \in \mathbb{R} ^ D$ contains the voltage phase angles at the buses, $\mathbf{z} \in \mathbb{R} ^ N$ is the vector of measurements, $\mathbf{H}\in\mathbb{R}^{N \times D}$ is the measurement Jacobian matrix and $\mathbf{n} \in \mathbb{R} ^ N$ is the measurement noise, which is assumed to have independent components \cite{s1}. The attack detection problem is defined as that of deciding whether or not there is an attack on the measurements. If the noise is distributed normally with zero mean, then a State Vector Estimation (SVE) method can be employed by computing	
\begin{equation}
\mathbf{\hat x} = ( \mathbf{H} ^T \mathbf{\Lambda} \mathbf{H}) ^{-1} \mathbf{H} ^T \mathbf{\Lambda} \mathbf{z} ,
\end{equation}
where $\mathbf{\Lambda}$ is a diagonal matrix whose diagonal elements are given by $\mathbf{\Lambda} _{ii} = \nu_i ^{-2}$, and $\nu_i ^{2}$ is the variance of $n_i$, ${\forall i=1,2,\ldots,N}$ \cite{s1,mi1}. The goal of the attacker is to inject a false data vector $\mathbf{a} \in \mathbb{R} ^ N$ into the measurements without being detected by the operator. The resulting observation model is
\begin{eqnarray}
\mathbf{\tilde {z}} = \mathbf{H} \mathbf{x} + \mathbf{a} + \mathbf{n}.
\label{eq:a1}
\end{eqnarray}
The false data injection vector, $\mathbf{a}$, is a nonzero vector, such that $\mathbf{a}_i \neq \mathbf{0}$, $\forall i \in \mathcal{A}$, where $\mathcal{A}$ is the set of indices of the measurement variables that will be attacked. The secure variables satisfy the constraint $\mathbf{a}_i = \mathbf{0}$, $\forall i \in \bar{\mathcal{A}}$, where $\bar{\mathcal{A}}$ is the set complement of $\mathcal{A}$ \cite{mi1}. 

In order to detect an attack, the \textit{measurement residual} \cite{s1,mi1} is examined in $\ell_2$-norm $\rho= \| \mathbf{\tilde {z}} - \mathbf{H} \mathbf{\hat { x}} \|^2 _2 $, where $\mathbf{\hat {x}} \in \mathbb{R} ^ D$ is the state vector estimate. If $ \rho > \tau$, where $\tau \in \mathbb{R} $ is an arbitrary threshold which determines the trade-off between the detection and false alarm probabilities, then the network operator declares that the measurements are attacked.

One of the challenging problems of this approach is that the Jacobian measurement matrices of power systems in the smart grid are sparse under the DC power flow model \cite{mi1,mi2}. Therefore, the sparsity of the systems determines the performance of sparse state vector estimation methods \cite{candes, donoho}. In addition, unobservable attacks can be constructed even if the network operator can estimate the state vector correctly. For instance, if $\mathbf{a}= \mathbf{H} \mathbf{c}$, where $\mathbf{c} \in \mathbb{R} ^ D$ is an attack vector, then the attack is \textit{unobservable} by using the measurement residual $\rho$ \cite{s1, kosut}. In this work, we show that statistical learning methods can be used to detect the unobservable attacks with performance higher than the attack detection algorithms that employ a state vector estimation approach. Following the motivation mentioned above, a new approach is proposed using statistical learning methods.

\subsection{Attack Detection using Statistical Learning Methods}

Given a set of samples $\mathcal{S}= \{ \mathbf{s}_i \} ^M _{i=1}$ and a set of labels $\mathcal{Y}= \{ y_i \} ^M _{i=1}$, where $(\mathbf{s}_i, y_i) \in \mathcal{S} \times \mathcal{Y}$ are independent and identically distributed (i.i.d.) with joint distribution $P$, the statistical learning problem can be defined as constructing a \textit{hypothesis function} $f: \mathcal{S} \to \mathcal{Y}$, that captures the relationship between the samples and labels \cite{stl}. Then, the attack detection problem is defined as a binary classification problem, where
\begin{equation}
y_i= 
\begin{cases}
\enspace 1, & \quad {\rm if} \quad \mathbf{a}_i \neq \mathbf{0} \\
-1, & \quad {\rm if} \quad \mathbf{a}_i = \mathbf{0}
\end{cases} .
\label{eq:ss}
\end{equation}
In other words, $y_i=1$, if the $i$-th measurement is attacked, and $y_i=-1$ when there is no attack. 

In this paper, the model proposed by Ozay et al. \cite{mi_jsac} is employed for attack construction where the measurements are observed in clusters in the network. Measurement matrices, and observation and attack vectors are partitioned into $G$ blocks, denoted by $\mathcal{G}_g$ with $|\mathcal{G}_g|=N_g$ for $g=1,2, \ldots, G$. Therefore, the observation model is defined as 
\begin{eqnarray}
\begin{bmatrix} 
\mathbf{\tilde{z}}_{1} \\ 
\vdots \\
\mathbf{\tilde{z}}_{G} \\
\end{bmatrix}  
= 
\begin{bmatrix} 
\mathbf{H}_{1} \\ 
\vdots \\
\mathbf{H}_{G} \\

\end{bmatrix}
\mathbf{x}
+
\begin{bmatrix} 
\mathbf{a}_{1} \\ 
\vdots \\
\mathbf{a}_{G} \\
\end{bmatrix}
+
\begin{bmatrix} 
\mathbf{n}_{1} \\ 
\vdots \\
\mathbf{n}_{G} \\
\end{bmatrix} \;,
\label{eq:dve1}
\end{eqnarray}  
where $\tilde{\mathbf{z}}_{g} \in \mathbb{R} ^ {N_g}$ is the measurement observed in the $g$-th cluster of nodes through measurement matrix $\mathbf{H_{g}}\in\mathbb{R}^{N_g \times D}$ and noise $\mathbf{n}_{g} \in \mathbb{R} ^ {N_g}$, and which is under attack $\mathbf{a}_g \in \mathbb{R} ^ {N_g}$ with $g=1,2,\ldots,G$ \cite{mi_jsac}. Within this framework, each observed measurement vector is considered as a sample, i.e., $\mathbf{s}_i  \triangleq \mathbf{\tilde{z}}_g$, where $\mathbf{\tilde{z}}_g \in \mathbb{R} ^ {N_g}$\footnote{For simplicity of notation, we use $i$ as the index of measurements $\tilde{\mathbf{z}}_{i}$, $\mathbf{z}_i$, and attack vectors $\mathbf{a}_i$, $\forall i=1,2,\ldots,M$.}. Taking this into account, the measurements are classified in two groups, \textit{secure} and \textit{attacked}, by computing $f(\mathbf{s}_i), \forall i=1,2,\ldots,M$. 

The crucial part of the traditional attack detection algorithm, which we call State Vector Estimation (SVE), is the estimation of $\mathbf{\hat { x}}$. If the attack vectors, $\mathbf{a}$, are constructed in the column space of $\mathbf{H}$, then they are annihilated in the computation of the residual \cite{s1}. Therefore, SVE cannot detect the attacks and these attacks are called \textit{unobservable}. On the other hand, we observe that the distance between the attacked and the secure measurement vectors is defined by the attack vector in $\mathcal{S}$. If the attacks are unobservable, i.e. $\mathbf{a}_i = \mathbf{H} \mathbf{c}_i $ and $\mathbf{a}_j= \mathbf{H} \mathbf{c}_j$, where $\mathbf{c}_i \in \mathbb{R} ^ D$ and $\mathbf{c}_j \in \mathbb{R} ^ D$ are the attack vectors, then the distance between $\mathbf{\tilde{z}}_i= \mathbf{z}_i + \mathbf{a}_i$ and $\mathbf{\tilde{z}}_j= \mathbf{z}_j + \mathbf{a}_j$ is computed as 
\begin{equation}
\| \mathbf{\tilde{z}}_i - \mathbf{\tilde{z}}_j \| _2 = 
\begin{cases}
 \| \mathbf{{z}}_i - \mathbf{{z}}_j \|_2 + \| \mathbf{{a}}_i - \mathbf{{a}}_j \|_2 , & \enspace {\rm if} \quad i,j \in \mathcal{A} \\
 \| \mathbf{{z}}_i - \mathbf{{z}}_j \|_2 + \| \mathbf{{a}}_i \|_2 , & \enspace {\rm if} \quad i \in \mathcal{A}, \; j \in \bar{\mathcal{A}} \\
   \| \mathbf{{z}}_i - \mathbf{{z}}_j \|_2 , & \enspace {\rm if} \quad i,j \in \bar{\mathcal{A}} \\ 
\end{cases} ,
\label{eq:dm}
\end{equation}
where $\mathbf{\tilde{z}}_i \in \mathcal{S}$ and $\mathbf{\tilde{z}}_j \in \mathcal{S}$. In \eqref{eq:dm}, we can extract information on the attack vectors by observing the measurements. Since the distances between secure and attacked measurements are discriminated by the attack vectors, the attacks can be recognized by the learning algorithms which use the information of these distances, even if the attacks are \textit{unobservable}. 

Two main assumptions from statistical learning theory need to be taken into account to classify measurements which satisfy \eqref{eq:dm}: 

\begin{enumerate}
\item We assume that $(\mathbf{s}_i, y_i) \in \mathcal{S} \times \mathcal{Y}$ are distributed according to a joint distribution $P$ \cite{kulkarni2}. In a smart grid setting, this \textit{distribution assumption} is satisfied for the attack models in which the measurements $\mathbf{\tilde{z}}$ are functions of $ \mathbf{a}$, and we can extract statistical information about both the attacked and secure measurements from the observations. 

\item We assume that $(\mathbf{s}_i, y_i)$, $\forall i$, are sampled from $P$, independently and identically. This assumption is also satisfied in the smart grid if the entries of $\mathbf{n}$ and $\mathbf{a}$ are i.i.d. random variables \cite{stl}.
\end{enumerate}

In order to explain the significance of the above assumptions in the smart grid, we consider the following example. Assume that measurements $1,2 \in \mathcal{A}$ and $3,4 \in \bar{\mathcal{A}}$, are given such that $y_1, y_2=1$ and $y_3,y_4=-1$. Furthermore, assume that $\mathbf{{z}}_1=3 \cdot \mathbf{I}$, $\mathbf{{z}}_2=5 \cdot \mathbf{I}$, $\mathbf{{z}}_3=2 \cdot \mathbf{I}$ and $\mathbf{{z}}_4=4 \cdot \mathbf{I}$, where $\mathbf{I}=(1, 1)^T$. If the attack vectors are \textit{identical} but not independent, then the attack vectors can be constructed as $\mathbf{{a}}_1=\mathbf{{a}}_2=-1 \cdot \mathbf{I}$. As a result, we observe that $\tilde{\mathbf{{z}}}_1=\tilde{\mathbf{{z}}}_3=2 \cdot \mathbf{I}$ and $\tilde{\mathbf{{z}}}_2=\tilde{\mathbf{{z}}}_4=4 \cdot \mathbf{I}$. Therefore, our assumption about the existence of a joint distribution $P$ is not satisfied and we cannot classify the measurements with the aforementioned approach.     

\section{Attack Detection using Machine Learning Methods}
\label{sec:methods}
In this section, the attack detection problem is modeled by statistical classification of measurements using machine learning methods. 


\subsection{Supervised Learning Methods}
\label{sec:supervised}
In the following, the classification function $f$ is computed in a \textit{supervised learning} framework by a network operator using a set of training data ${\sf Tr}= \{ (\mathbf{s}_i, y_i) \} _{i=1} ^{M^{\sf Tr}}$. The class label, $y'_{i}$, of a new observation, $\mathbf{s}'_i$, is predicted using $y'_{i}=f(\mathbf{s}'_{i})$. We employ four learning algorithms for attack detection.

\subsubsection{Perceptron}
Given a sample $\mathbf{s}_{i}$, a perceptron predicts $y_i$ using the classification function  $f(\mathbf{s}_{i})= {\sf sign}( \mathbf{w} \cdot \mathbf{s}_{i} )$, where $\mathbf{w} \in \mathbb{R} ^ {N_i}$ is a weight vector and ${\sf sign}( \mathbf{w} \cdot \mathbf{s}_{i} )$ is defined as \cite{kulkarni2}
\begin{equation}
{\sf sign}( \mathbf{w} \cdot \mathbf{s}_{i} )= 
\begin{cases}
 -1, & \quad {\rm if} \quad \mathbf{w} \cdot \mathbf{s}_{i} < 0 \\
\enspace 1, & \quad {\rm otherwise.}  
\end{cases} 
\label{eq:sign}
\end{equation} 

In the training phase, the weights are adjusted at each iteration $t=1, 2, \ldots, T$ of the algorithm for each training sample using 
\begin{equation}
\mathbf{w}(t+1) := \mathbf{w}(t) + \Delta \mathbf{w},
\end{equation}
where $\Delta \mathbf{w} = \gamma (y_i - f(\mathbf{s}_{i}) ) \mathbf{s}_{i}$ and $\gamma$ is the \textit{learning rate}. The algorithm is iterated until a stopping criterion, such as the number of algorithm steps, or an error threshold, is achieved. In the testing phase, the label of a new test sample is predicted by $f(\mathbf{s'}_{i})={\sf sign}( \mathbf{w}(T) \cdot \mathbf{s'}_{i} )$.

Despite its success in various machine learning applications, the convergence of the algorithm is assured only when the samples are linearly separable \cite{kulkarni2}. For that reason, the perceptron can be successfully used for the detection of the attacks only if the measurements can be separated by a hyperplane. In the following sections, we give examples of classification algorithms which overcome this limitation by employing non-linear classification rules or feature extraction methods.

\subsubsection{$k$-Nearest Neighbor ($k$-NN)}

This algorithm labels an unlabeled sample $\mathbf{s}'_{i}$ according to the labels of its $k$-nearest neighborhood in the feature space \cite{kulkarni2}. Specifically, the observed measurements $\mathbf{s}_i \in \mathcal{S}$, $\forall i=1,2,\ldots,M$, are taken as feature vectors. The set of $k$-nearest neighbors of $\mathbf{s}'_{i}$,  $ \aleph(\mathbf{s}'_{i})= \{ \mathbf{s}_{i(1)},\mathbf{s}_{i(2)}, \ldots,\mathbf{s}_{i(k)} \}$, is constructed by computing the Euclidean distances between the samples \cite{k2}, where $i(1), i(2), \ldots, i(M)$ are defined as 
\begin{equation} 
\| \mathbf{{s'}}_i - \mathbf{s}_{i(1)} \|_2 \leq \| \mathbf{{s'}}_i - \mathbf{s}_{i(2)} \|_2 \leq \ldots \leq \| \mathbf{{s'}}_i - \mathbf{s}_{i(M)} \|_2 .
\end{equation}
Then, the most frequently observed class label is computed using majority voting among the class labels of the samples in the neighborhood, and assigned as the class label of $\mathbf{s}'_{i}$ \cite{teo}. One of the challenges of $k$-NN is the \textit{curse of dimensionality}, which is the difficulty of the learning problem when the sample size is small compared to the dimension of the feature vector \cite{kulkarni2,teo,duda}. In attack detection, this problem can be handled using the following approaches:
\begin{itemize}
\item Feature selection algorithms can be used to reduce the dimension of the feature vectors \cite{teo,duda}. Development of feature selection algorithms may be a promising direction for smart grid security, and is an interesting topic for future work. 
\item Kernel machines, such as SVMs, can be used to map the feature vectors in $\mathcal{S}$ to Hilbert spaces, where the feature vectors are processed implicitly in the mappings and the computation of the learning models. We give the details of the kernel machines and SVMs in the following sections.
\item The samples can be processed in small sizes, e.g. by selecting a single measurement vector as a sample, which leads to one-dimensional samples. We employ this approach in Section IV. If the sample size is large, distributed learning and optimization methods can be used \cite{mi_jsac,mi2}.
\end{itemize}

\begin{figure}[ht!]
\centering
\subfloat[Attack detection using the linearly separable dataset.]{\includegraphics[width=2.90in, height=2.20in]{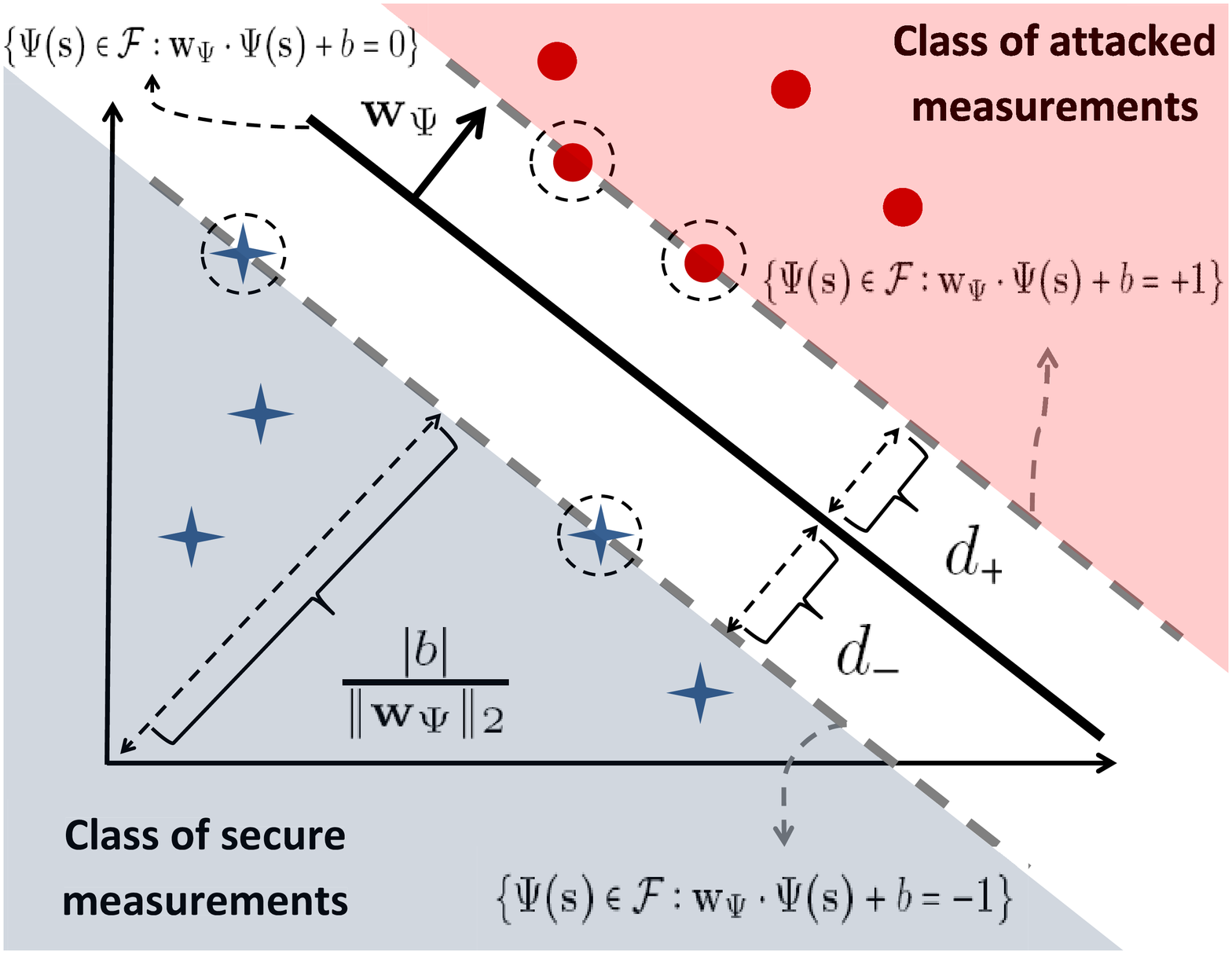}\label{fig:svm_lin}}  \\
\subfloat[Attack detection using the linearly non-separable dataset.]{\includegraphics[width=2.9in, height=2.20in]{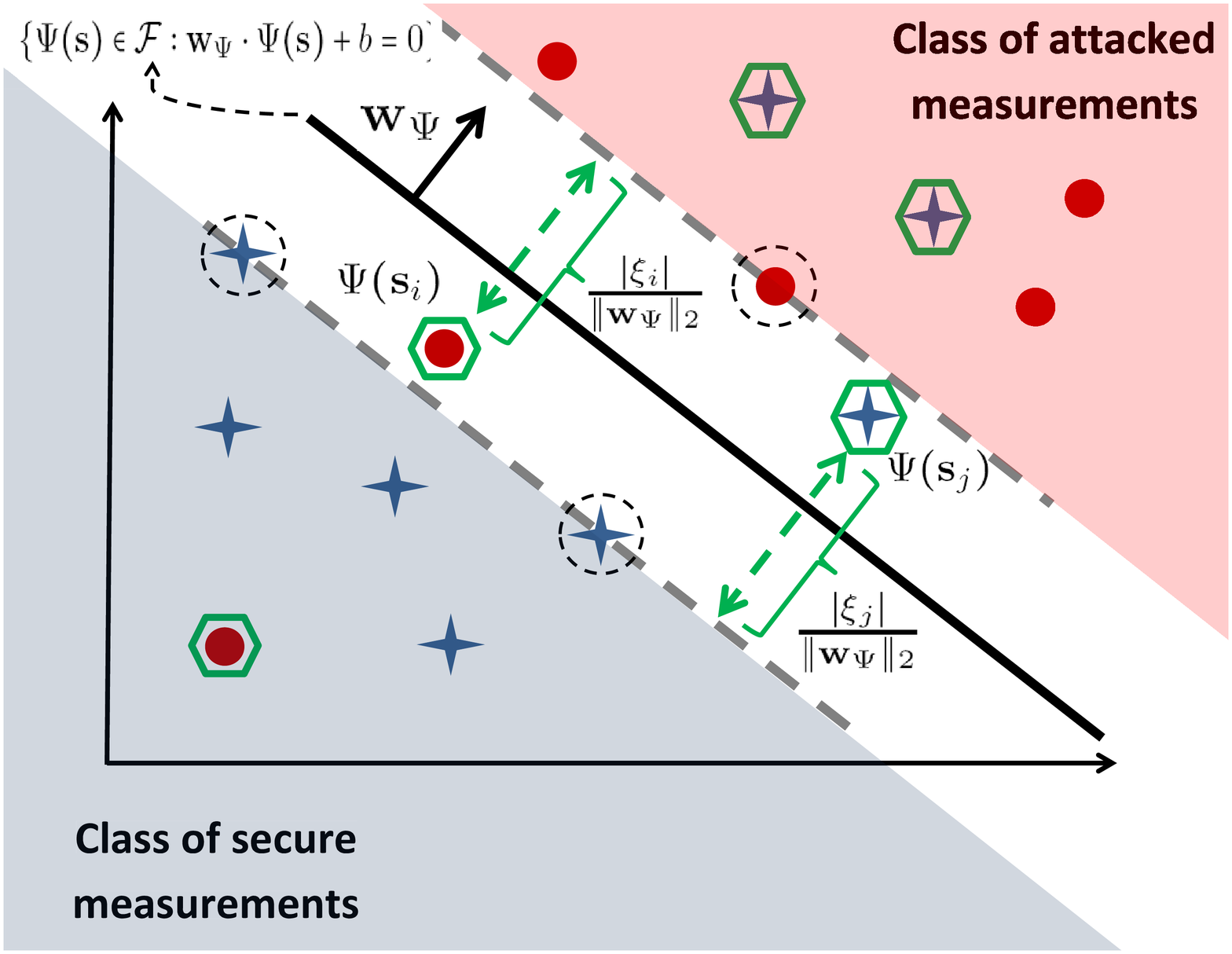}\label{fig:svm_nonlin}} 
\caption{Classification using SVM. Positive and negative samples which belong to the class of attacked and secure measurements are depicted by disk and star markers, respectively. Support vectors and misclassified samples are depicted by dashed circles and hexagonal markers, respectively.}
\label{fig:svm}
\end{figure}

\subsubsection{Support Vector Machines}

We seek a hyperplane that linearly separates attacked and secure measurements   into two half spaces using hyperplanes in a $D'$  dimensional feature space, $\mathcal{F}$, which is constructed by a non-linear mapping $\Psi : \mathcal{S} \to \mathcal{F}$ \cite{mi1,svm}. A hyperplane is represented by a weight vector $\mathbf{w} _{\Psi} \in \mathbb{R}^{D'}$ and a bias variable $b \in \mathbb{R}$, which results in 
\begin{eqnarray}
\begin{matrix} 
\mathbf{w}_{\Psi} \cdot \Psi(\mathbf{s}) + b=0,
\end{matrix}  
\label{eq:hyper}
\end{eqnarray}
where $\Psi(\mathbf{s})$ is the feature vector of the sample that lies on the hyperplane in $\mathcal{F}$ as shown in Fig.~\ref{fig:svm}. We choose the hyperplane that is at the largest distance from the closest positive and negative samples. This constraint can be formulated as 
\begin{eqnarray}
\begin{matrix} 
y_i(\mathbf{w}_{\Psi} \cdot \Psi(\mathbf{s}) + b) -1 \geq 0, \quad \forall i=1,2,\ldots,M^{\sf Tr}.
\end{matrix}  
\label{eq:margin}
\end{eqnarray}
Since $d_+ = d_- = \frac{1}{ \| \mathbf{w}_{\Psi} \| _2 }$, where $d_+$ and $d_-$ are the shortest distances from the hyperplane to the closest positive and negative samples respectively, a maximum margin hyperplane can be computed by minimizing $\| \mathbf{w}_{\Psi} \| _2$. 

If the training examples in the transformed space are not linearly separable (see Fig. 1.b), then the optimization problem can be modified by introducing slack variables $\xi_i\geq 0$, ${\forall i=1,2,\ldots,M^{\sf Tr}}$, in \eqref{eq:margin} which yields
\begin{eqnarray}
\begin{matrix} 
y_i(\mathbf{w}_{\Psi} \cdot \Psi(\mathbf{s}_i) + b) -1 + \xi_i \geq 0 \; , \forall i=1,2,\ldots,M^{\sf Tr}.
\end{matrix}  
\label{eq:margin2}
\end{eqnarray}
The hyperplane $\mathbf{w}_{\Psi} $ is computed by solving the following optimization problem in primal or dual form \cite{svm,kulkarni,primal_svm}
\begin{eqnarray}
\begin{matrix} 
\text{minimize} & \| \mathbf{w}_{\Psi} \| _2 ^2 + \displaystyle C \sum \limits ^{M^{\sf Tr}} _{i=1} \xi_i  \\
\text{subject to} &\;\quad\quad\quad\quad\quad y_i(\mathbf{w}_{\Psi} \cdot \Psi(\mathbf{s}_i) + b) -1 + \xi_i \geq 0 \\
&\quad\quad\quad\xi_i \geq 0,\quad \forall i=1,2,\ldots,M^{\sf Tr} 
\end{matrix} 
\label{eq:primal_svm} 
\end{eqnarray}
where $C$ is a constant that penalizes (an upper bound on) the training error of the soft margin SVM.

\subsubsection{Sparse Logistic Regression}
In utilizing this approach for attack detection, we solve the classification problem using the Alternating Direction Method of Multipliers (ADMM) \cite{admm} considering the sparse state vector estimation approach of Ozay et al. \cite{mi_jsac}. Note that, the hyperplanes defined in \eqref{eq:hyper} can be computed by employing the generalized logistic regression models presented in \cite{teo}, which provide the distributions
\begin{eqnarray}
P(y_i | \mathbf{s}_i)&=& \frac{1}{1+ \exp(-y_i (\mathbf{w} \cdot \mathbf{s}_i + b)) }, \\
P(y_i | \Psi(\mathbf{s}_i))&=&\frac{1}{1+ \exp(-y_i(\mathbf{w}_{\Psi} \cdot \Psi(\mathbf{s}_i) + b))},
\label{eq:log_reg2}
\end{eqnarray}
in $\mathcal{S}$ and $\mathcal{F}$, respectively. For this purpose, we minimize the \textit{logistic loss} functions
\begin{equation}
\mathcal{L}(\mathbf{s}_i,y_i)= \log\left(1+ \exp\left(-y_i (\mathbf{w} \cdot \mathbf{s}_i + b)\right)\right), 
\label{eq:log_reg3_1}
\end{equation}
\begin{equation}
\mathcal{L}(\Psi(\mathbf{s}_i),y_i)= \log\left(1+ \exp(-y_i(\mathbf{w}_{\Psi} \cdot \Psi(\mathbf{s}_i) + b))\right).
\label{eq:log_reg3}
\end{equation}
Defining a feature matrix $\mathbf{S} =  (\mathbf{s} _1 ^{T},\mathbf{s} _2 ^{T},\ldots,\mathbf{s} ^{T} _{M_{tr}})^T$ and a label vector $\mathbf{Y} = (y_1,y_2,\ldots,y_{M_{tr}}) ^T$,  the ADMM optimization problem \cite{admm} is constructed as 
\begin{eqnarray}
\begin{matrix} 
& \text{minimize} && \mathcal{L}(\mathbf{S},\mathbf{Y})+ \mu(\mathbf{r}) \\
& \text{subject to} && \mathbf{ w} - \mathbf{r} =\mathbf{0}
\end{matrix}  
\label{eq:dist_log_reg}
\end{eqnarray}
where $\mathbf{ w}$ is a weight vector, $\mathbf{r}$ is a vector of optimization variables, $\mu(\mathbf{r})= \lambda \| \mathbf{r} \| _1$ is a regularization function, and $\lambda$ is a regularization parameter which is introduced to control the sparsity of the solution \cite{admm}.

\subsection{Semi-supervised Learning Methods}
In semi-supervised learning methods, the information obtained from the unlabeled test samples is used during the computation of the learning models \cite{semibook}. 

In this section, a semi-supervised Support Vector Machine algorithm, called Semi-supervised SVM (S3VM) \cite{s3vm,semiopt} is employed to establish the analytical relationship between supervised and semi-supervised learning algorithms. In this setting, the unlabeled samples are incorporated into cost function of the optimization problem (\ref{eq:primal_svm}) as
\begin{eqnarray}
\begin{matrix} 
\text{minimize} & \| \mathbf{w} \| _2 ^2 + \displaystyle C_1 \sum \limits ^{M^{\sf Tr}} _{i=1} L^{\sf Tr}(\mathbf{s}_i,y_i) + C_2 \sum \limits ^{M^{\sf Te}} _{i=1} L^{\sf Te}(\mathbf{s'}_i)  ,
\end{matrix}  
\label{eq:s3vm}
\end{eqnarray}
where $C_1$ and $C_2$ are confidence parameters, and $ L^{\sf Tr}(\mathbf{s}_i,y_i)= {\rm max} (0,1-y_i( \mathbf{w} \mathbf{s}_i +b))$ and $ L^{\sf Te}(\mathbf{s'}_i)= {\rm max} (0, 1- \| \mathbf{s'}_i \| _1 )$ are the loss functions of the training and test samples, respectively. 

The main assumption of the S3VM is that the samples in the same cluster have the same labels and the number of sub-clusters is not large \cite{semiopt}. In other words, attacked and secure measurement vectors should be clustered in distinct regions in the feature spaces. Moreover, the difference between the number of attacked and secure measurements should not be large in order to avoid the formation of sub-clusters. 

This requirement can be validated by analyzing the feature space. Following (\ref{eq:dm}), if $ \| \mathbf{{z}}_i - \mathbf{{z}}_j \|_2 + \| \mathbf{{a}}_i - \mathbf{{a}}_j \|_2 \leq \| \mathbf{{a}}_i \|_2 + \| \mathbf{{a}}_j \|_2$, and $ \| \mathbf{{z}}_k - \mathbf{{z}}_l \|_2 \leq \| \mathbf{{a}}_i \|_2 + \| \mathbf{{a}}_j \|_2$, $\forall  i,j \in \mathcal{A}$ and $ \forall k,l \in \bar{\mathcal{A}}$, then the samples belonging to different classes are well-separated in different classes. Moreover, this requirement is satisfied in (\ref{eq:s3vm}) by adjusting $C_2$ \cite{semiopt}. A survey of the methods which are used to provide \textit{optimal} $C_2$ and solve (\ref{eq:s3vm}) is given in \cite{semiopt}. 

\subsection{Decision and Feature Level Fusion Methods}

One of the challenges of statistical learning theory is to find a classification rule that performs better than a set of rules of individual classifiers, or to find a feature set that represents the samples better than a set of individual features. One approach to solve this problem is to combine a collection of classifiers or a set of features to boost the performance of the individual classifiers. The former approach is called decision level fusion or ensemble learning, and the latter approach is called feature level fusion. In this section, we consider Adaboost \cite{adaboost} and Multiple Kernel Learning (MKL) \cite{mkl} for ensemble learning and feature level fusion.

\subsubsection{Ensemble Learning for Decision Level Fusion}
\label{sec:ensemble}
Various methods such as bagging, boosting and stacking have been developed to combine classifiers in ensemble learning situations \cite{kulkarni2,comb}. In the following, Adaboost is explained as an ensemble learning approach, in which a collection of \textit{weak} classifiers are generated and combined using a combination rule to construct a \textit{stronger} classifier which performs better than the \textit{weak} classifiers \cite{kulkarni2,adaboost,boost_book}.  

At each iteration $t=1,2,\ldots, T$ of the algorithm, a decision or hypothesis $f_t( \cdot )$ of the weak classifier is computed with respect to the distribution on the training samples $D_t( \cdot )$ at $t$ by minimizing the weighted error ${\epsilon _t = \sum \limits ^{M^{\sf Tr}} _{i=1} D_t(i)I( f_t( \mathbf{s}_i) \neq y_i)}$, where $I( \cdot )$ is the indicator function. The distribution is initialized uniformly $D_1(i)=\frac{1}{M^{\sf Tr}}$ at $t=1$, and is updated by a parameter $\alpha_t=\frac{1}{2} \log ( \frac{1-\epsilon_t}{\epsilon_t} )$ as follows \cite{boost_book}
\begin{equation}
D_{t+1}(i)=\frac{ D_{t}(i) \exp ^{- \alpha_t y_i f_t( \mathbf{s}_i)} } {Z_t} ,
\label{eq:adaboost}
\end{equation} 
where $Z_t$ is a normalization parameter, called the \textit{partition function}. At the output of the algorithm, a strong classifier $H( \cdot )$ is constructed for a sample $\mathbf{s'}$ using $H( \mathbf{s'} )= {\sf sign}( \sum  \limits ^T _{t=1} \alpha_t f_t(\mathbf{s'}) )$.

\subsubsection{Multiple Kernel Learning for Feature Level Fusion}

Feature level fusion methods combine the feature spaces instead of the decisions of the classifiers. One of the feature level fusion methods is MKL in which different feature mappings are represented by kernels that are combined to produce a new kernel which represents the samples better than the other kernels \cite{mkl}. Therefore, MKL provides an approach to solve the feature mapping selection problem of SVM. In order to see this relationship, we first give the dual form of \eqref{eq:primal_svm}
\begin{eqnarray}
\begin{matrix} 
\text{maximize} & \sum \limits ^{M^{\sf Tr}} _{i=1} \beta_i - \frac{1}{2} \sum \limits ^{M^{\sf Tr}} _{i=1} \sum  \limits ^{M^{\sf Tr}} _{j=1} \beta_i \beta_j y_i y_j k(\mathbf{s}_i,\mathbf{s}_j) \\ 
\text{subject to} &\; \sum \limits ^{M^{\sf Tr}} _{i=1} \beta_i y_i=0 \\
& 0 \leq \beta_i \leq C, \quad \forall i=1,2,\ldots,M^{\sf Tr} , 
\end{matrix} 
\label{eq:dual_svm} 
\end{eqnarray}
where $\beta_i$ is the dual variable and $k(\mathbf{s}_i,\mathbf{s}_j)=\Psi(\mathbf{s}_i) \cdot \Psi(\mathbf{s}_j)$ is the kernel function. Therefore, \eqref{eq:dual_svm} is a single kernel learning algorithm which employs a single kernel matrix $K \in \mathbb{R} ^{ M^{\sf Tr} \times M^{\sf Tr} }$ with elements $K(i,j)=k(\mathbf{s}_i,\mathbf{s}_j)$. If we define the weighted combination of $U$ kernels as $K=\sum \limits ^{U} _{u=1} d_u K_u$, where $d_u \geq 0$ are the normalized weights such that $\sum \limits ^{U} _{u=1} d_u =1$, then we obtain the following optimization problem of the MKL \cite{simplemkl}:
\begin{eqnarray}
\begin{matrix} 
\text{maximize} & \sum \limits ^{M^{\sf Tr}} _{i=1} \beta_i - \frac{1}{2} \sum \limits ^{M^{\sf Tr}} _{i=1} \sum  \limits ^{M^{\sf Tr}} _{j=1} \beta_i \beta_j y_i y_j \sum \limits ^{U} _{u=1} d_u K_u(\mathbf{s}_i,\mathbf{s}_j)  \\
\text{subject to} &\; \sum \limits ^{M^{\sf Tr}} _{i=1} \beta_i y_i=0 \\
& 0 \leq \beta_i \leq C, \quad \forall i=1,2,\ldots,M^{\sf Tr} . 
\end{matrix} 
\label{eq:dual_mkl} 
\end{eqnarray}
In \eqref{eq:dual_mkl}, the kernels with $d_u=0$ are eliminated, and therefore MKL can be considered as a kernel selection method. In the experiments, SVM algorithms are implemented with different kernels and these kernels are combined under MKL.

\subsection{Online Learning Methods for Real-time Attack Detection} 
\label{sec:online}
In the smart grid, the measurements are observed in real-time where the samples are collected sequentially in time. In this scenario, we relax the distribution assumption of Section II.B, since the samples are observed in an arbitrary sequence \cite{kakade}. Moreover, \textit{smart} PMUs which employ learning algorithms, may be required to detect the attacks when the measurements are observed without processing the whole set of training samples. 
In order to solve these challenging problems, we may use online versions of the learning algorithms given in the previous sections.

In a general online learning setting, a sequence of training samples (or a single sample) is given to the learning algorithm at each observation or algorithm processing time. Then, the algorithm computes the learning model using only the given samples and predicts the labels. The learning model is updated with respect to the error of the algorithm which is computed using a loss function on the given samples. Therefore, the perceptron and Adaboost are convenient for online learning in this setting. For instance, an online perceptron is implemented by predicting the label $y_i$ of a single sample $\mathbf{s}_i$ at each time $t$, and updating the weight vector $\mathbf{w}$ using $\Delta \mathbf{w}$ for the misclassified samples with $y_i \neq {\sf sign} (f(\mathbf{s}_{i}))$ \cite{online}. This simple approach is applied for the development of online MKL \cite{online} and regression algorithms \cite{online_l1}. 

\subsection{Performance Analysis} 

In smart grid networks, the major concern is not just the detection of attacked variables, but also that of the secure variables with high performance. In other words, we require the algorithms to predict the samples with high precision and recall performance in order to avoid  false alarms. Therefore, we measure the true positives (\textit{tp}), the true negatives ({\it tn}), the false positives ({\it fp}), and the false negatives ({\it fn}), which are defined in Table \ref{tab:addlabel}.  
\begin{table}[htbp]
  \centering
  \caption{Definitions of performance measures }
\begin{tabular}{r|c|c|}
\multicolumn{1}{r}{}
 &  \multicolumn{1}{c}{Attacked}
 & \multicolumn{1}{c}{Secure} \\
\cline{2-3}
Classified as Attacked & \textit{tp} & \textit{fp} \\
\cline{2-3}
Classified as Secure & \textit{fn} & \textit{tn} \\
\cline{2-3}
\end{tabular}
  \label{tab:addlabel}%
\end{table}%

In addition, the learning abilities and memorization properties of the algorithms are measured by Precision (\textit{Prec}), Recall (\textit{Rec}) and Accuracy (\textit{Acc}) values which are defined as \cite{mi1}
\begin{eqnarray}
\begin{matrix} 
& Prec=\frac{tp}{tp+fp}, & Rec=\frac{tp}{tp+fn}, & Acc=\frac{tp+tn}{tp+tn+fp+fn}.
\end{matrix} 
\label{eq:perf} 
\end{eqnarray}

Precision values give information about the prediction performance of the algorithms. On the other hand, Recall values measure the degree of \textit{attack retrieval}. Finally, the total classification performance of the algorithms is measured by Accuracy. For instance, if $Prec=1$, then none of the secure measurements is misclassified as attacked. If $Rec=1$, then none of the attacked measurements is misclassified as secure. If $Acc=1$, then each measurement classified as attacked is actually exposed to an attack, and each measurement classified as secure is actually a secure measurement.  

\section{Experiments}
\label{sec:exp}

The classification algorithms are analyzed in IEEE 9-bus, 57-bus and 118-bus test systems in the experiments. The measurement matrices $\mathbf{H}$ of the systems are obtained from the MATPOWER toolbox \cite{mp}. The operating points of the test systems provided in the MATPOWER case files are used in generating $\mathbf{z}$. Training and test data are generated by repeating this process $50$ times for each simulated point and dataset. In the  experiments, we assume that the attacker has access to $\kappa$ measurements which are randomly chosen to generate a $\kappa$-sparse attack vector $\mathbf{a}$ with Gaussian distributed  nonzero elements with the same mean and variance as the entries of $\mathbf{z}$ \cite{mi_jsac,mi1,mi2}. We assume that concept drift \cite{k_shift} and dataset shift \cite{datashift} do not occur. Therefore, we use $G=N$ in the simulations following the results of Ozay et al. \cite{mi_jsac}.

We analyze the behavior of each algorithm on each system for both observable and unobservable attacks by generating attack vectors with different values of $\frac{\kappa}{N} \in [0,1]$. More precisely, if ${\kappa\geq N-D+1}$, then attack vectors that are not observable by SVE, i.e. \textit{unobservable} attacks, are generated \cite{mi_jsac}. Otherwise, the generated attacks are \textit{observable}. 

The LIBSVM  \cite{libsvm} implementation is used for the SVM, and the ADMM \cite{admm} implementation is used for Sparse Logistic Regression (SLR). $k$ values of the $k$-NN algorithm are optimized by searching $k \in \{ 1,2,\ldots,\sqrt{M^{\sf Tr}} \}$ using leave-one-out cross-validation, where $M^{\sf Tr}$ is the number of training samples. Both the linear and Gaussian kernels are used for the implementation of SVM. A grid search method \cite{libsvm,liblinear,params} is employed to search the parameters of the SVM in an interval $\mathcal{I} = [ \mathcal{I}_{min}, \mathcal{I}_{max} ]$, where $\mathcal{I}_{min}$ and $\mathcal{I}_{max}$ are user defined values. In order to follow linear paths in the search space, $\log$ values of parameters are considered in the grid search method \cite{params}. Keerthi and Lin \cite{params} analyzed the asymptotic properties of the SVM for $\mathcal{I} = [0, \infty)$. In the experiments, ${\mathcal{I}_{min} = -10}$ is chosen to compute a lower limit $2^{-10}$ of the parameter values  following the theoretical results given in \cite{libsvm} and \cite{params}. Since the classification performance of the SVM does not change for parameter values that are greater than a threshold \cite{params}, we used $\mathcal{I}_{max} =10$ as employed in the experimental analyses in \cite{params}. Therefore, the kernel width parameter $\sigma$ of a Gaussian kernel is searched in the interval $\log(\sigma)\in [-10,10]$  and the cost penalization parameter $C$ of the SVM is searched in the interval $\log(C)\in [-10,10]$. The regularization parameter of the SLR is computed as 
\begin{equation}
\lambda = \Omega \lambda_{max},
\label{eq:lambda}
\end{equation} 
where $\lambda_{max}= \| \mathbf{H} \tilde{\mathbf{z}} \| _{\infty}$ determines the critical value of $\lambda$ above which the solution of the ADMM problem is $\mathbf{0}$ and $\Omega$ is searched for in the interval $\Omega \in [10^{-3},1]$ \cite{mi_jsac,admm,par_log}. An \textit{optimal} $\hat{\lambda}$ is computed by analyzing the solution (or regularization) path of the LASSO type optimization algorithms using a given training dataset. As the sparsity of the systems that generate datasets increases, lower values are calculated for $\Omega$ \cite{mi_jsac,admm,par_log}.  The absolute and relative tolerances, which determine values of upper bounds on the Euclidean norms of primal and dual residuals, are chosen as $10^{-4}$ and $10^{-2}$, respectively \cite{admm}. The penalty parameter is initially selected as $1$ and dynamically updated at each iteration of the algorithm \cite{admm}. The maximum number of iterations is chosen as $10^4$ \cite{mi_jsac,admm}. 

In the experiments, we observe that the selection of tolerance parameters does not affect the convergence rates if their relative values do not change. In addition,  selection of the initial value of the penalty parameter also does not affect the convergence rate if relative values of tolerance parameters are fixed \cite{admm}. For instance, similar convergence rates are observed when we chose $10^{-4}$ and $10^{-2}$, or $10^{-6}$ and $10^{-4}$, as tolerance parameters. 
$ \| \mathbf{\tilde{z}} - \mathbf{H} \mathbf{\hat x_b} \| \leq \tau$ is computed in order to decide whether there is an attack using the SVE and assuming a chi-square test with $95 \%$ confidence in the computation of $\tau$ \cite{book,mi1}.

\subsection{Results for Supervised Learning Algorithms}

\begin{figure}[t]
\centering
\subfloat[SVE.]{\includegraphics[width=1.7in, height=1.5in]{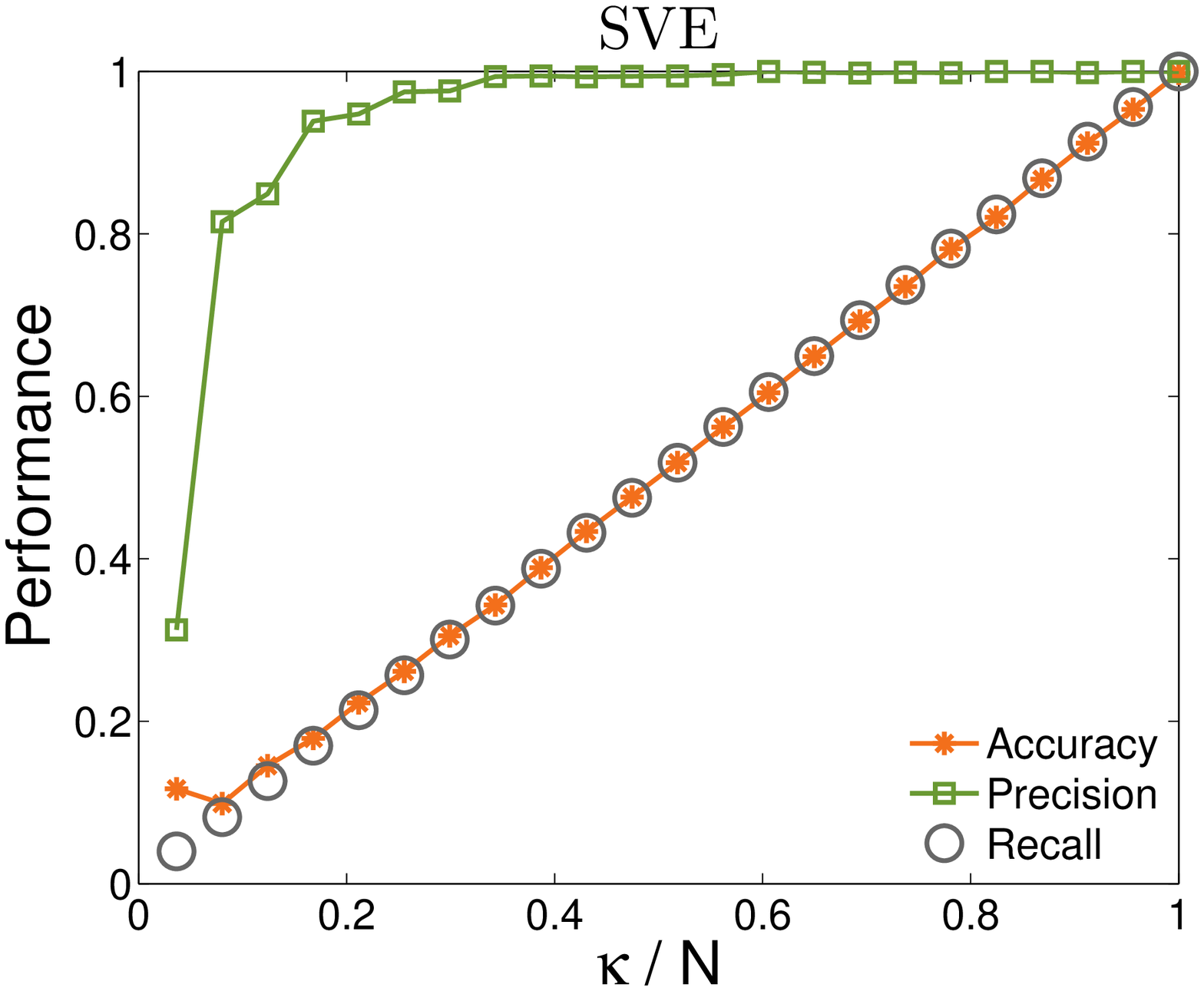}\label{fig:sve_57}} 
\subfloat[Perceptron.]{\includegraphics[width=1.7in, height=1.5in]{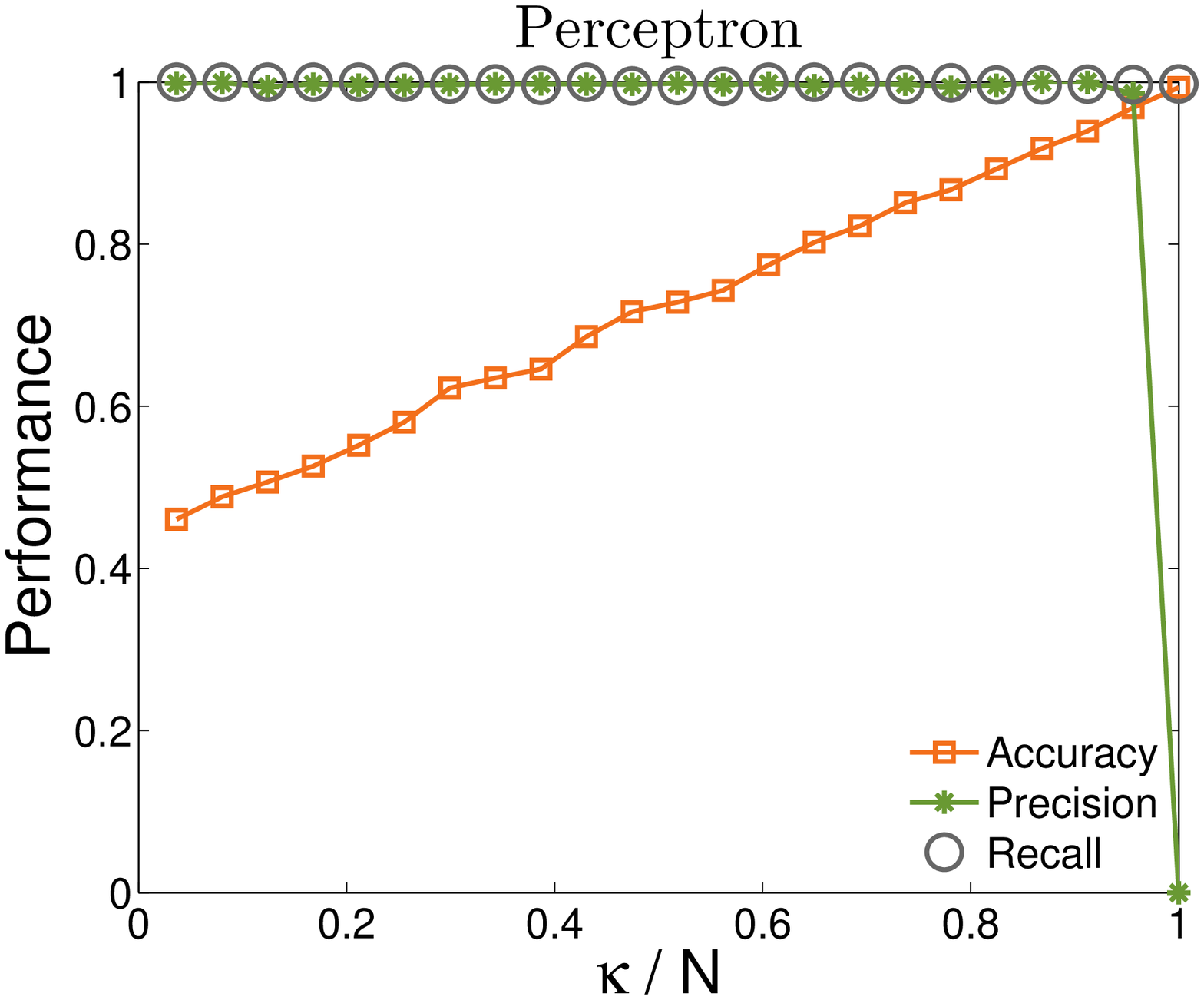}\label{fig:perceptron_57}} \\
\subfloat[SVM with linear kernel.]{\includegraphics[width=1.7in, height=1.5in]{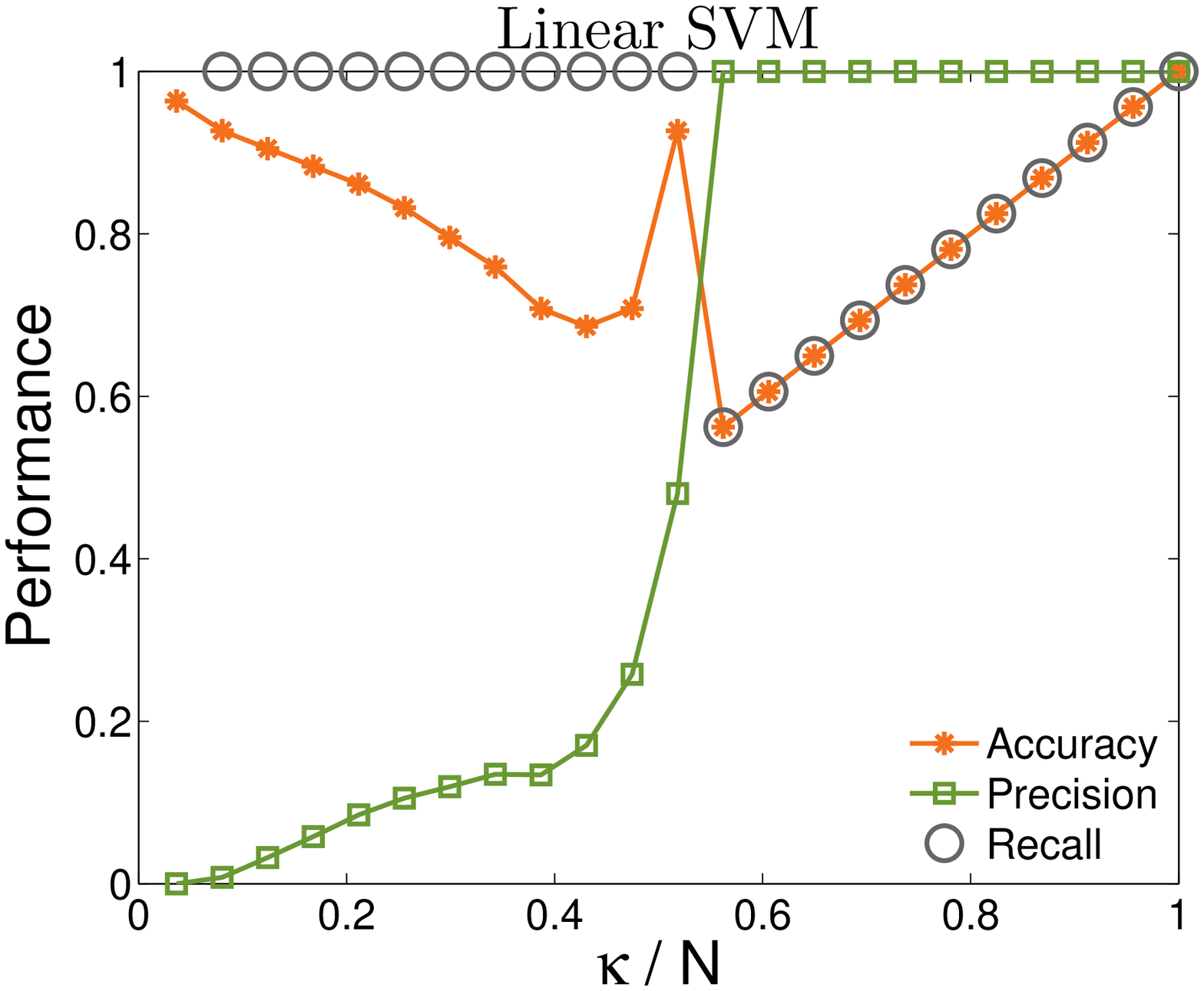}\label{fig:svm_57}} 
\subfloat[$k$-NN.]{\includegraphics[width=1.7in, height=1.5in]{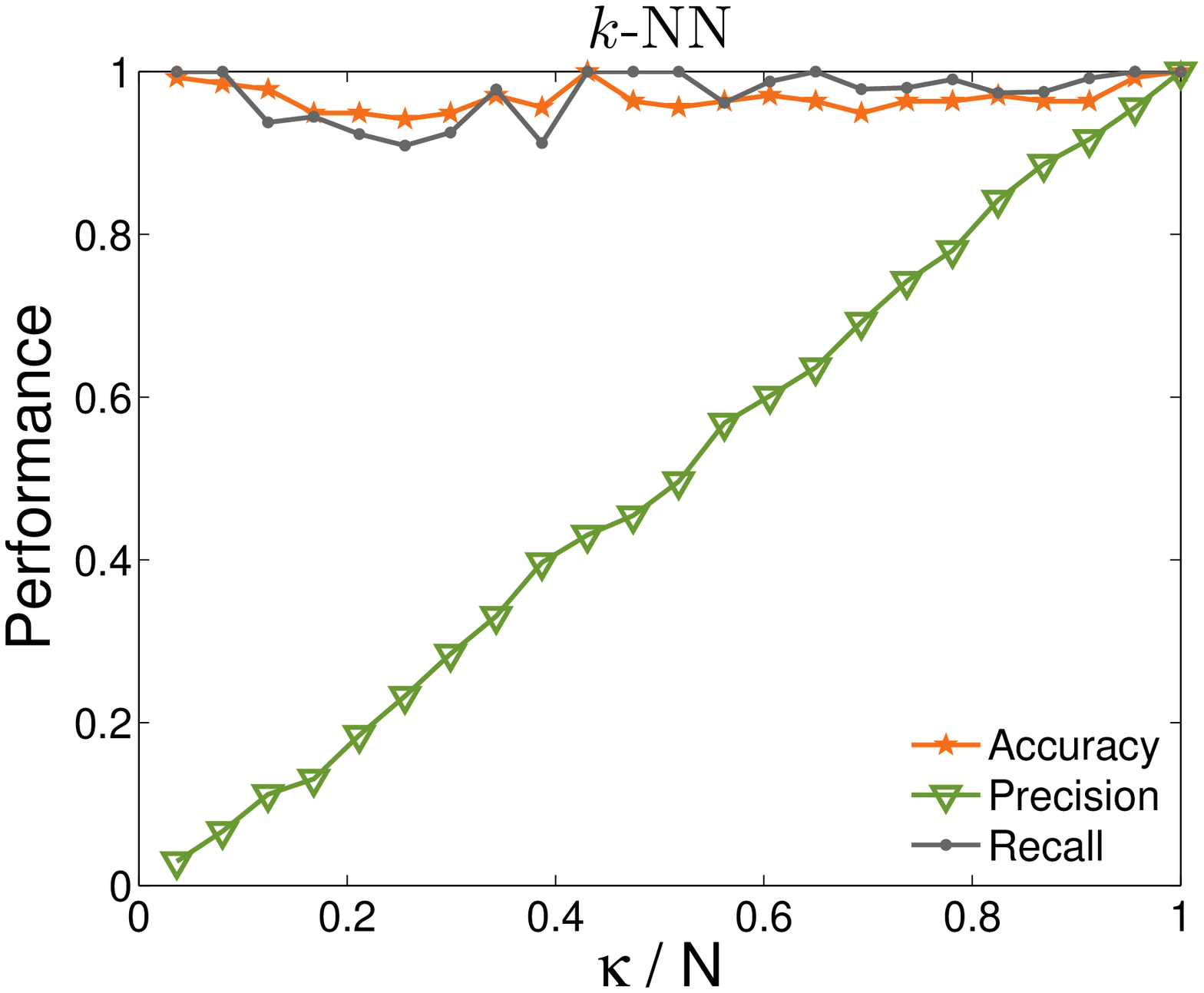}\label{fig:knn_57}} \\
\subfloat[SLR.]{\includegraphics[width=1.7in, height=1.5in]{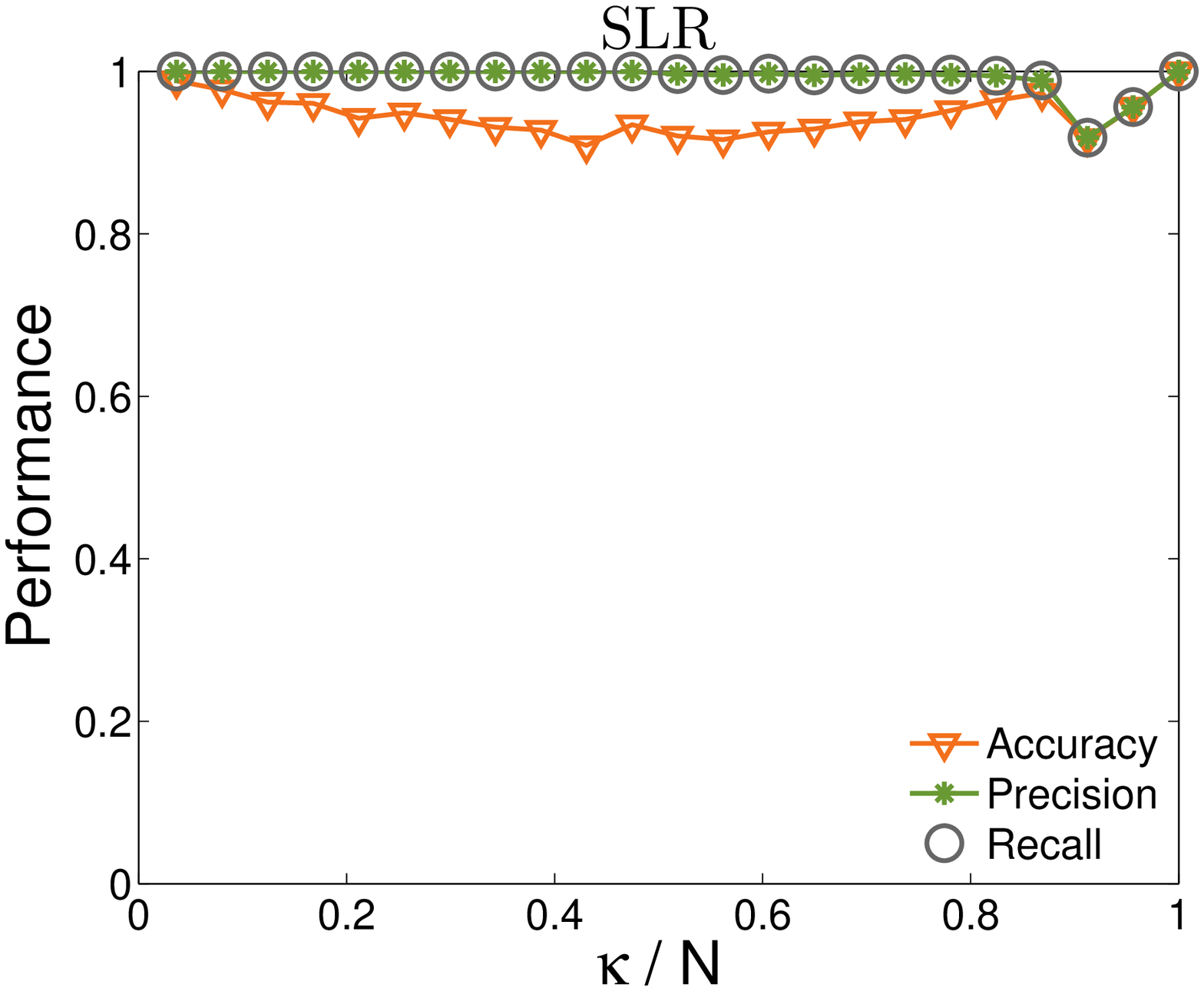}\label{fig:slr_57}} 
\caption{Results for the IEEE 57-bus system. Accuracy values of the SVE and perceptron increase while Precision values of the {$k$-NN} and SLR increase as $\frac{\kappa}{N}$ increases. Both Accuracy and Precision values of the SVM increase and phase transitions occur.}
\label{fig:1}
\end{figure}

The performance of different algorithms is compared for the IEEE 57-bus system in Fig.~\ref{fig:1}.
Accuracy values of the SVE and perceptron increase as $\frac{\kappa}{N}$ increases in Fig.~\ref{fig:1}.a and Fig.~\ref{fig:1}.b. Additionally, Recall values of the SVE increase linearly as $\frac{\kappa}{N}$ increases. Precision values of the perceptron are high and do not decrease, and Accuracy and Recall values increase, since \textit{fn} values decrease and \textit{tn} values increase. In Fig.~\ref{fig:1}.c, a phase transition around $\kappa^*= N-D+1$, is observed for the performance of the SVM. Since the distance between measurement vectors of attacked and secure variables increases as $\frac{\kappa}{N}$ increases following \eqref{eq:dm}, we observe that the Accuracy, Precision and Recall values of the $k$-NN increase in Fig. \ref{fig:1}.d. Accuracy and Recall values of the $k$-NN and SLR are above $0.9$ and do not change as $\frac{\kappa}{N}$ increases in Fig.~\ref{fig:1}.e.

\begin{figure}[t]
\centering
\subfloat[Performance values for Class-1.]{\includegraphics[width=1.7in, height=1.5in]{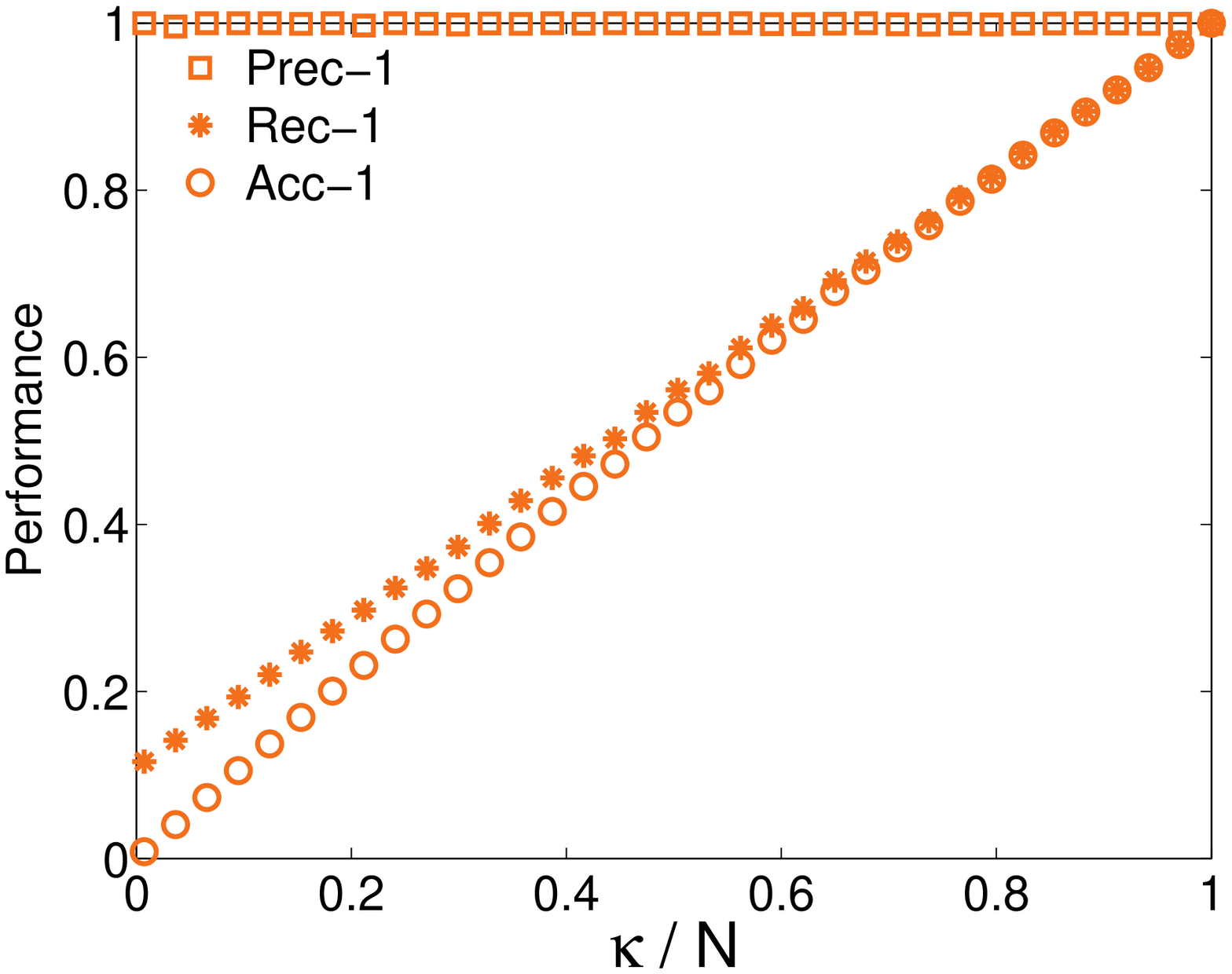}}
\subfloat[Performance values for Class-2.]{\includegraphics[width=1.7in, height=1.5in]{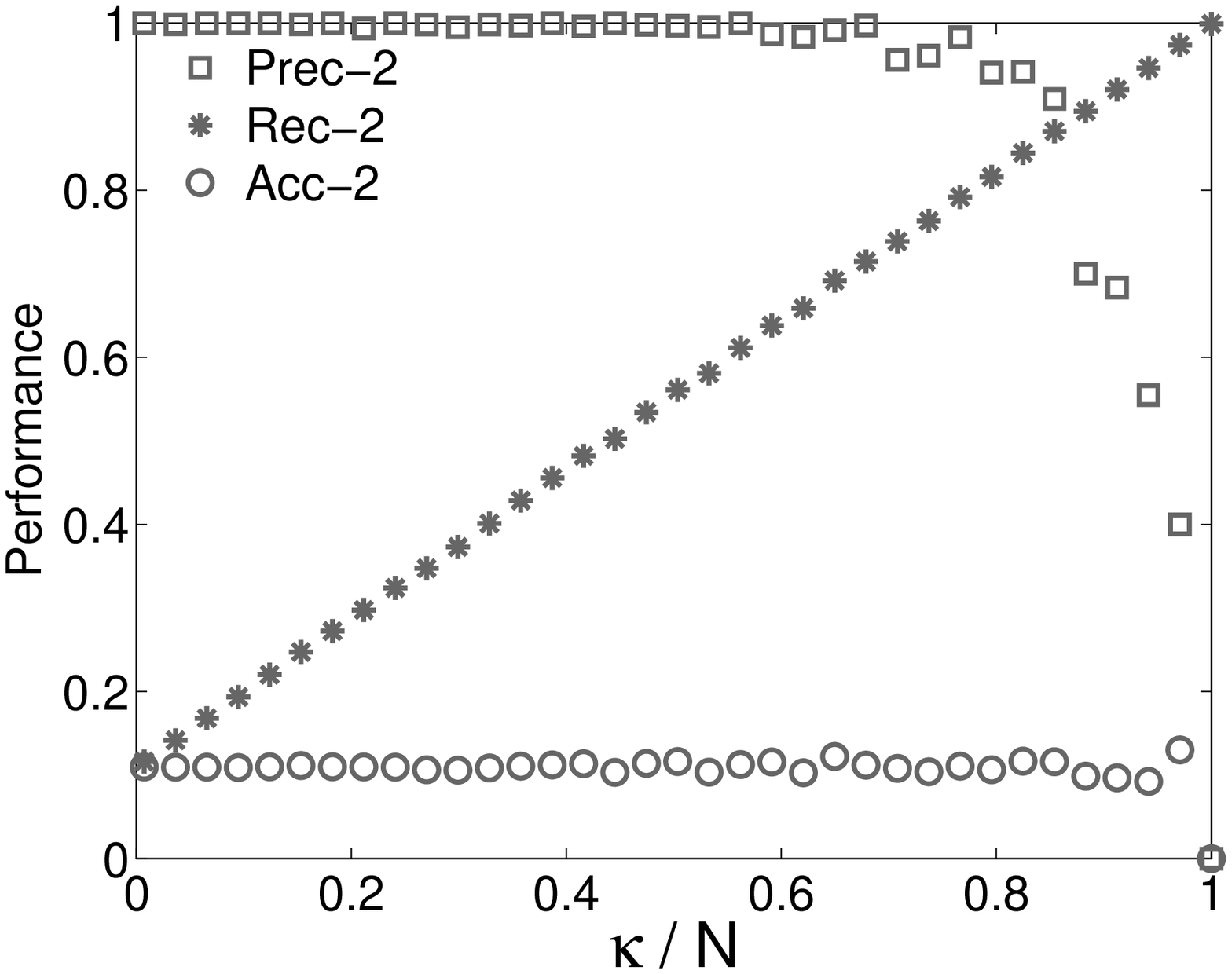}}
\caption{Experiments using the SVE for the IEEE 57-bus test system. Note that $fp$ values increase as $\frac{\kappa}{N}$ increases.}
\label{fig:2}
\end{figure}

The class-based performance values of the algorithms are measured using class-wise  performance indices, where Class-1 and Class-2 denotes the class of attacked and secure variables, respectively. The class-wise performance indices are defined as follows:
\begin{eqnarray}
 {\rm Class-1: } & Prec-1=\frac{tp}{tp+fp}, & Rec-1=\frac{tp}{tp+fn}, \\
 {\rm Class-2: } & Prec-2=\frac{tn}{tn+fn}, & Rec-2=\frac{tn}{fp+tn}.
\end{eqnarray}


\begin{figure}[t]
\centering
\subfloat[Results for the IEEE 57-bus.]{\includegraphics[width=1.75in, height=1.5in]{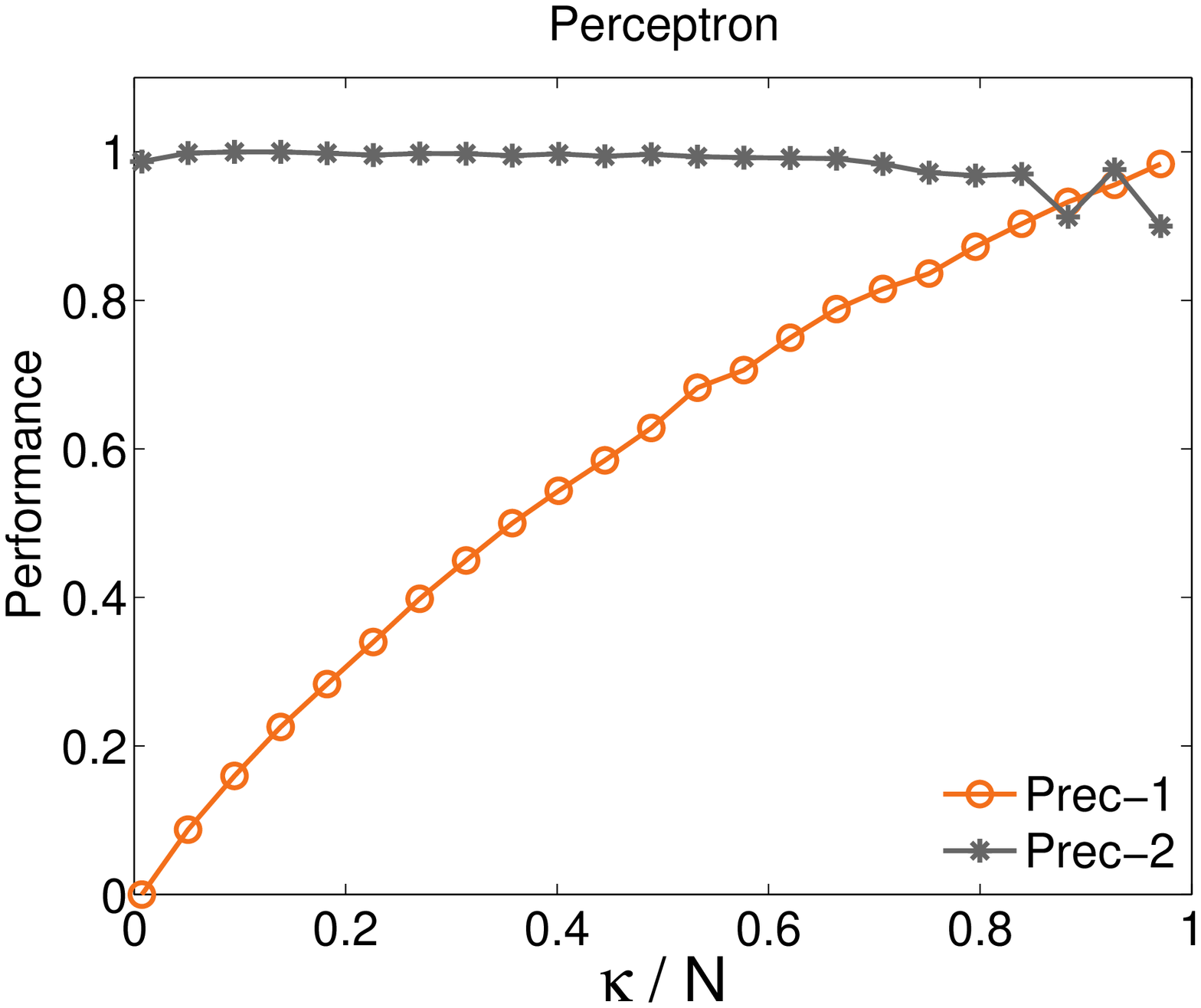}} 
\subfloat[Results for the IEEE 57-bus.]{\includegraphics[width=1.75in, height=1.5in]{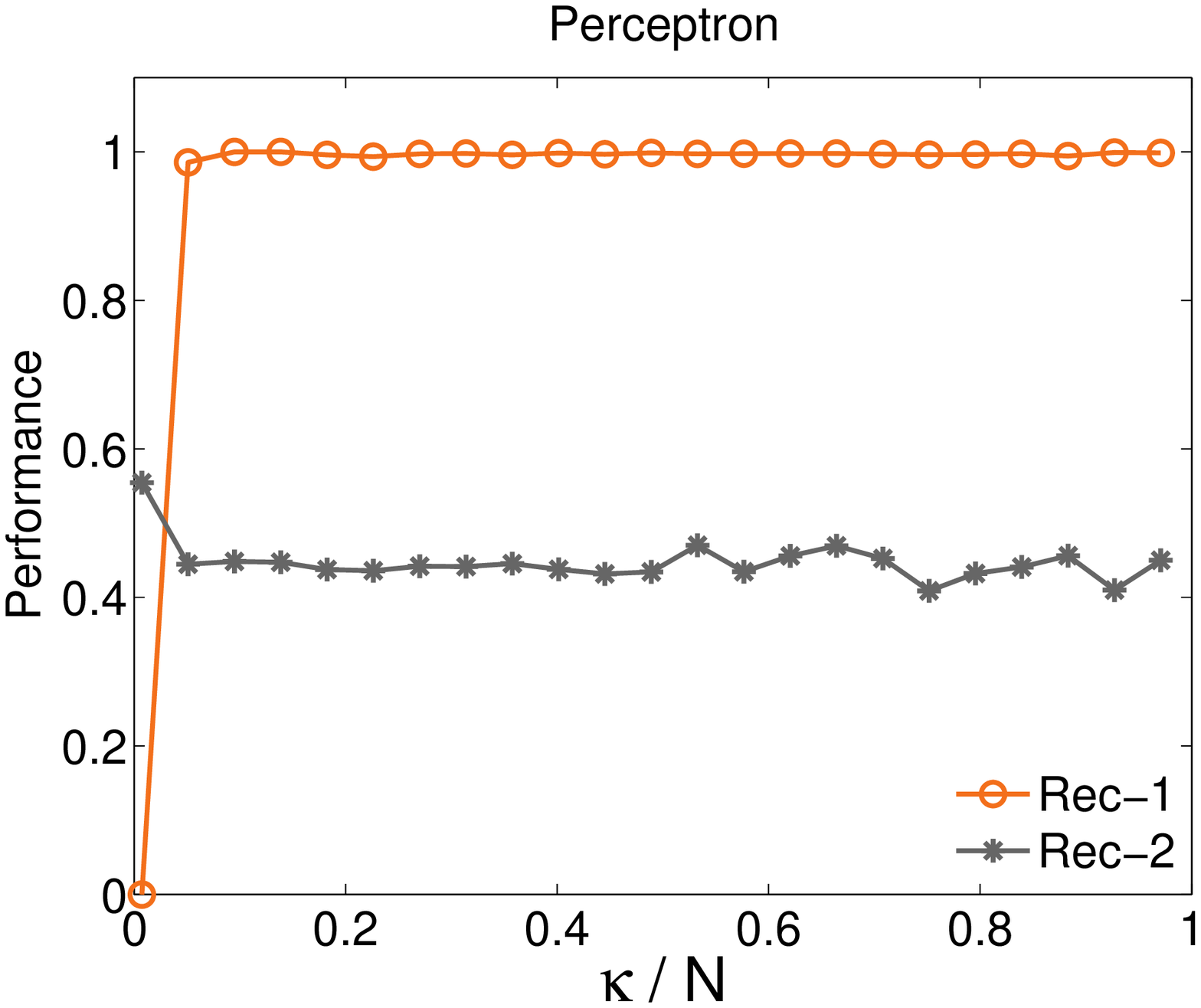}} \\
\subfloat[Results for the IEEE 118-bus.]{\includegraphics[width=1.75in, height=1.5in]{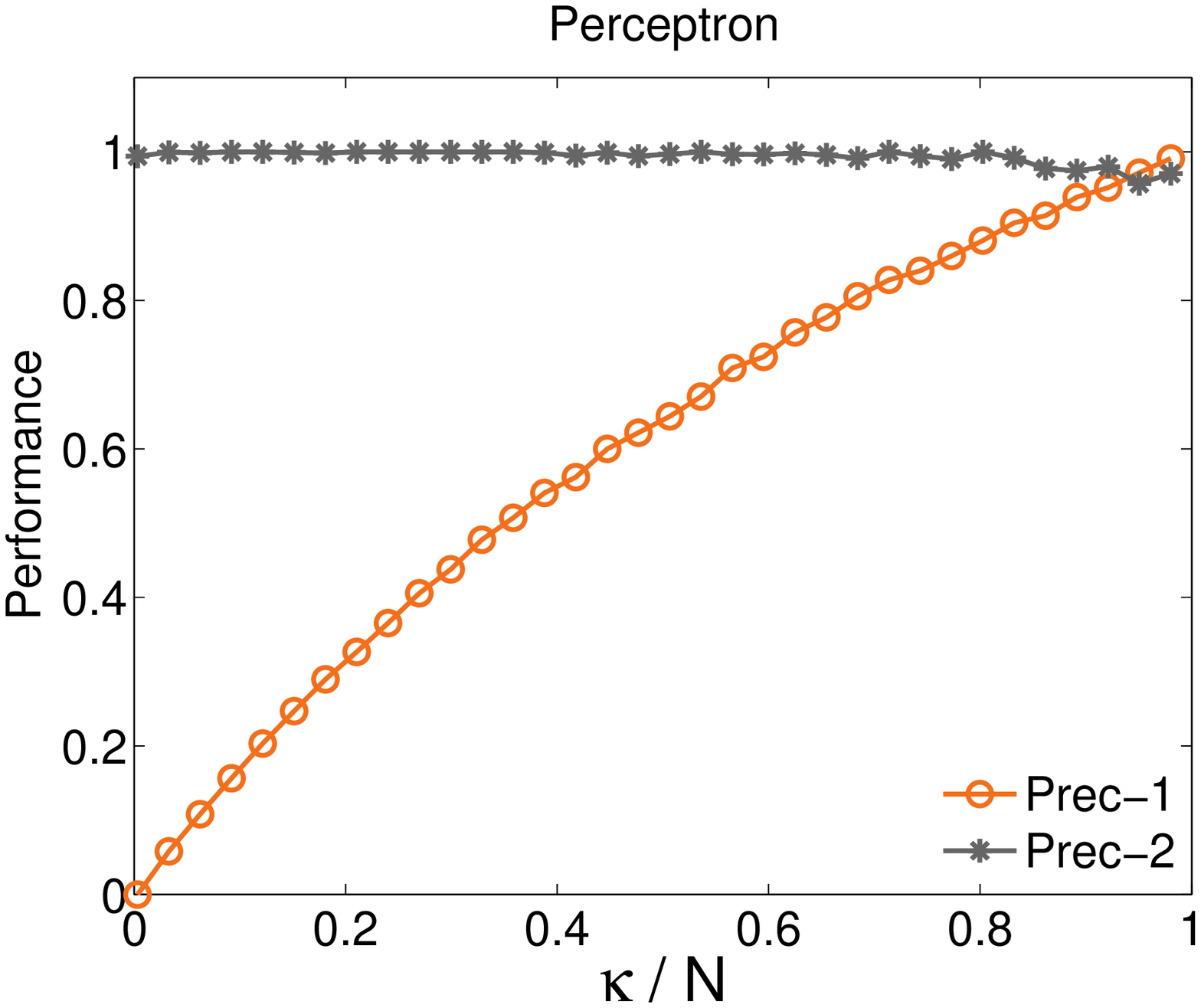}} 
\subfloat[Results for the IEEE 118-bus.]{\includegraphics[width=1.75in, height=1.5in]{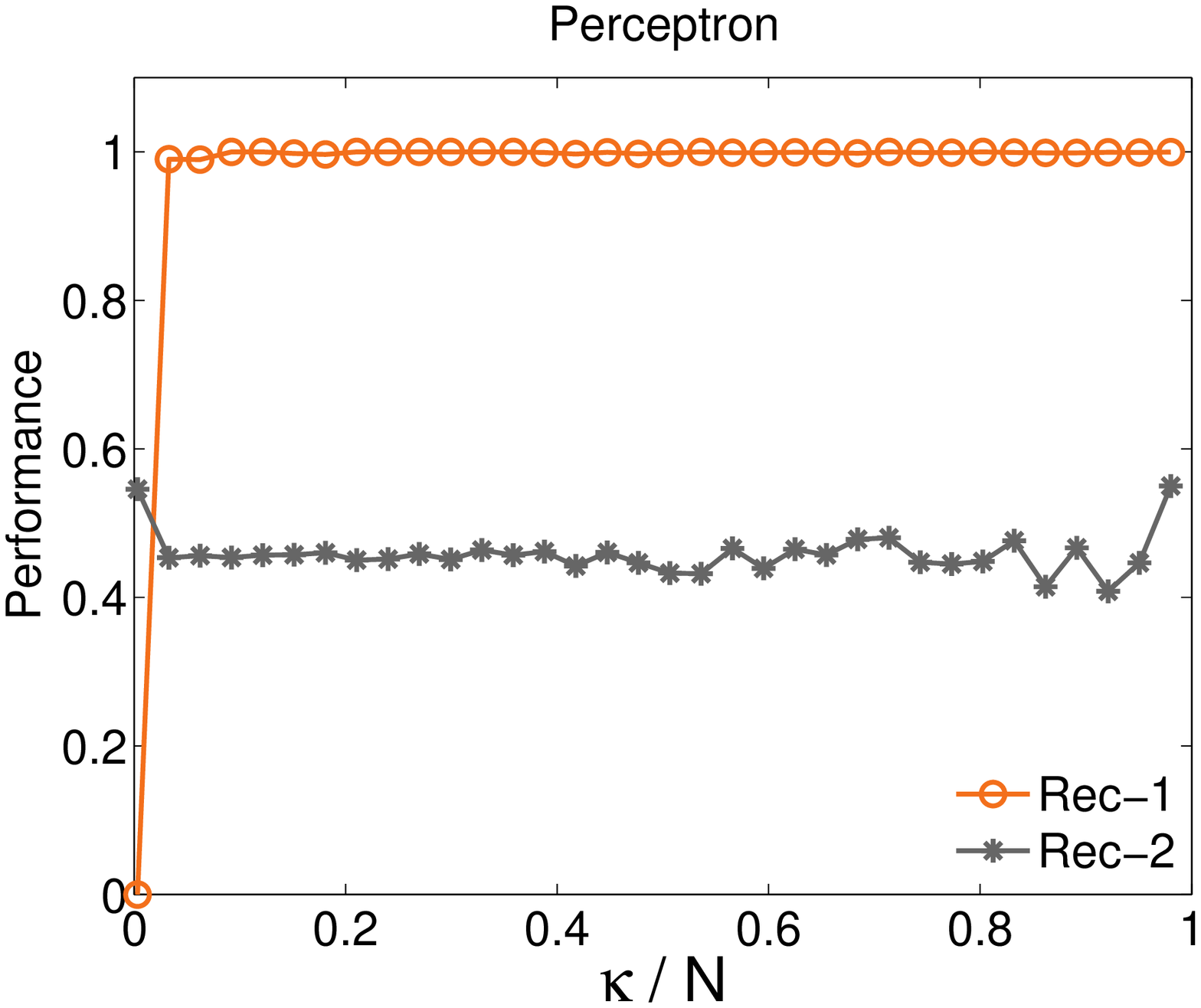}} 
\caption{Performance analysis of the perceptron.}
\label{fig:per}
\end{figure}

In Fig.~\ref{fig:2}.a, we observe that the Precision, Recall and Accuracy values of the SVE increase as $\frac{\kappa}{N}$ increases for {Class-1}. Note that the first value of Acc-1 is observed at $0.008$. In Fig.~\ref{fig:2}.b, Precision values for Class-2 decrease with the percentage of attacked variables, i.e. the number of secure variables that are incorrectly classified by the SVE increases as the number of attacked variables increases. Although the SVE may correctly detect the attacked variables as $\frac{\kappa}{N}$ increases, the secure variables are incorrectly labelled as attacked variables, and therefore, the SVE gives more false alarms than the other algorithms. 

Performance values for the perceptron are given in Fig. \ref{fig:per}. We observe that Precision values for Class-1 increase and Recall values do not change drastically for both of the classes as $\frac{\kappa}{N}$ increases. Moreover, we do not observe any performance increase for the Recall values of the secure class in the perceptron.

\begin{figure}[h!]
\centering
\subfloat[Prec. values for the IEEE 57-bus.]{\includegraphics[width=1.75in, height=1.5in]{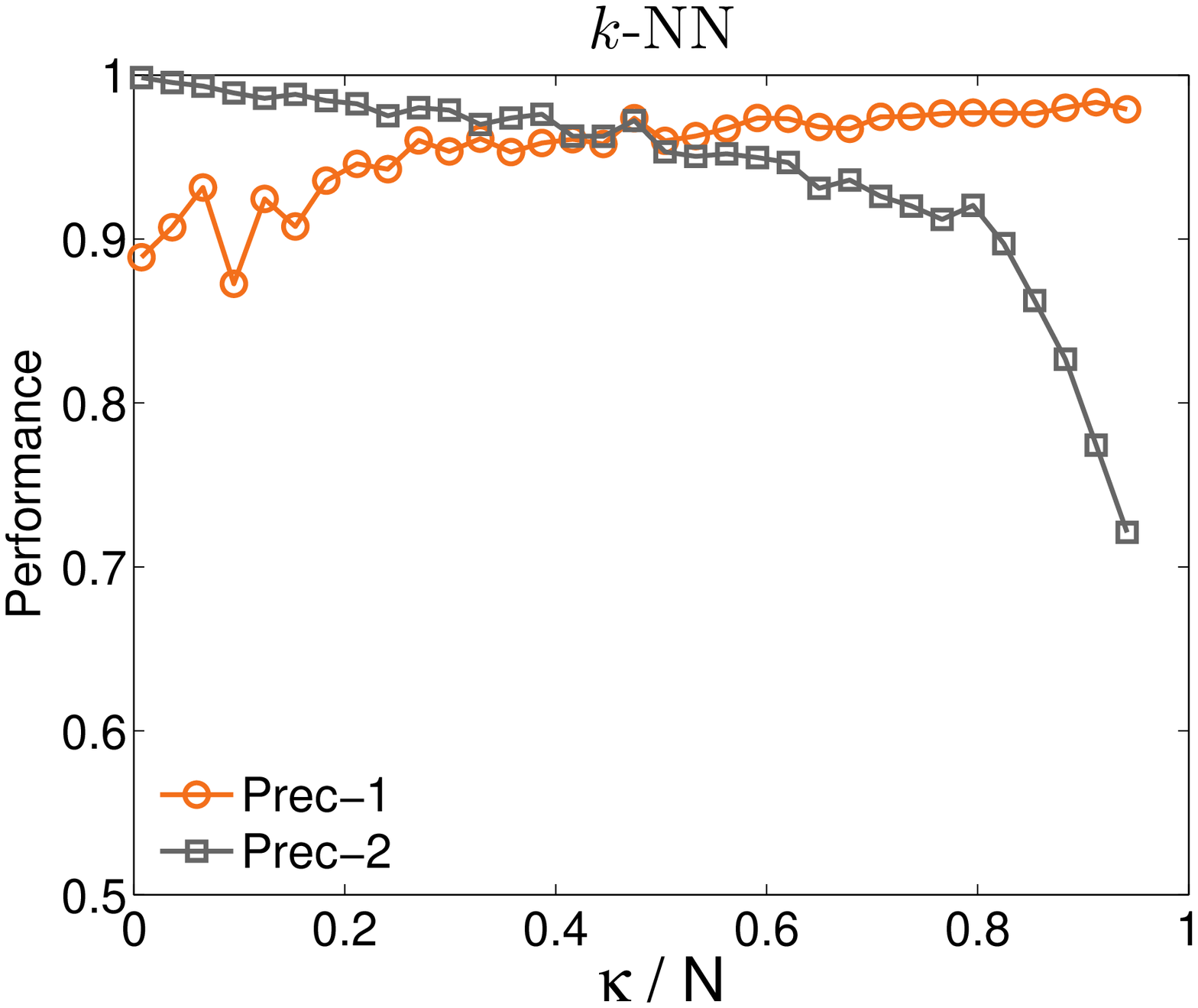}} 
\subfloat[Rec. values for the IEEE 57-bus.]{\includegraphics[width=1.75in, height=1.5in]{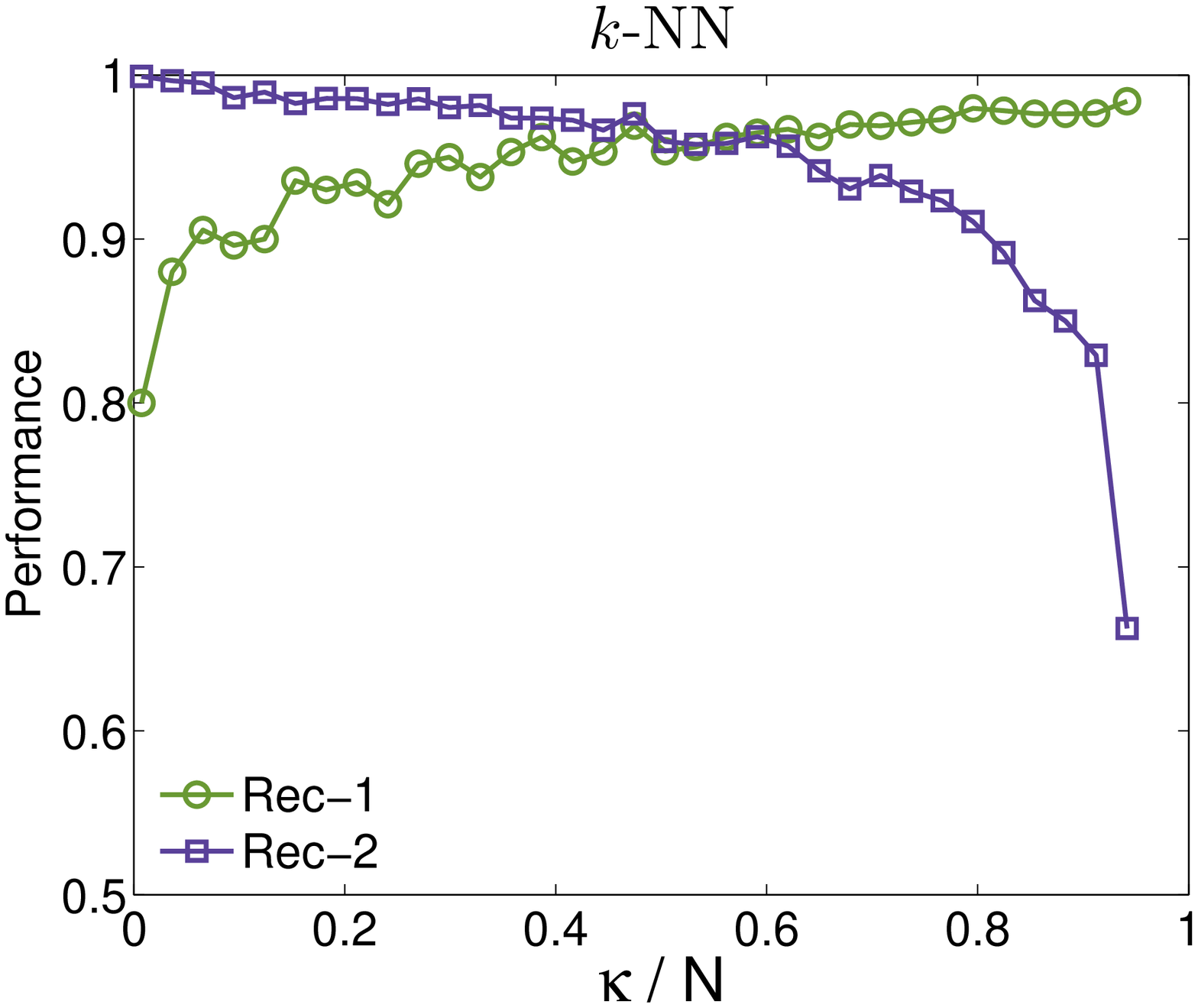}} \\
\subfloat[Prec. values for the IEEE 118-bus.]{\includegraphics[width=1.75in, height=1.5in]{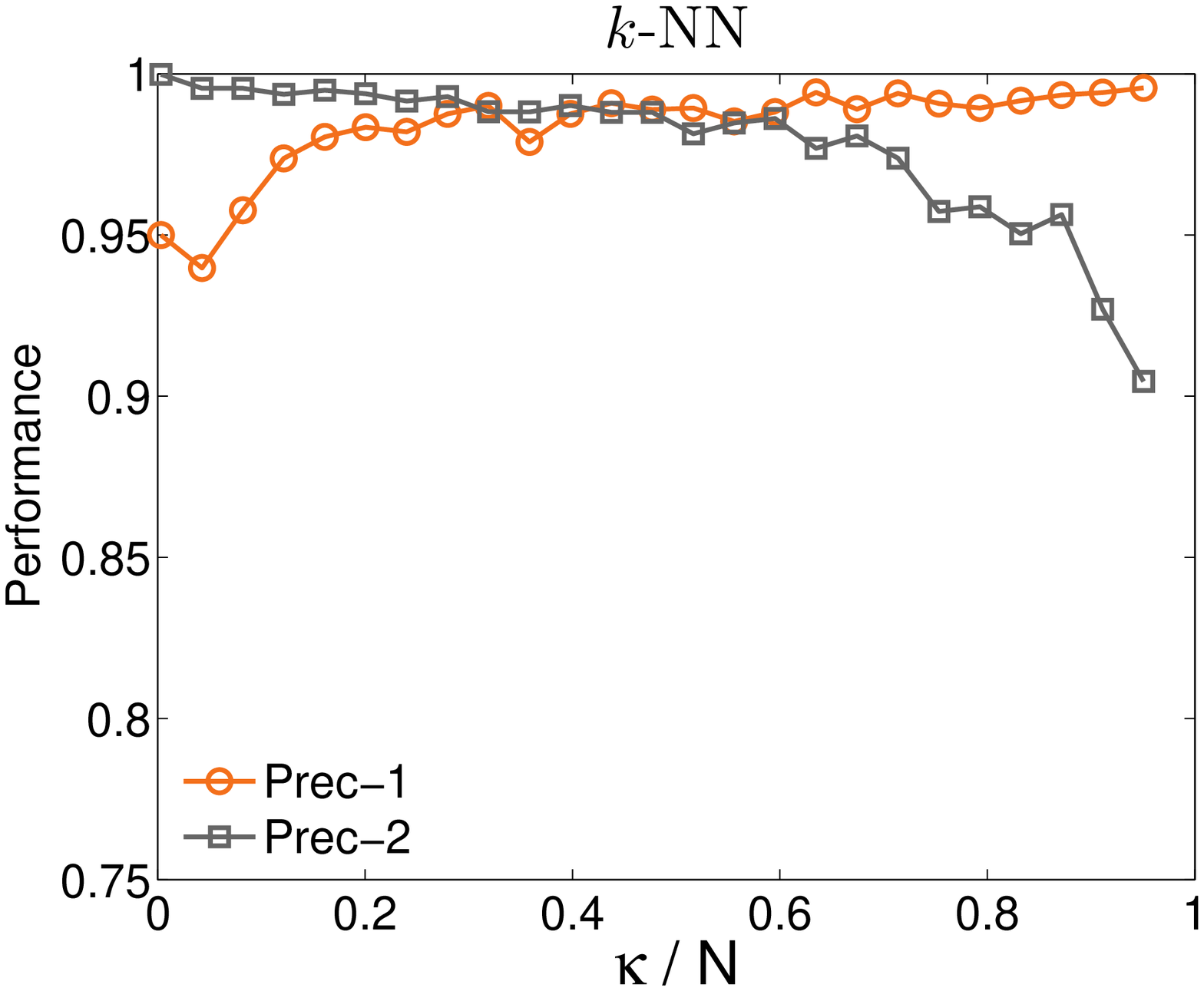}} 
\subfloat[Rec. values for the IEEE 118-bus.]{\includegraphics[width=1.75in, height=1.5in]{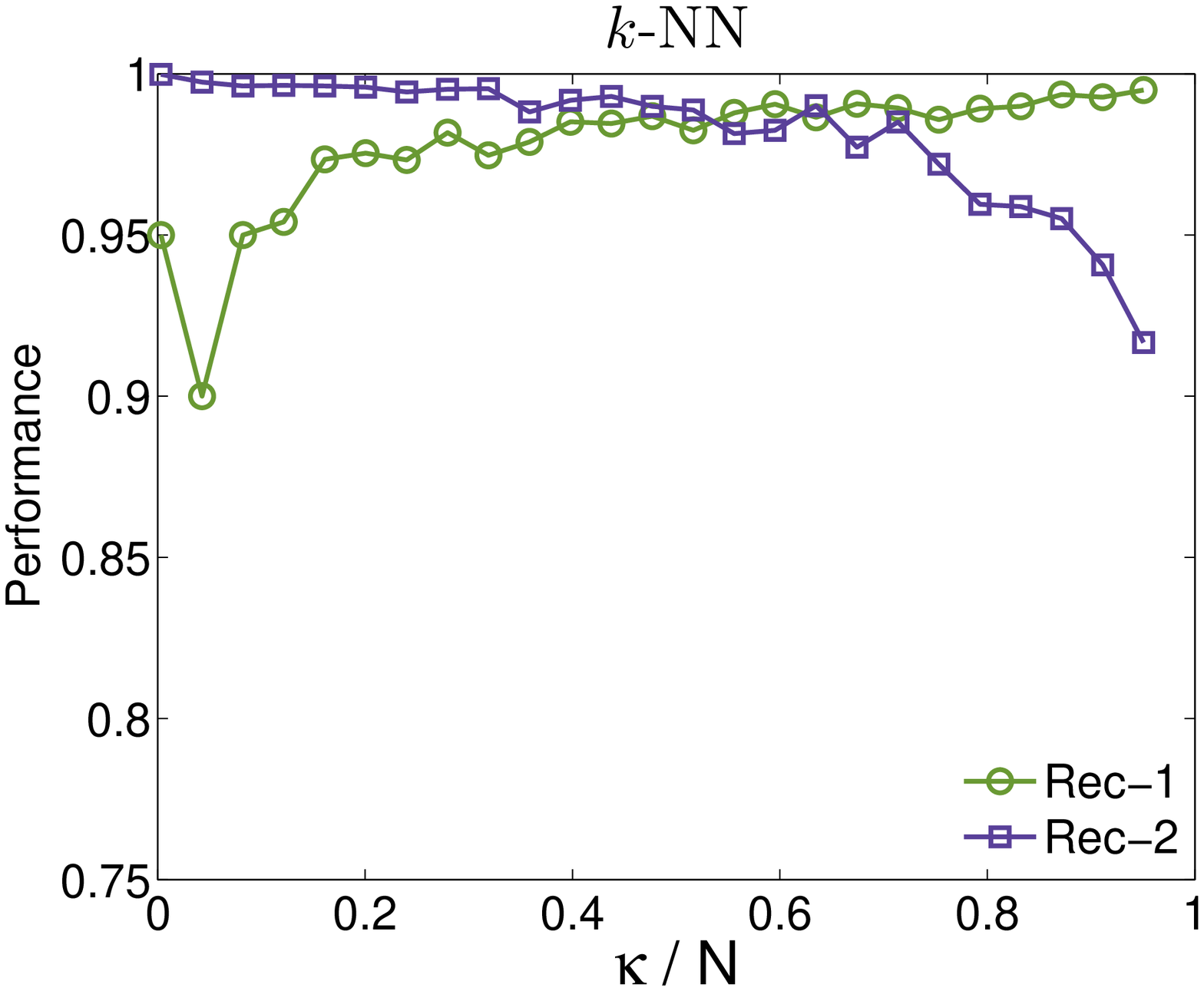}} 
\caption{Since the $k$-NN is sensitive to class-balance and sparsity of the data, performance values for Class-1 increase and the values for Class-2 decrease as $\frac{\kappa}{N}$ increases. Note that the performance curves intersect at the critical values $\kappa^*$.}
\label{fig:3}
\end{figure}

In Fig. \ref{fig:3}, the results for $k$-NN are shown. We observe that performance values for Class-1 increase and the values for Class-2 decrease as $\frac{\kappa}{N}$ increases since $k$-NN is sensitive to class-balance and sparsity of the data \cite{devroye}. In addition, classification hypotheses are computed by forming neighborhoods in Euclidean spaces, and the $\ell_2$ norm of vectors of attacked measurements increases as $\frac{\kappa}{N}$ increases in \eqref{eq:dm}; therefore, decision boundaries of the hypotheses are biased towards Class-1.

\begin{figure}[t]
\centering
\subfloat[Results for the IEEE 57-bus.]{\includegraphics[width=1.7in, height=1.5in]{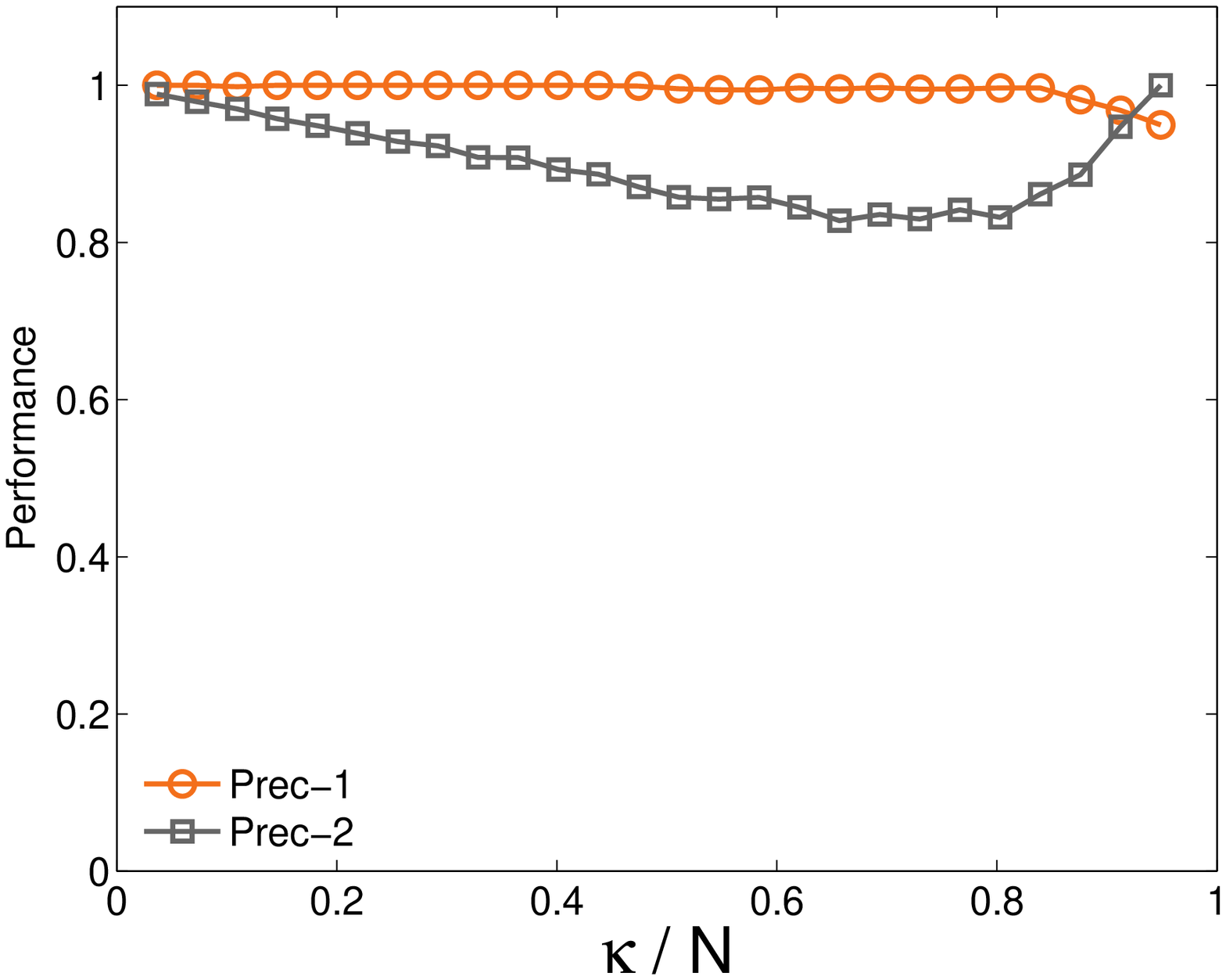}} 
\subfloat[Results for the IEEE 57-bus.]{\includegraphics[width=1.7in, height=1.5in]{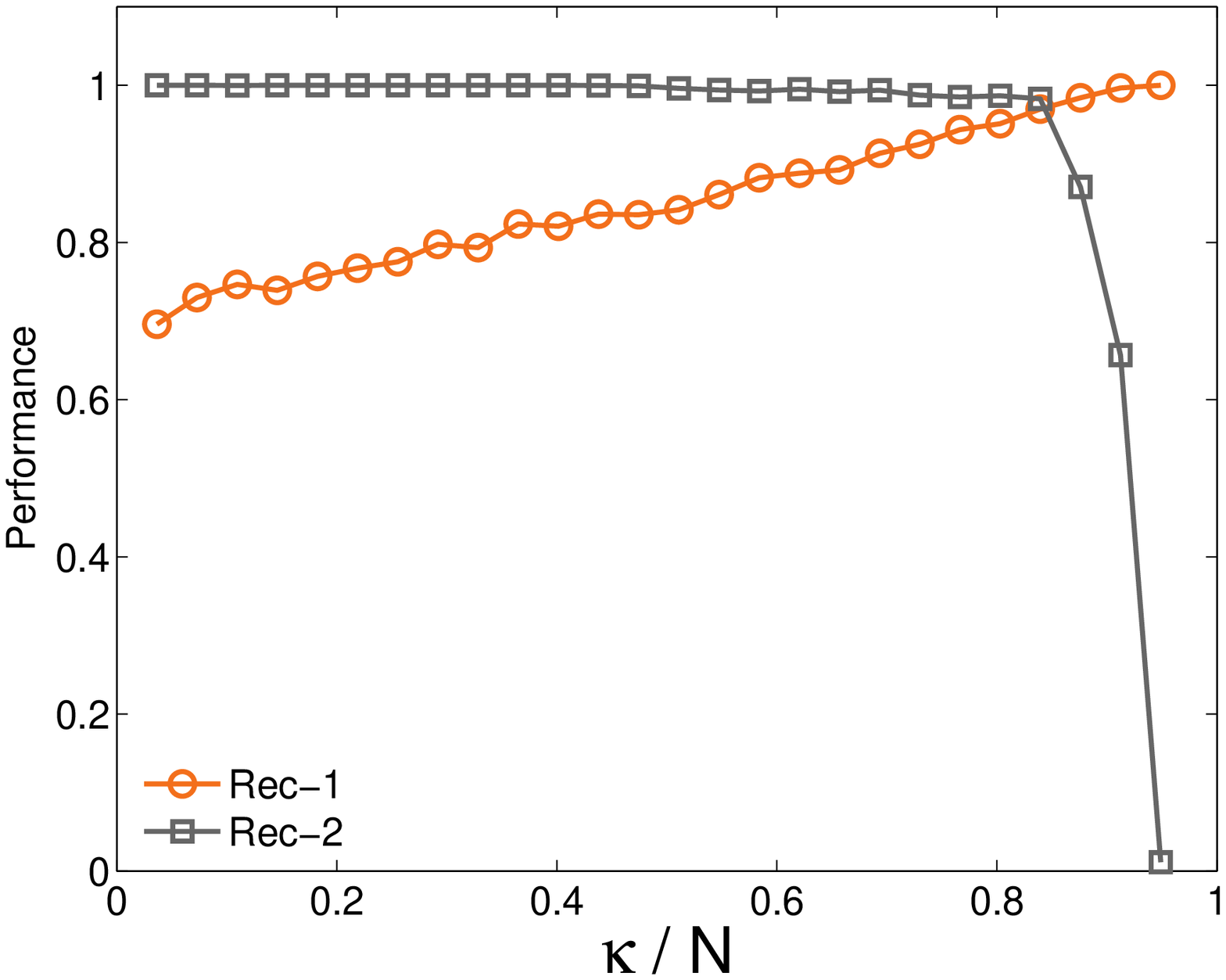}} \\ 
\subfloat[Results for the IEEE 118-bus.]{\includegraphics[width=1.7in, height=1.5in]{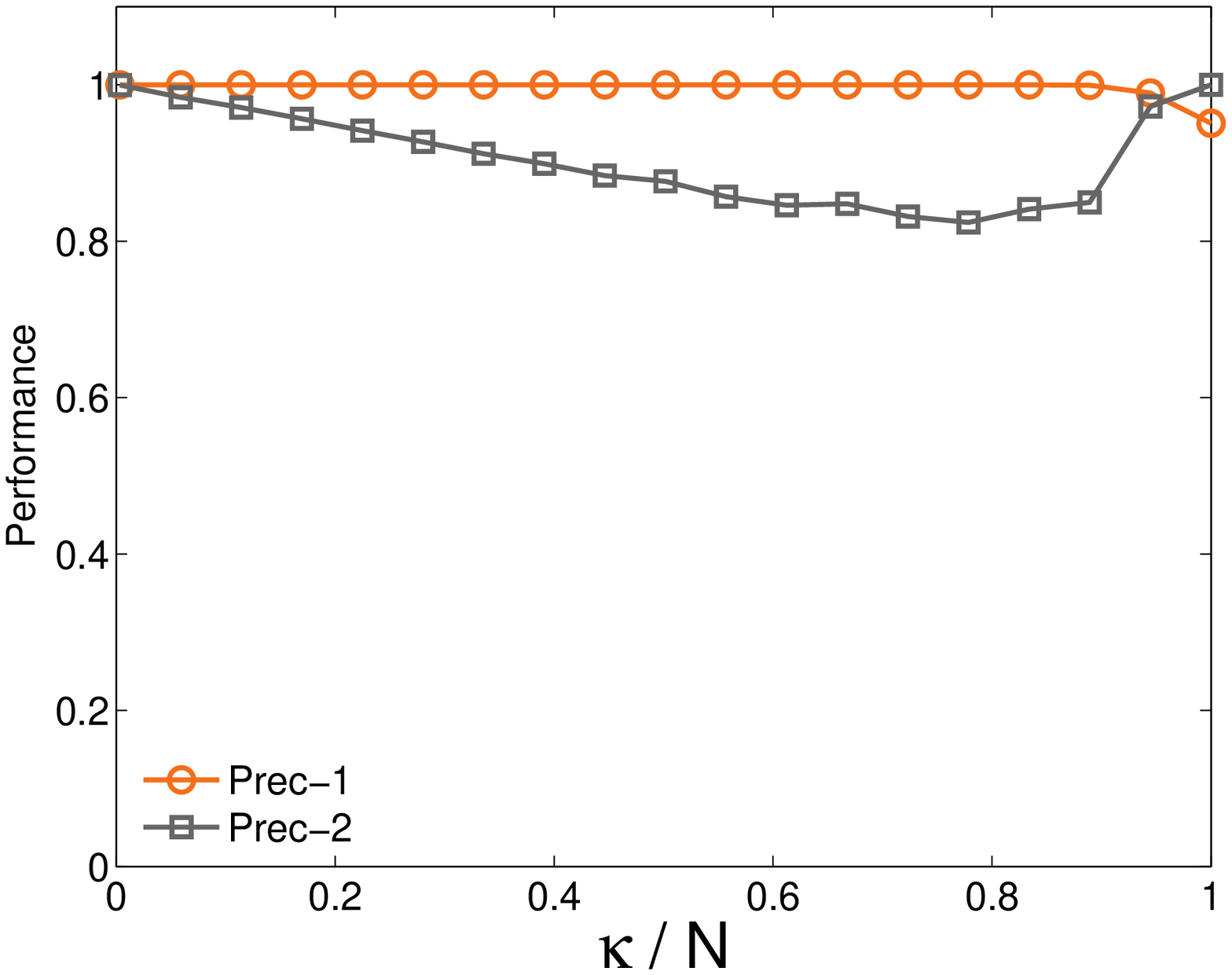}} 
\subfloat[Results for the IEEE 118-bus.]{\includegraphics[width=1.7in, height=1.5in]{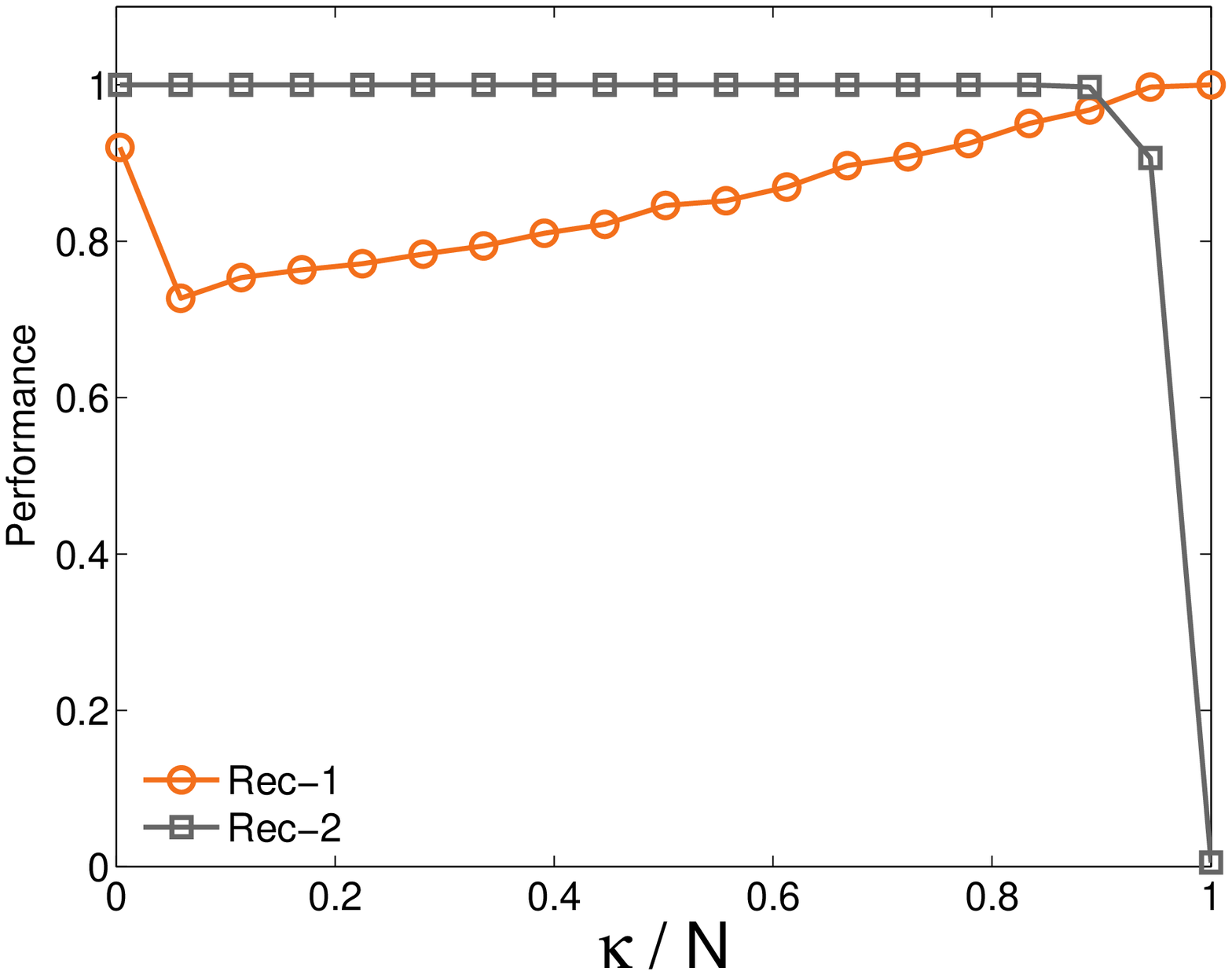}}
\caption{Experiments using the SLR. Note that the SLR can handle the variety in the sparsity of the data as $\frac{\kappa}{N}$ changes.}
\label{fig:5}
\end{figure}

Fig.~\ref{fig:5} depicts the results for the SLR, where the performance values for Class-2 (secure variables) increase as the system size increases. Moreover, we observe that the performance values for Class-2 do not decrease rapidly as $\frac{\kappa}{N}$ increases, compared to the other supervised algorithms. In addition, the performance values for Class-1 are higher than the values of the other algorithms, especially for lower $\frac{\kappa}{N}$ values. The reason is that the SLR can handle the variety in the sparsity of the data as $\frac{\kappa}{N}$ changes. This task is accomplished by controlling and learning the sparsity of the solution in \eqref{eq:dist_log_reg} using the training data in order to learn the sparse structure of the measurements defined in the observation model \eqref{eq:dve1}.

The results of the experiments for the SVM are shown in Fig.~\ref{fig:4}, where a phase transition for the performance values is observed. It is worth noting that the values of $\kappa$ at which the phase transition occurs correspond to the minimum number of measurement variables, $\kappa^*$, that the attacker needs to compromise in order to construct unobservable attacks \cite{s1}. $\kappa^*$ is depicted as a vertical dotted line in Fig.~\ref{fig:4}. For instance, $\kappa^*=10$ and $\frac{\kappa^*}{N} =0.56$ for the IEEE 9-bus test system. The transitions are observed before the critical points when the linear kernel SVM is employed in the experiments for IEEE 57-bus and 118-bus systems. In addition, the phase transitions of performance values occur at the critical points when Gaussian kernels are used.

\subsection{Results for Semi-supervised Learning Algorithms}

\begin{figure*}[t]
\centering
\subfloat[Linear SVM.]{\includegraphics[width=1.8in, height=1.45in]{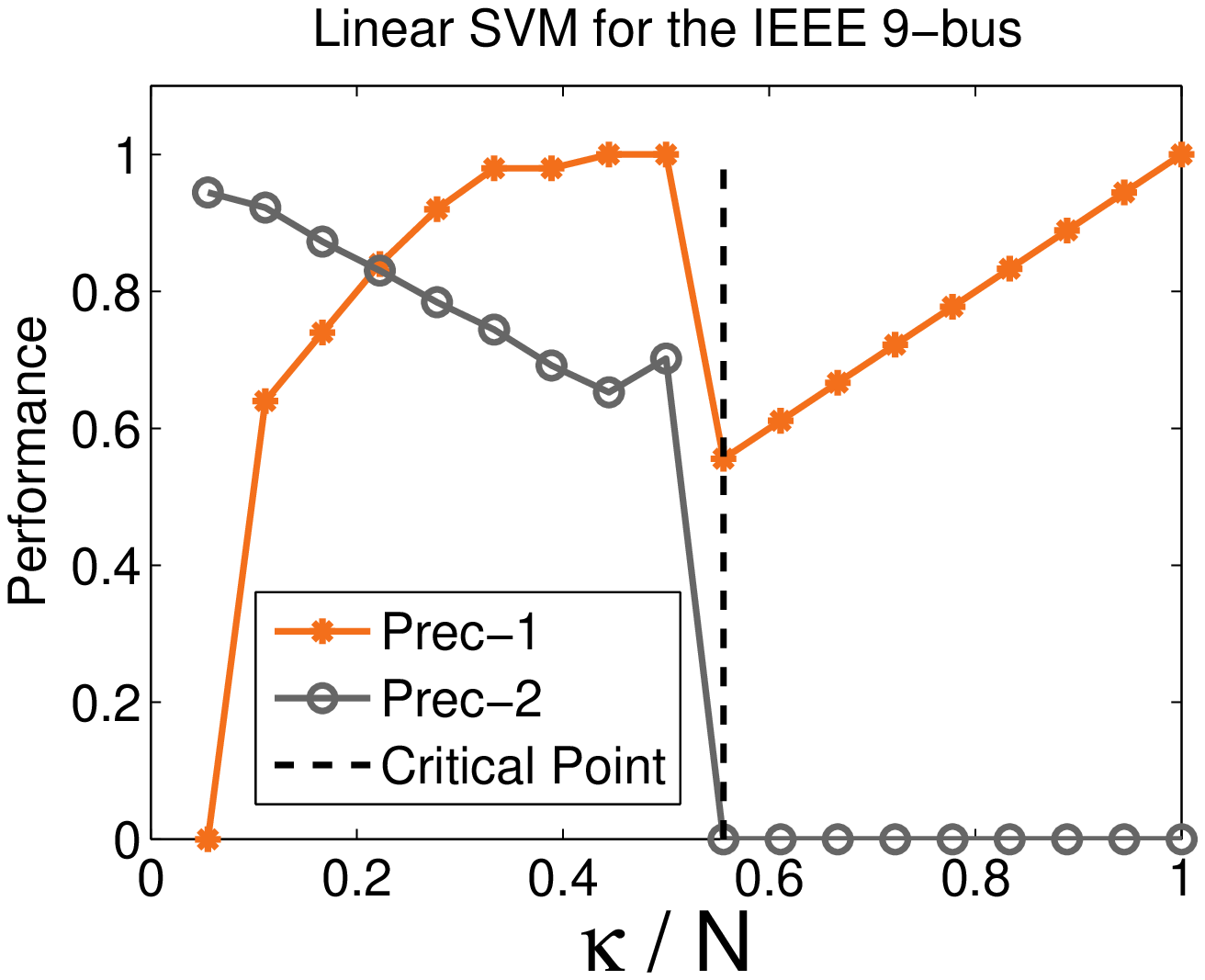}} 
\subfloat[Linear SVM.]{\includegraphics[width=1.8in, height=1.45in]{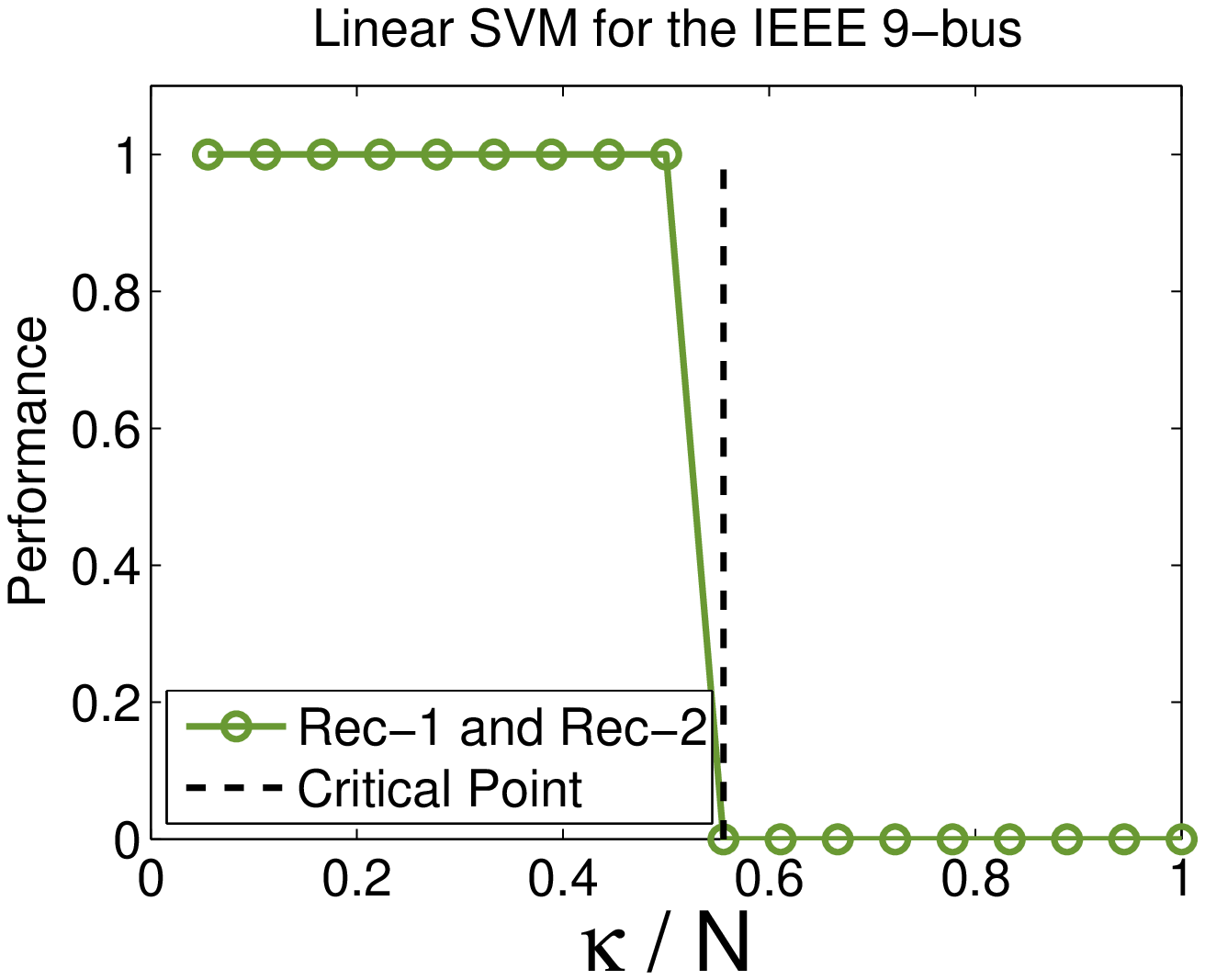}} 
\subfloat[Gaussian SVM.]{\includegraphics[width=1.8in, height=1.45in]{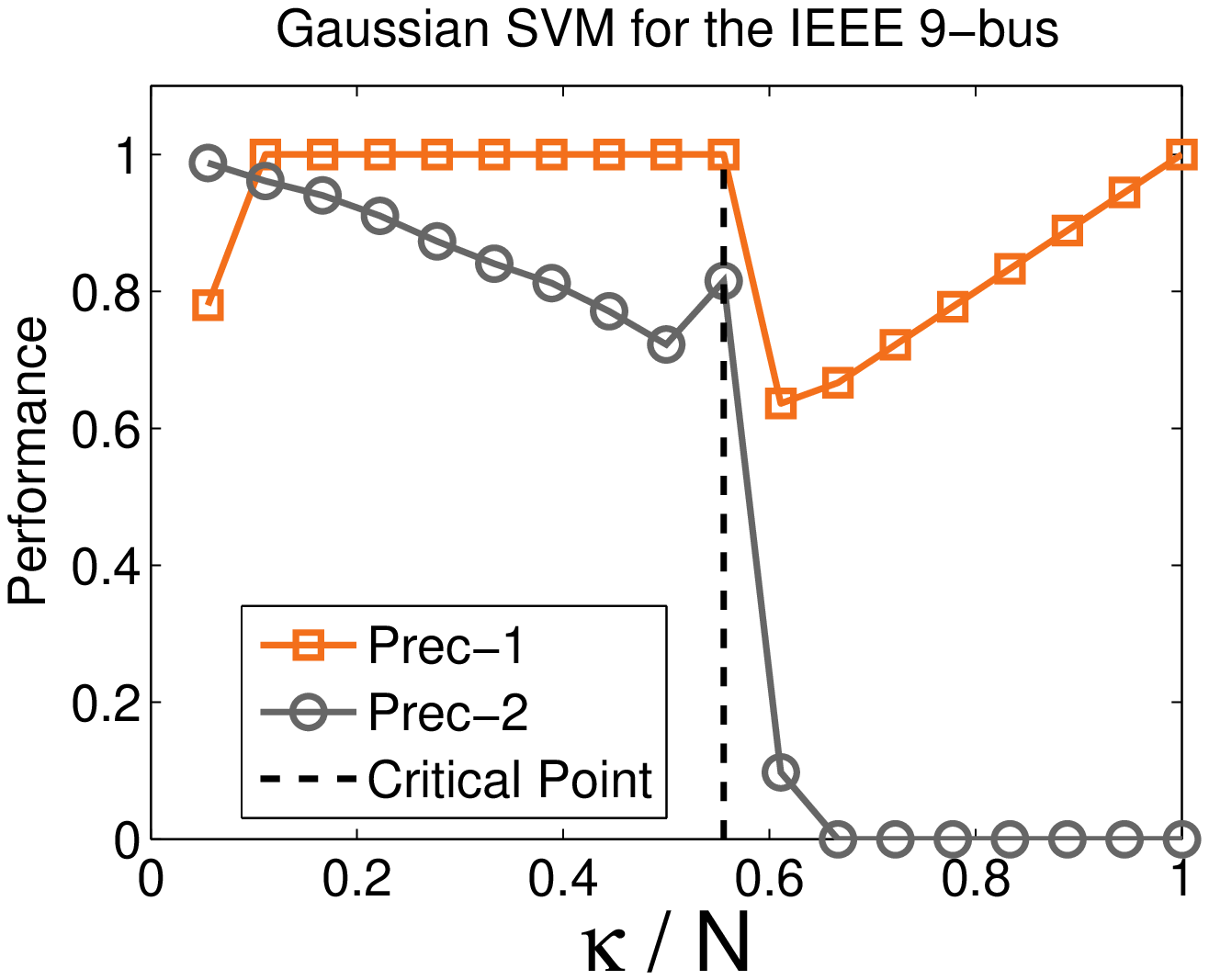}}
\subfloat[Gaussian SVM.]{\includegraphics[width=1.8in, height=1.45in]{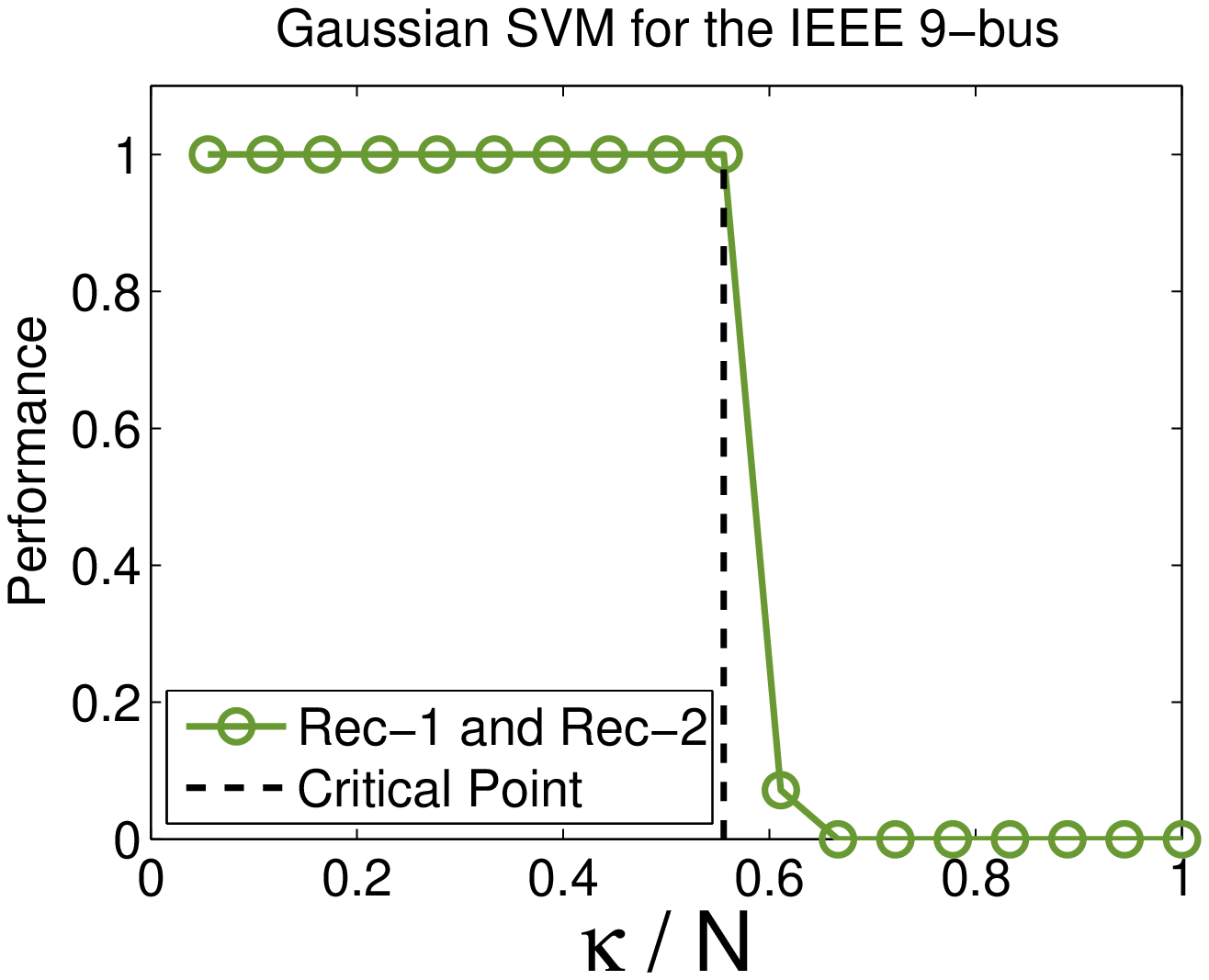}} \\
\subfloat[Linear SVM.]{\includegraphics[width=1.8in, height=1.45in]{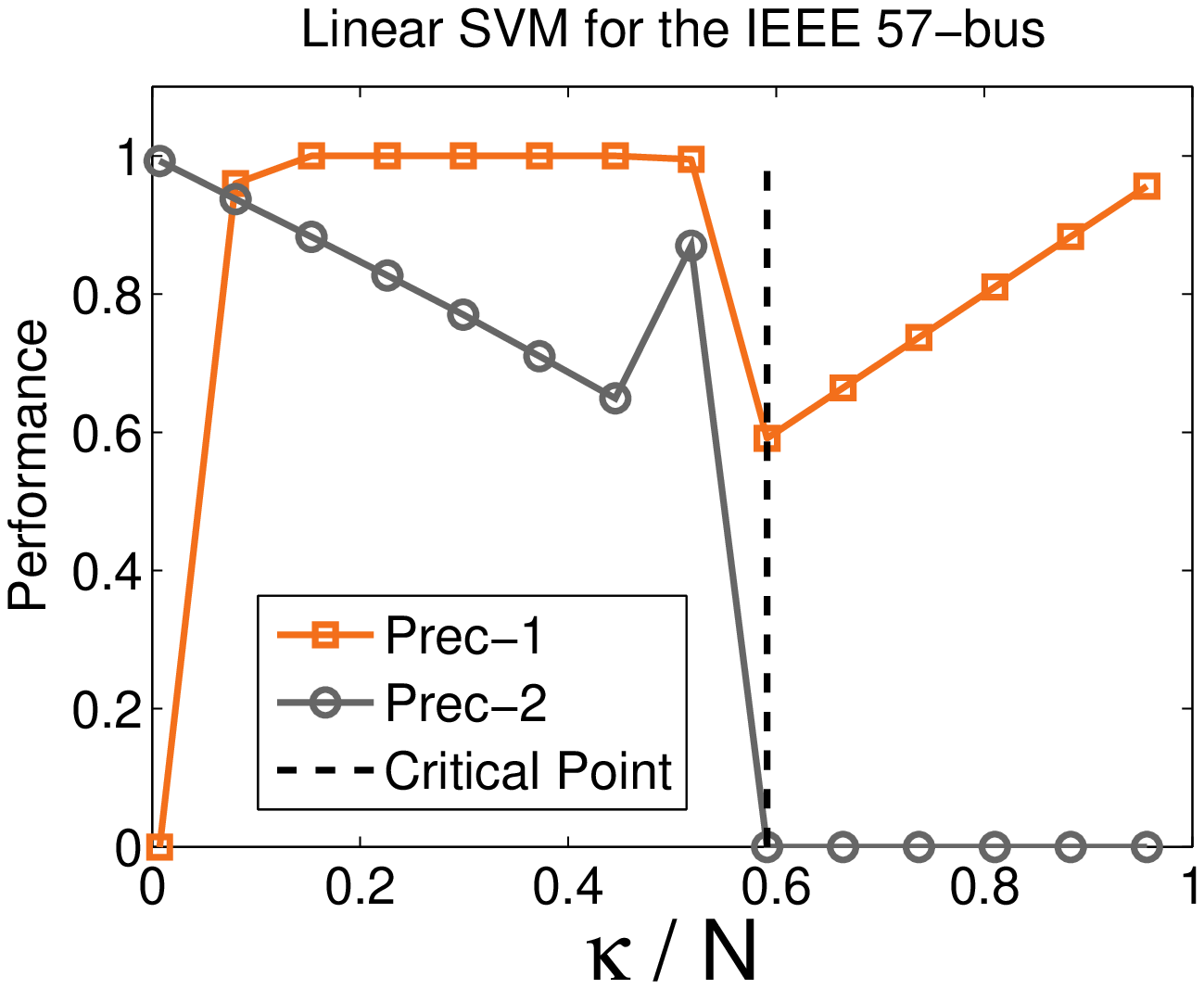}} 
\subfloat[Linear SVM.]{\includegraphics[width=1.8in, height=1.45in]{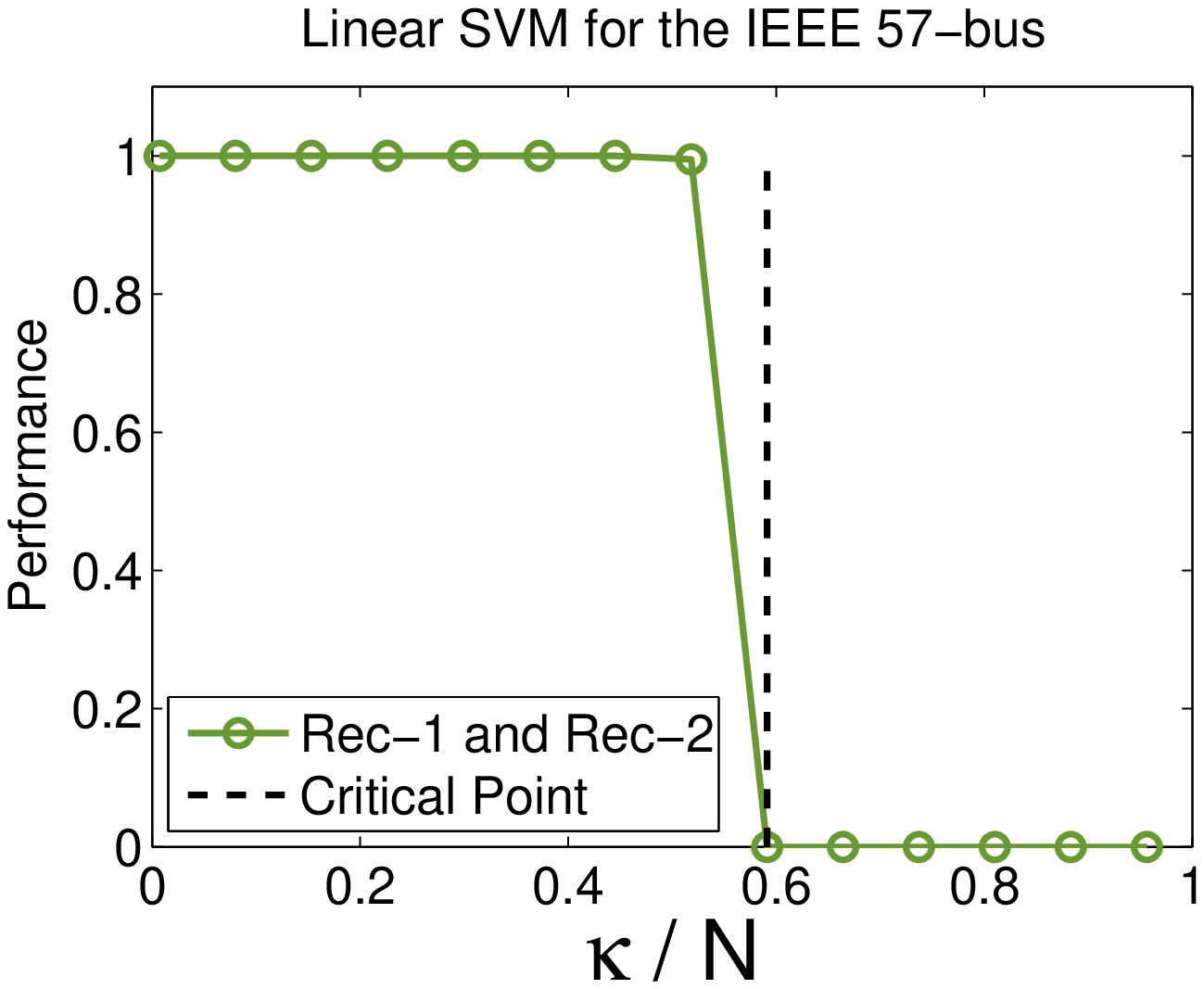}} 
\subfloat[Gaussian SVM.]{\includegraphics[width=1.8in, height=1.45in]{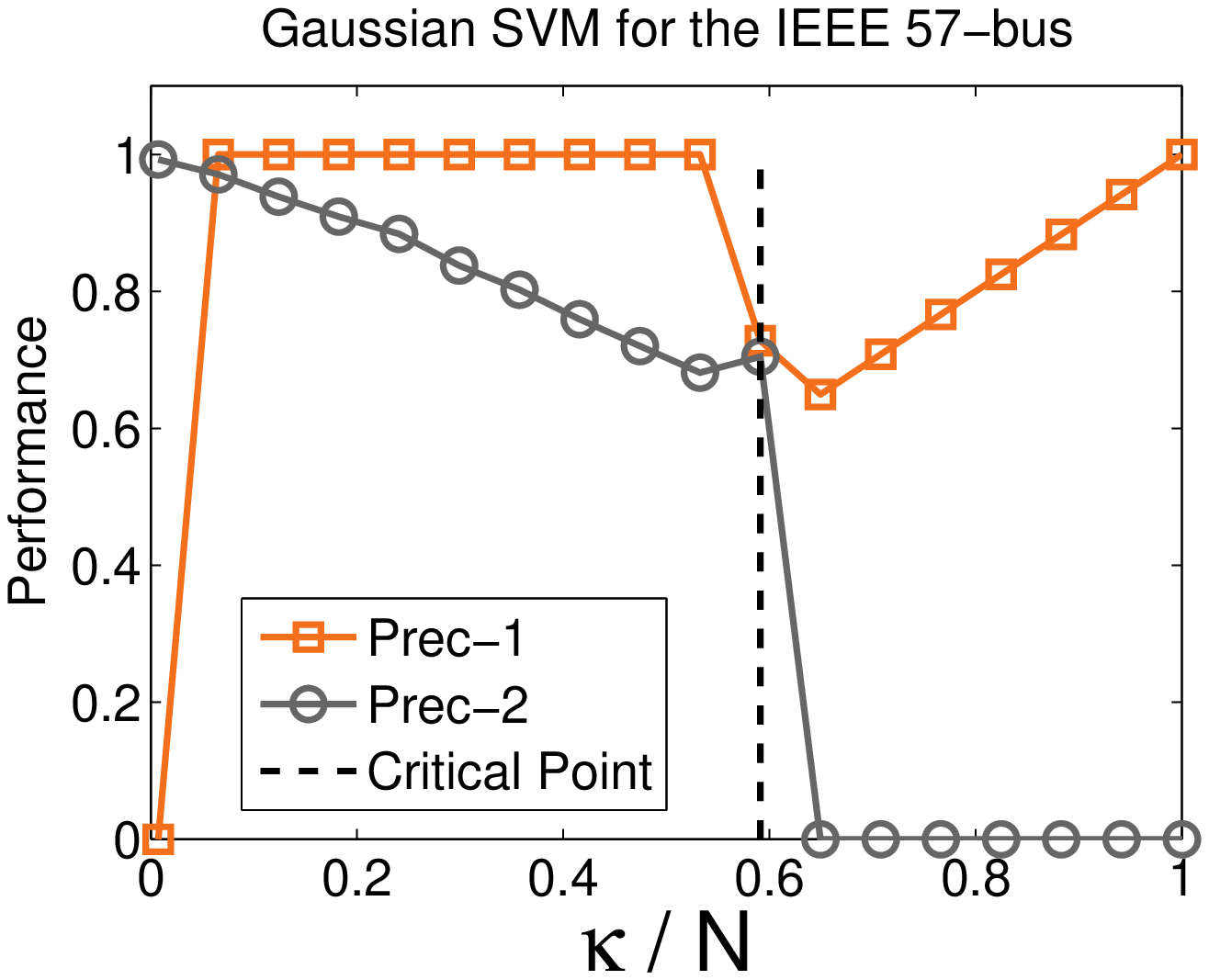}}
\subfloat[Gaussian SVM.]{\includegraphics[width=1.8in, height=1.45in]{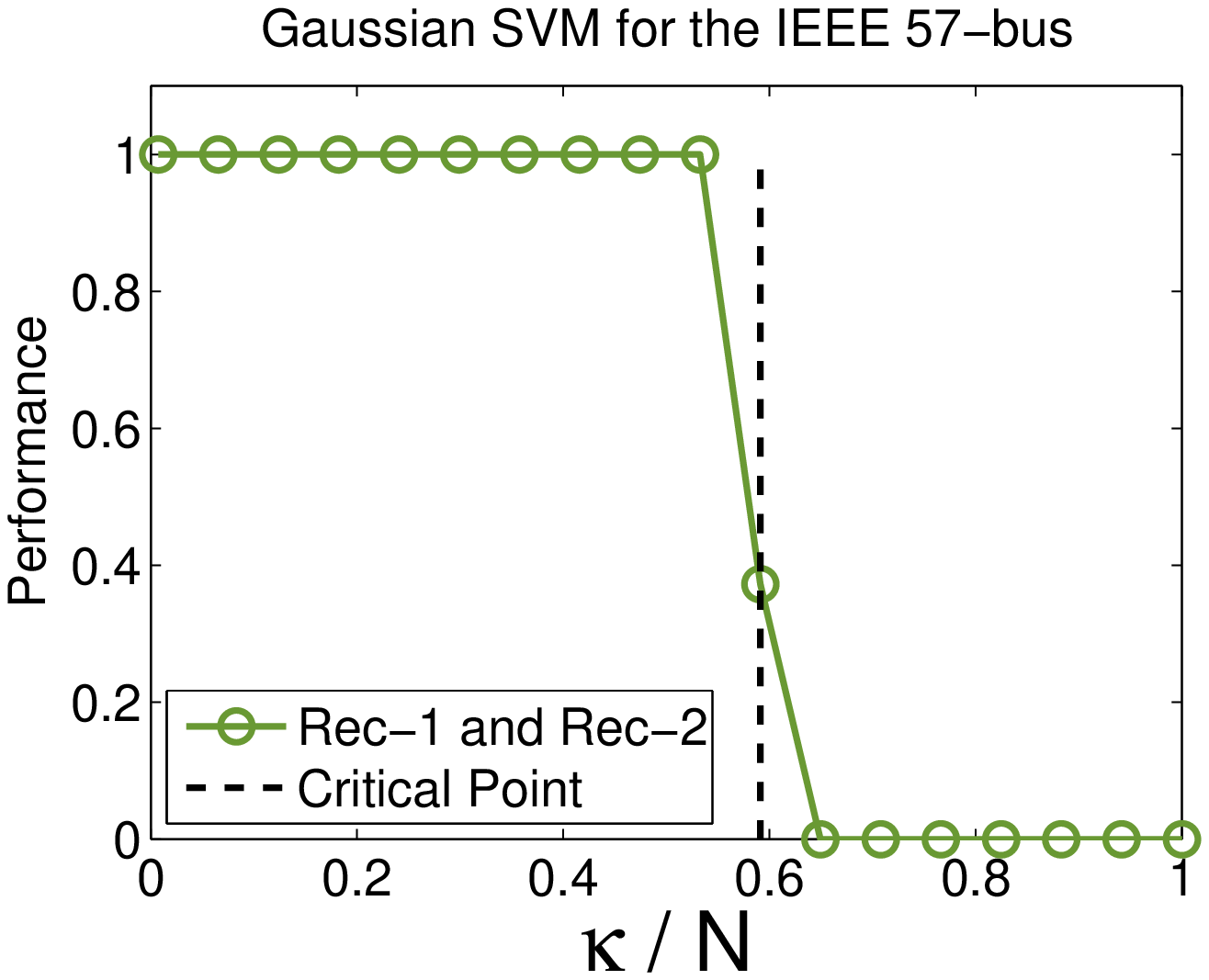}}\\
\subfloat[Linear SVM.]{\includegraphics[width=1.8in, height=1.45in]{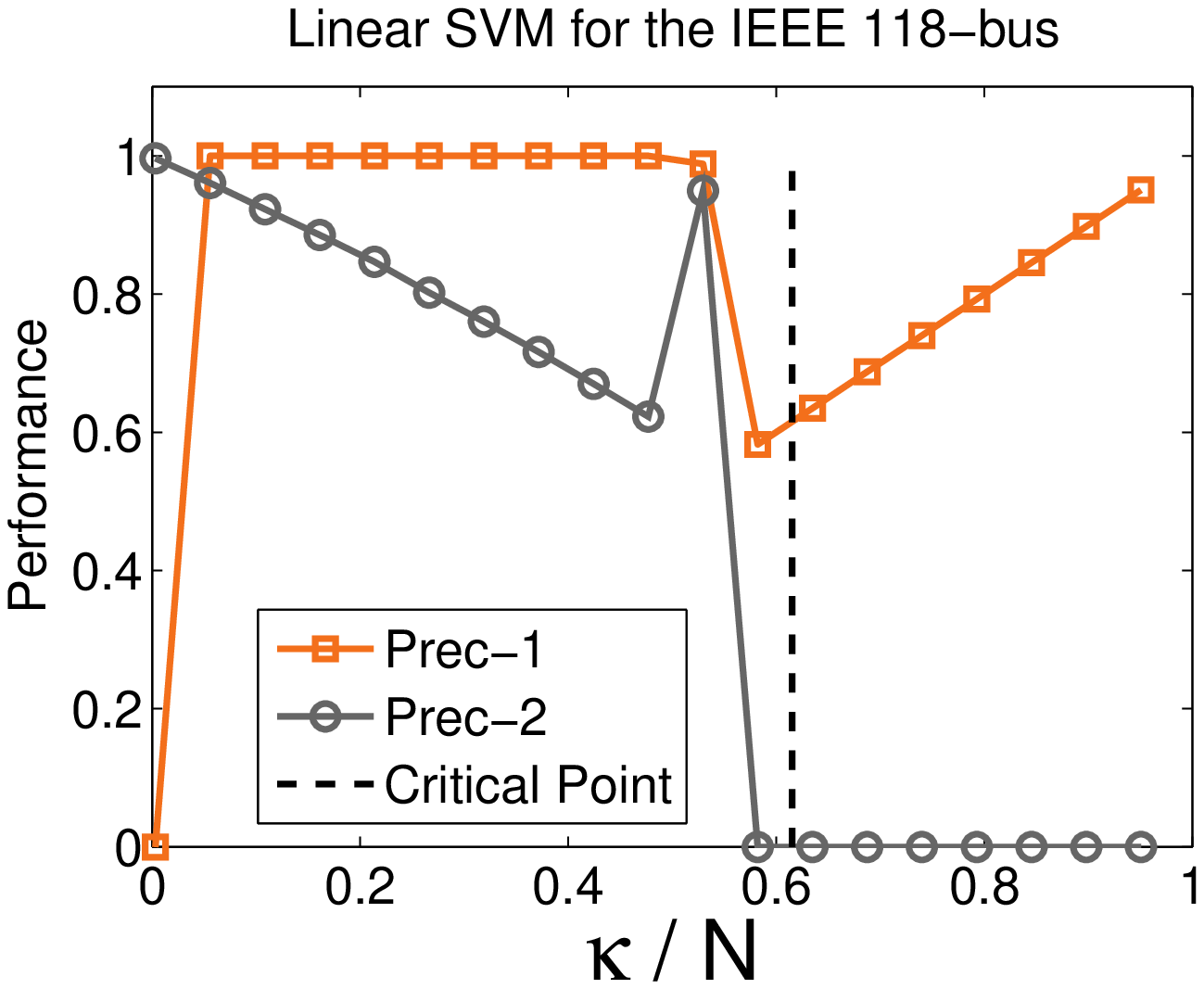}} 
\subfloat[Linear SVM.]{\includegraphics[width=1.8in, height=1.45in]{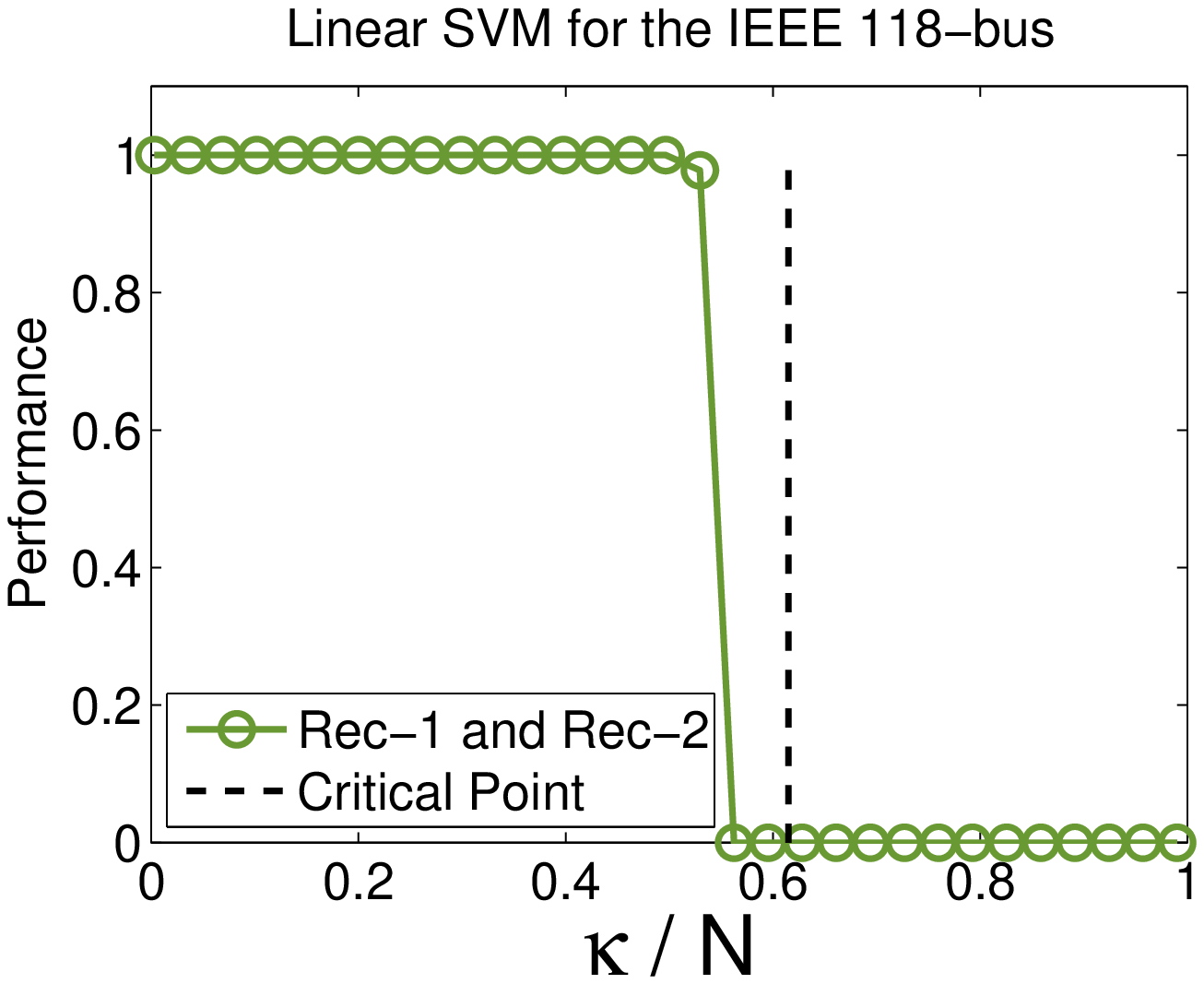}} 
\subfloat[Gaussian SVM.]{\includegraphics[width=1.8in, height=1.45in]{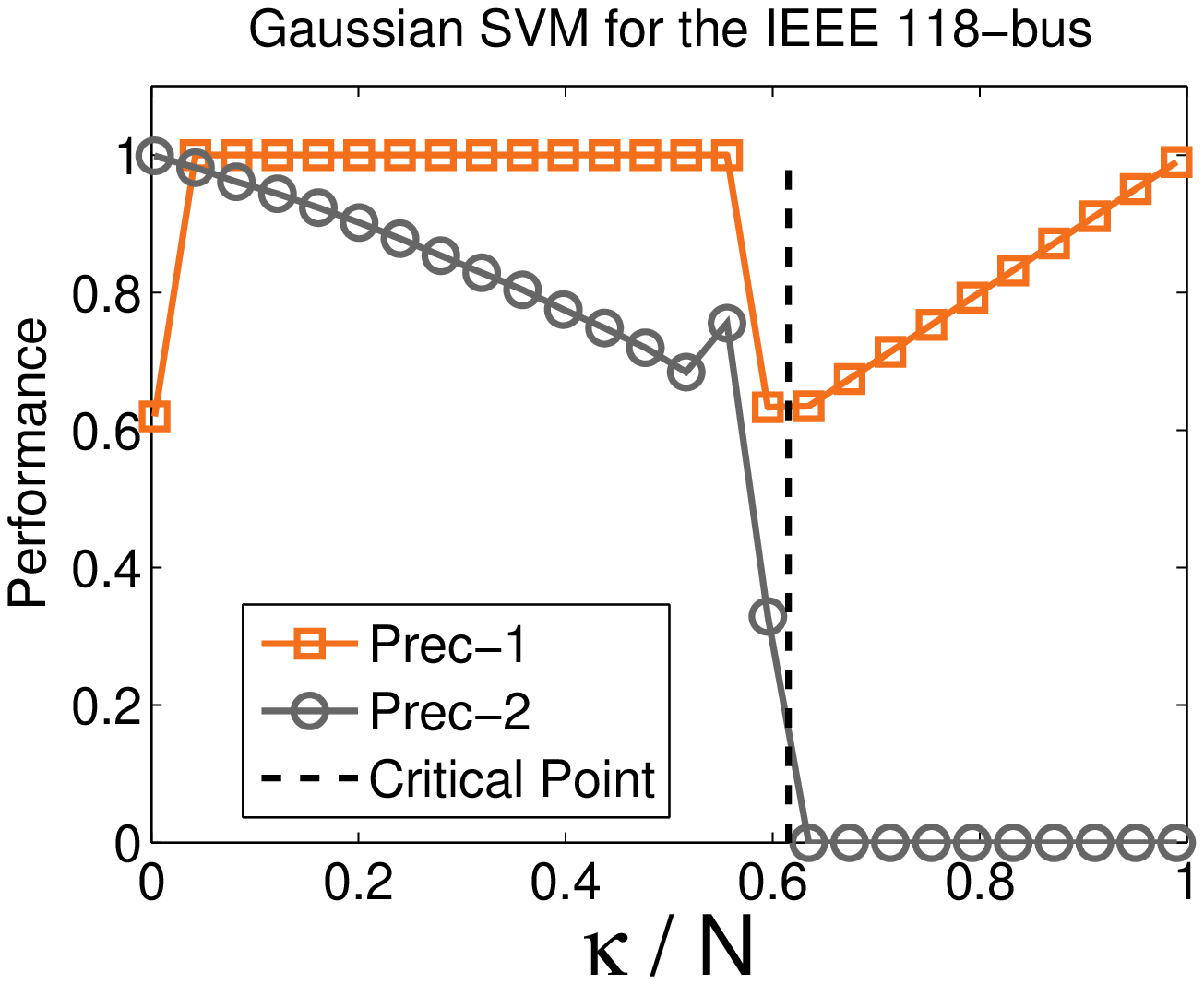}}
\subfloat[Gaussian SVM.]{\includegraphics[width=1.8in, height=1.45in]{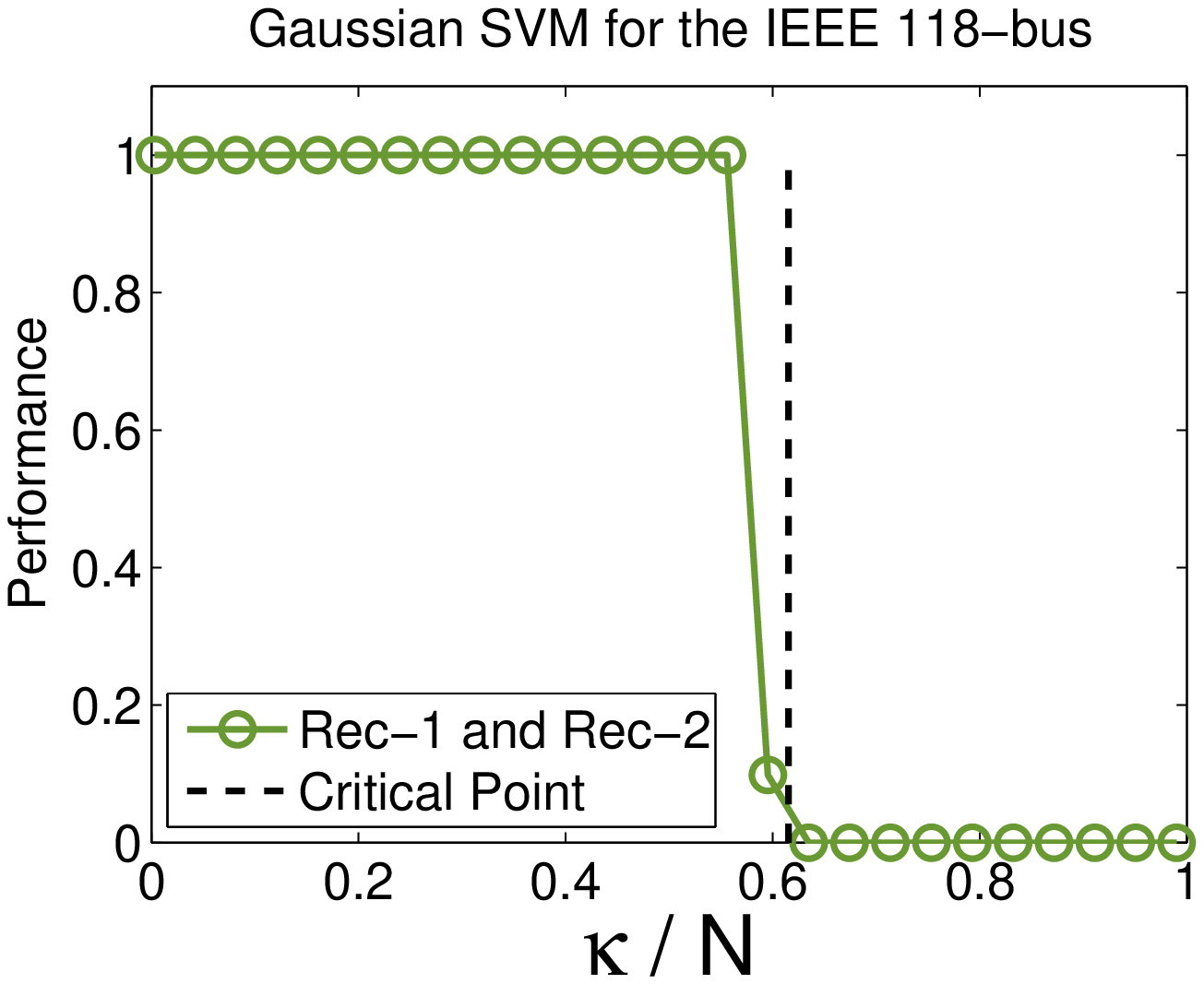}}  
\caption{Experiments using the SVM with linear and Gaussian kernels. Phase transitions of performance values occur at the critical point $\kappa^*$. See the text for more detailed explanation.}
\label{fig:4}
\end{figure*}

\begin{figure*}[ht!]
\centering
\subfloat[Results for the IEEE 57-bus.]{\includegraphics[width=1.75in, height=1.7in]{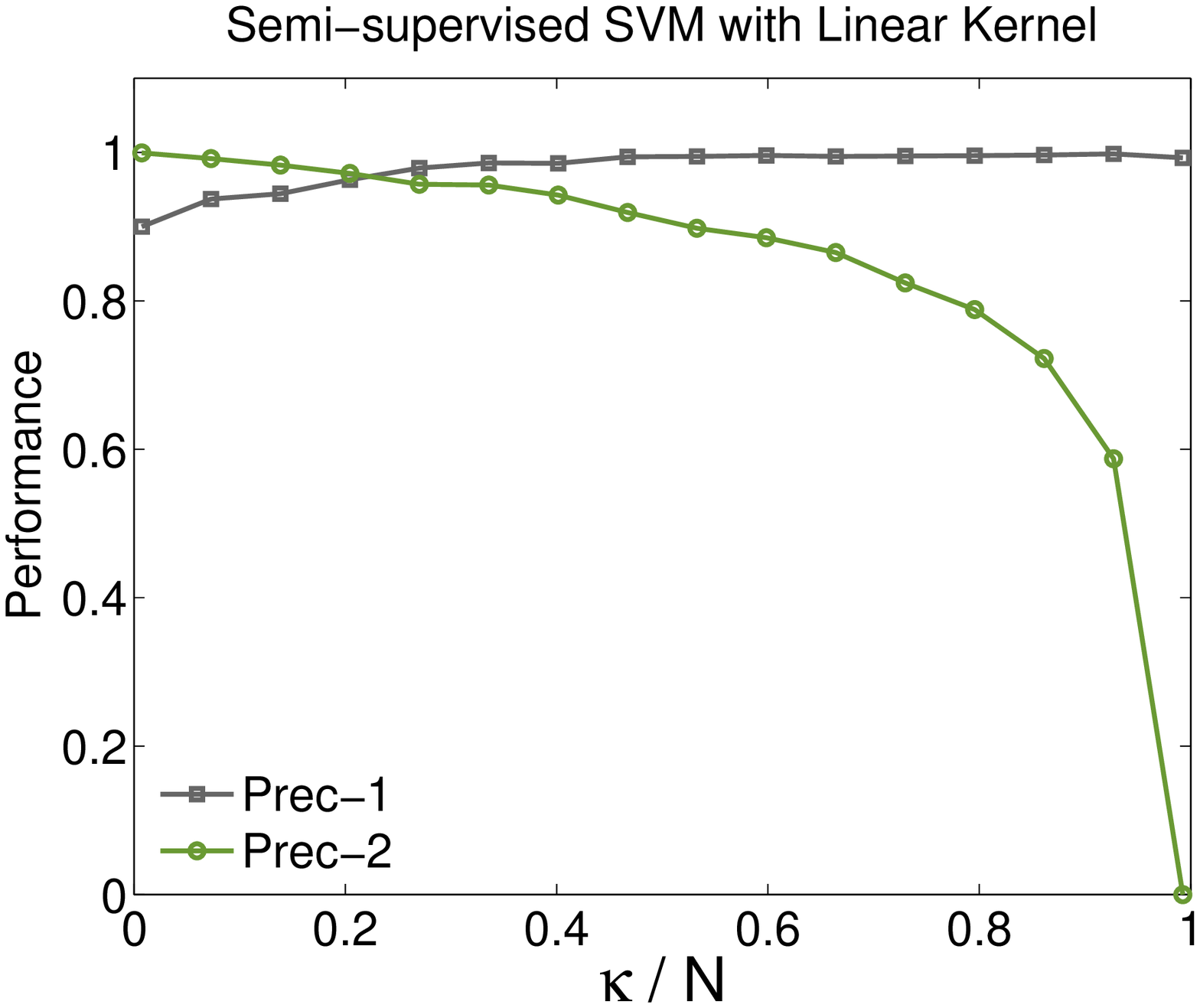}}
\subfloat[Results for the IEEE 57-bus.]{\includegraphics[width=1.75in, height=1.7in]{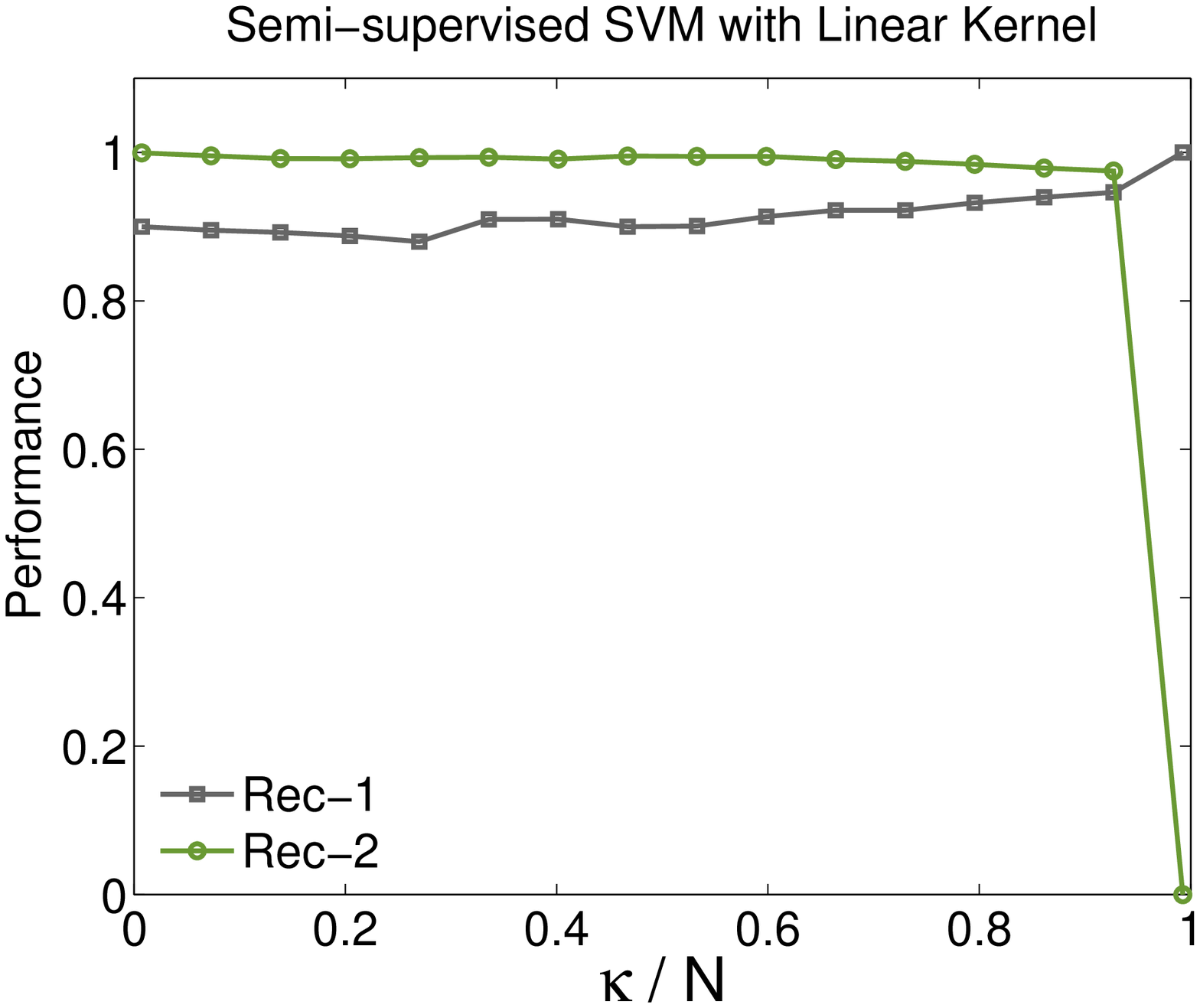}}
\subfloat[Results for the IEEE 118-bus.]{\includegraphics[width=1.75in, height=1.7in]{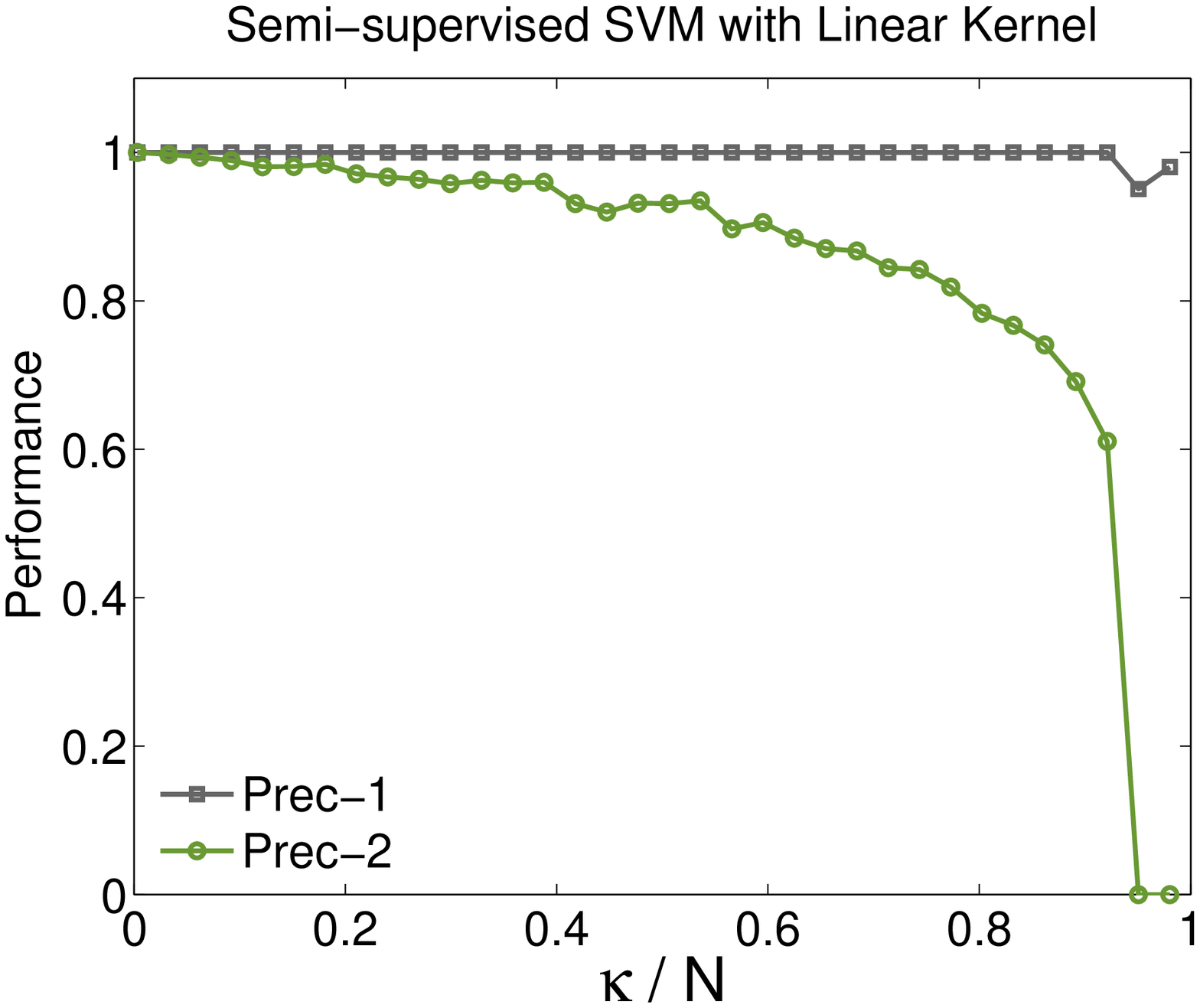}} 
\subfloat[Results for the IEEE 118-bus.]{\includegraphics[width=1.75in, height=1.7in]{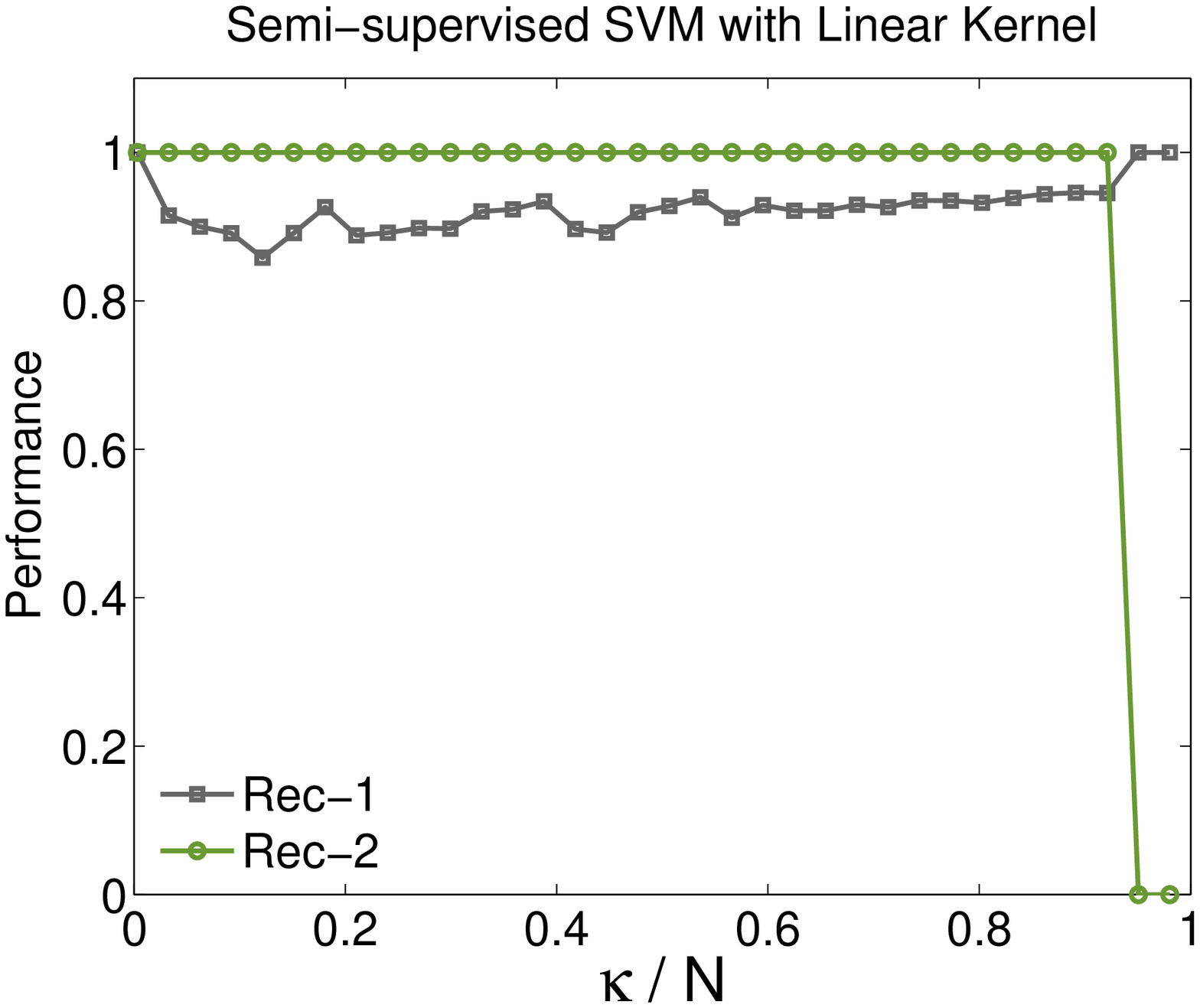}} \\
\subfloat[Results for the IEEE 57-bus.]{\includegraphics[width=1.75in, height=1.7in]{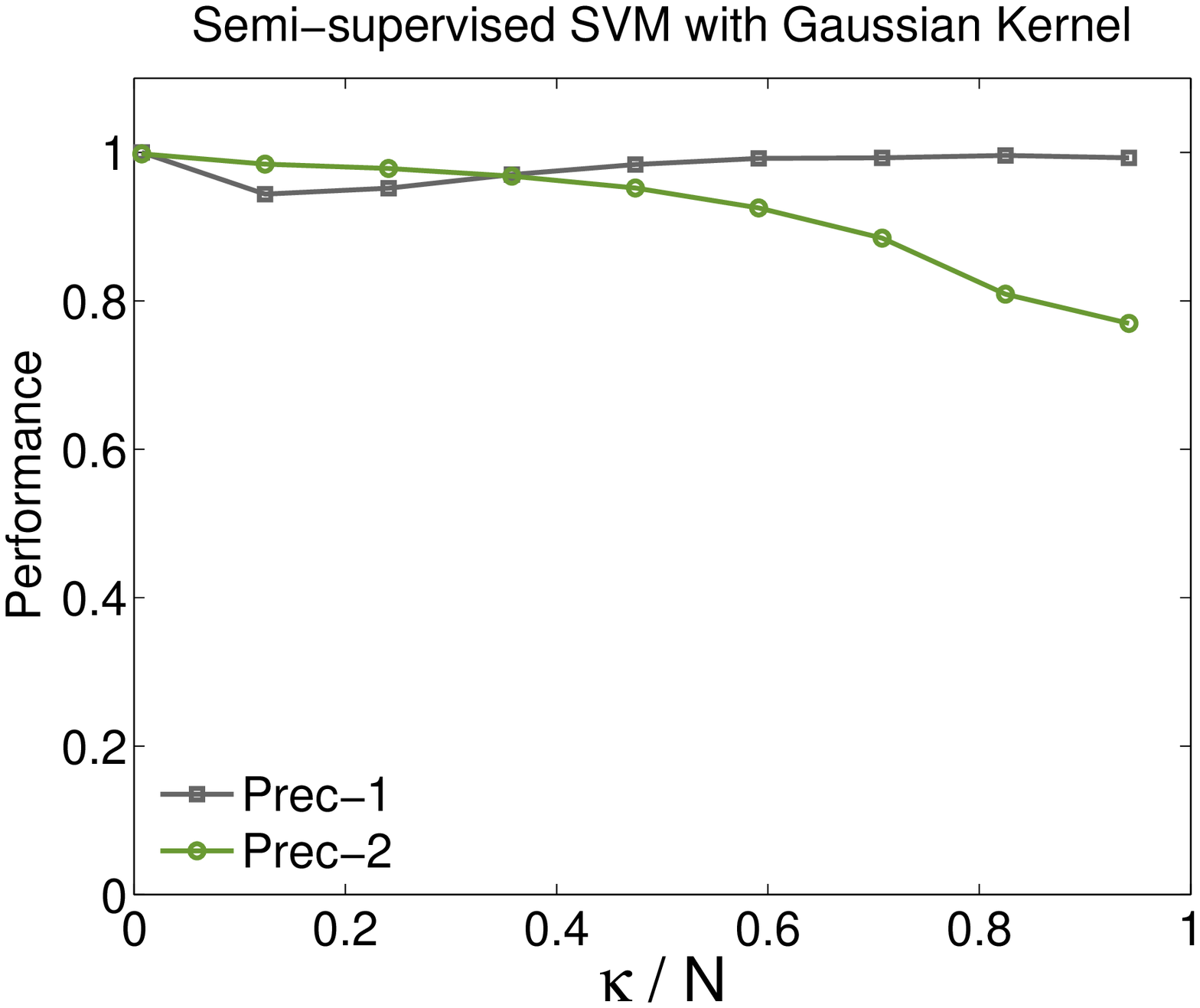}}
\subfloat[Results for the IEEE 57-bus.]{\includegraphics[width=1.75in, height=1.7in]{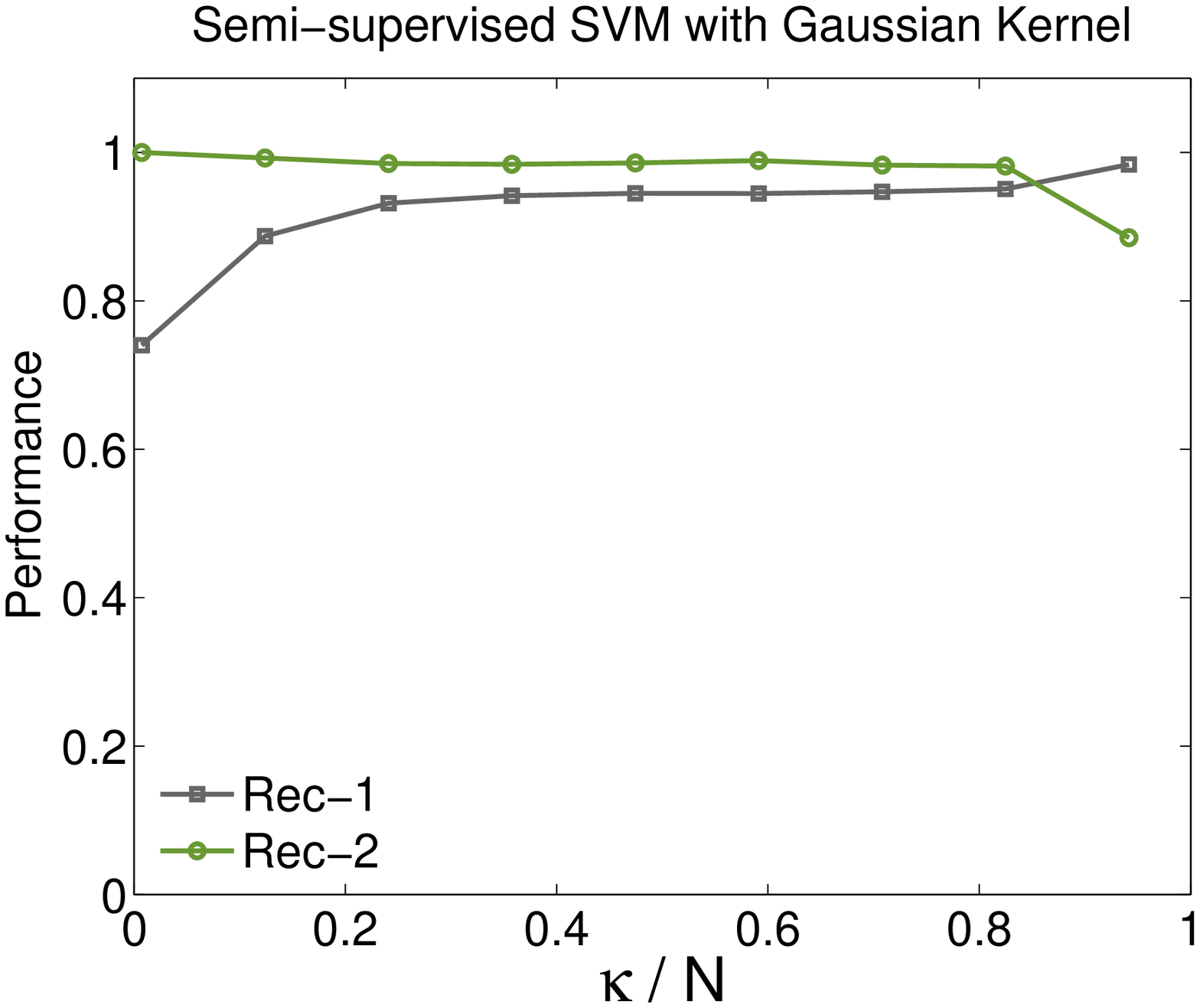}}
\subfloat[Results for the IEEE 118-bus.]{\includegraphics[width=1.75in, height=1.7in]{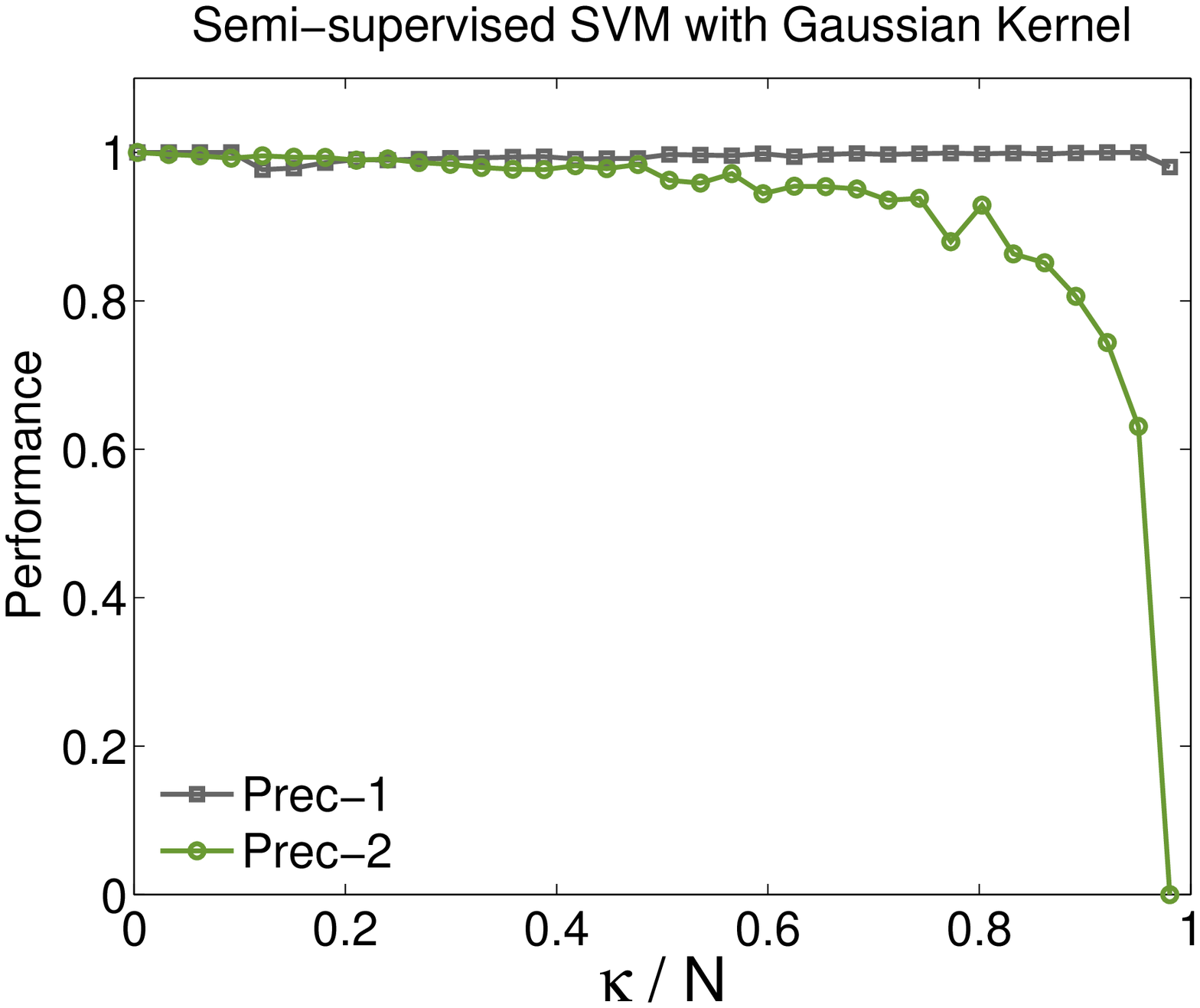}}
\subfloat[Results for the IEEE 118-bus.]{\includegraphics[width=1.75in, height=1.7in]{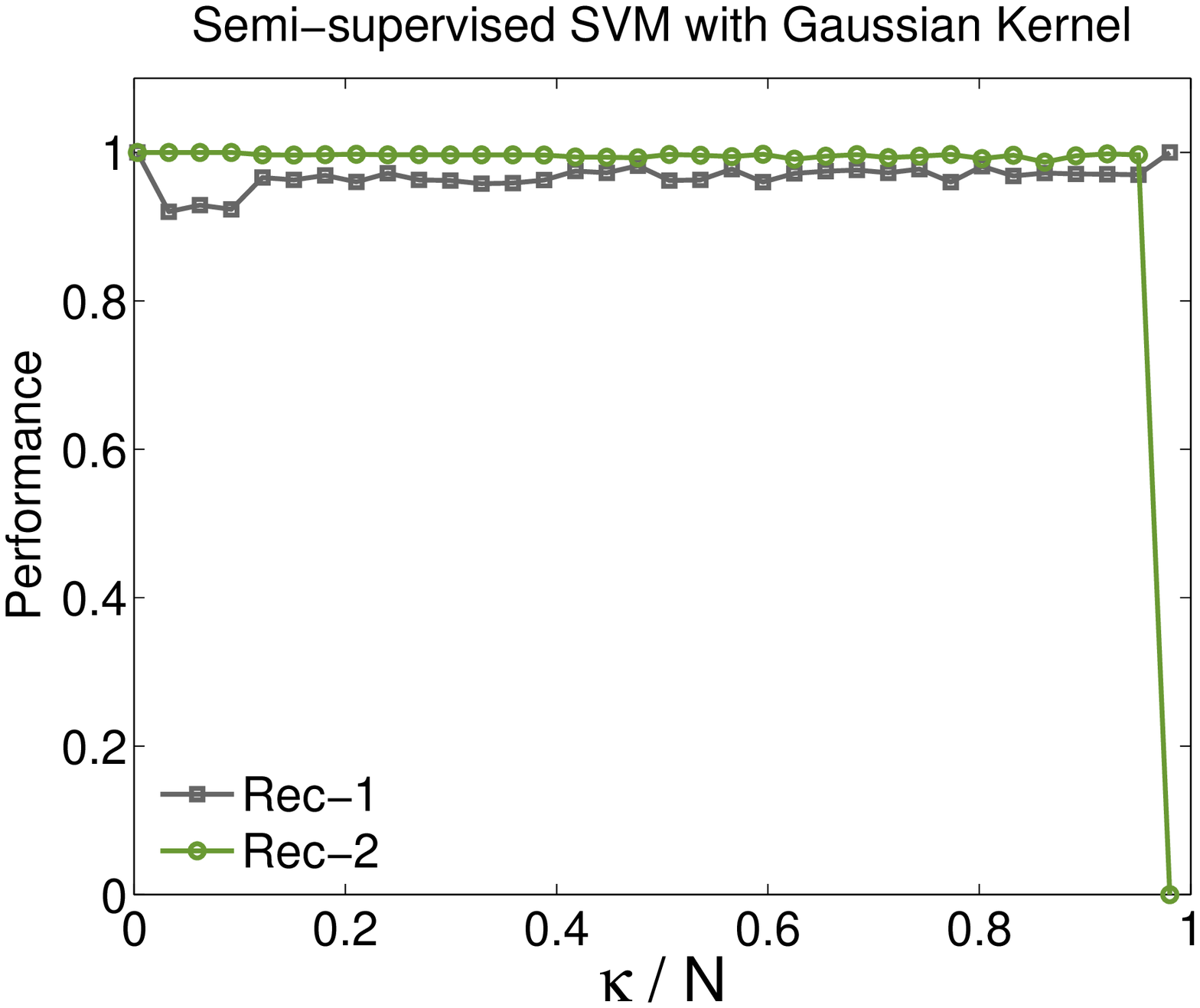}}
\caption{Sharp phase transitions are not observed in the semi-supervised SVM unlike the supervised SVM, since the information obtained from unlabeled data contributes to the performance values in the computation of the learning models.}
\label{fig:7}
\end{figure*}

We use the S3VM with default parameters as suggested in \cite{s3vm2}. The results of the semi-supervised SVM are shown in Fig.~\ref{fig:7}. We do not observe sharp phase transitions in the semi-supervised SVM unlike the supervised SVM, since the information obtained from unlabeled data contributes to the performance values in the computation of the learning models. For instance, Precision values of Class-2 decrease sharply near the critical point for the supervised SVM in Fig.~\ref{fig:4}. However, the semi-supervised SVM employs the unlabeled samples during the computation of the learning model in \eqref{eq:s3vm}, and partially solves this problem.

\begin{figure*}[ht!]
\centering
\subfloat[Adaboost for the IEEE 57-bus.]{\includegraphics[width=1.75in, height=1.7in]{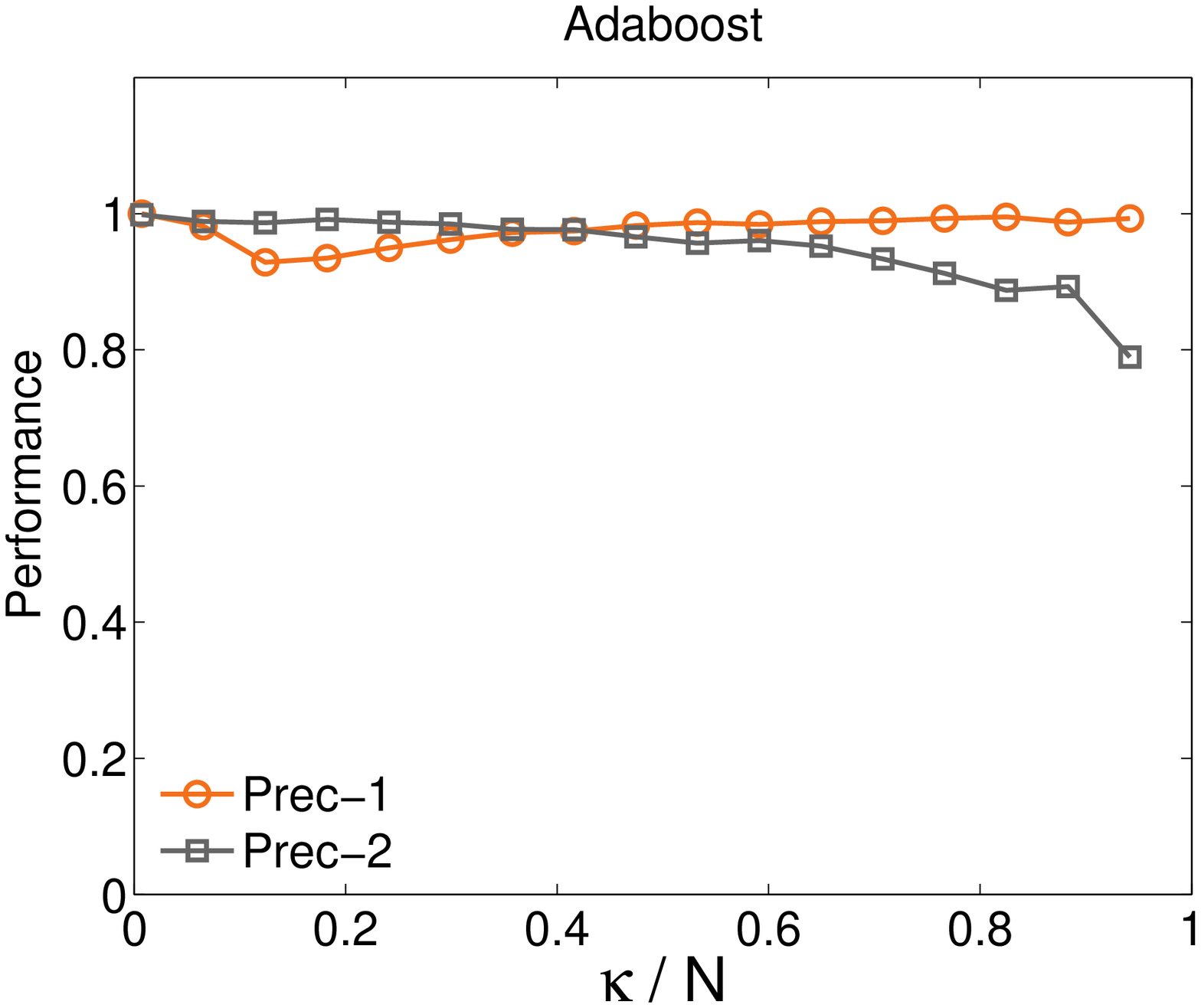}}
\subfloat[Adaboost for the IEEE 57-bus.]{\includegraphics[width=1.75in, height=1.7in]{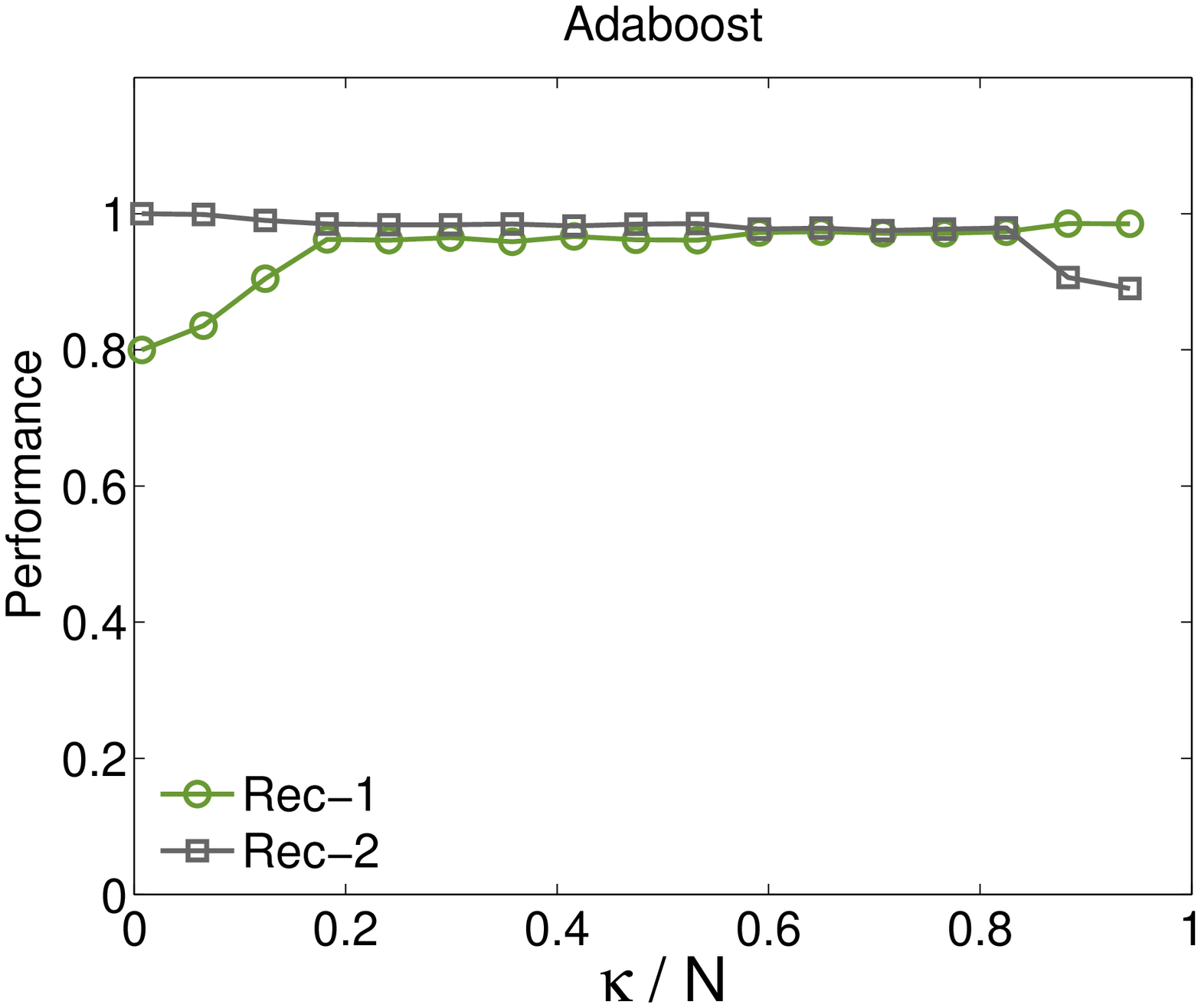}}
\subfloat[Adaboost for the IEEE 118-bus.]{\includegraphics[width=1.75in, height=1.7in]{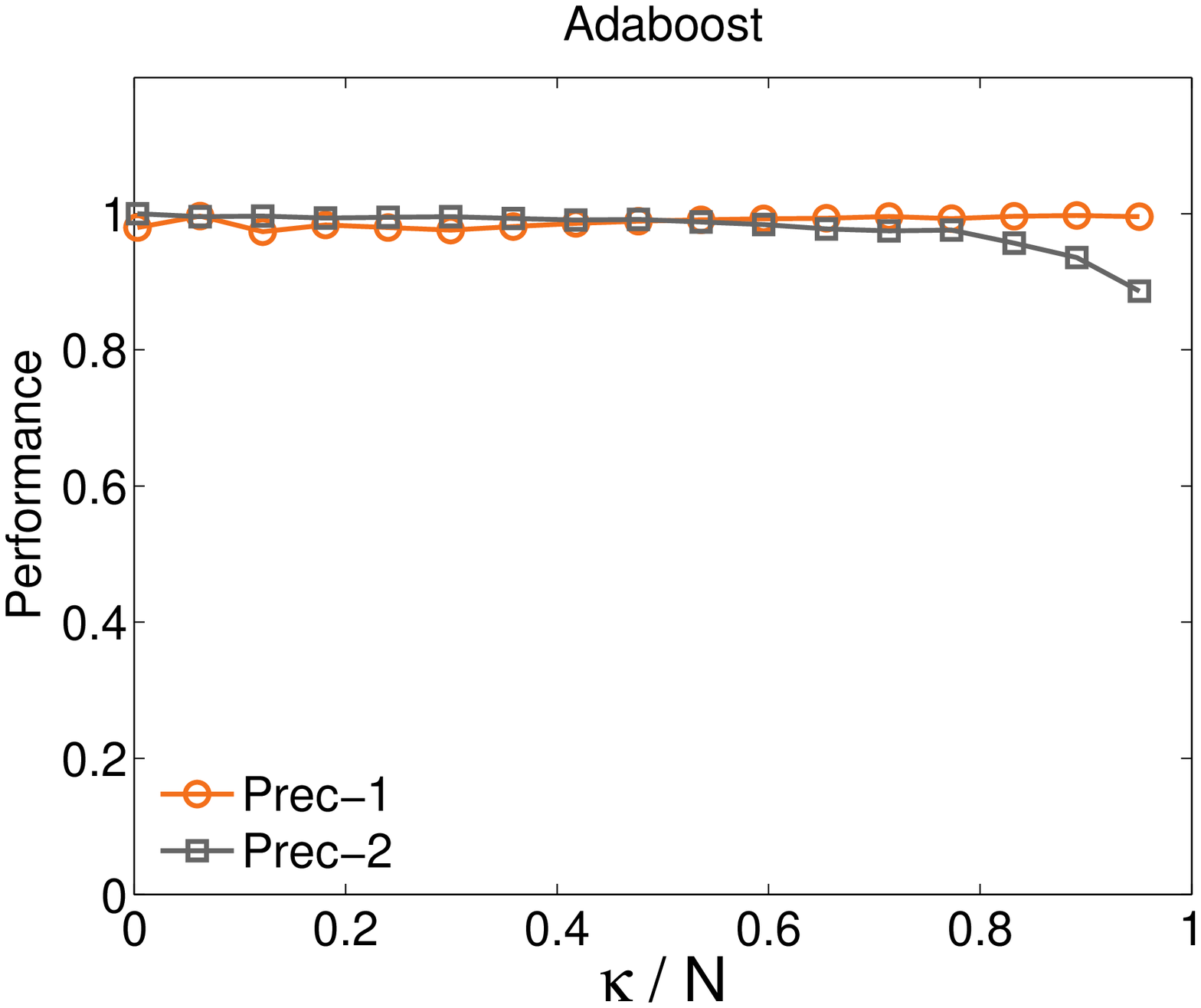}} 
\subfloat[Adaboost for the IEEE 118-bus.]{\includegraphics[width=1.75in, height=1.7in]{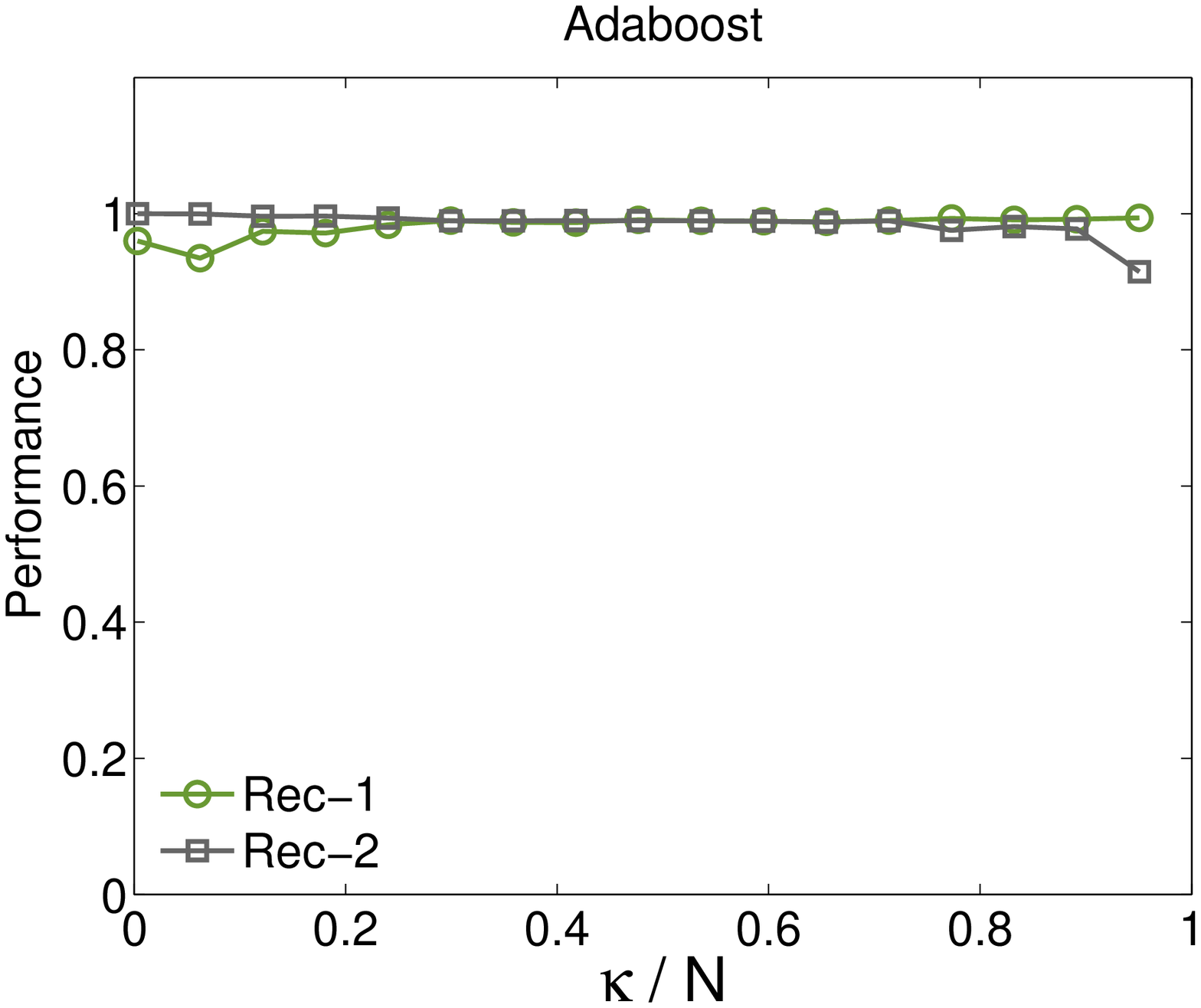}} \\
\subfloat[MKL for the IEEE 57-bus.]{\includegraphics[width=1.75in, height=1.7in]{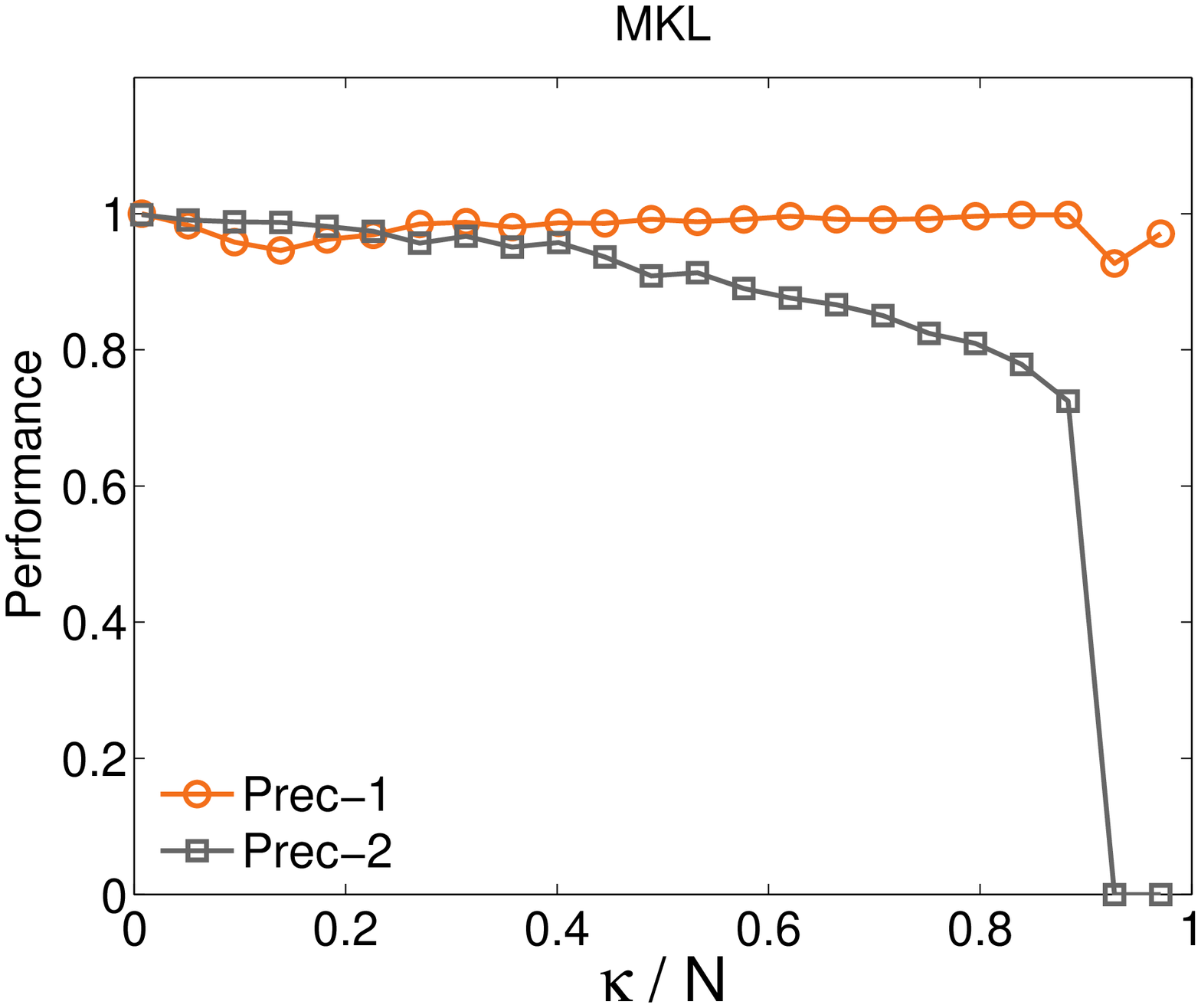}}
\subfloat[MKL for the IEEE 57-bus.]{\includegraphics[width=1.75in, height=1.7in]{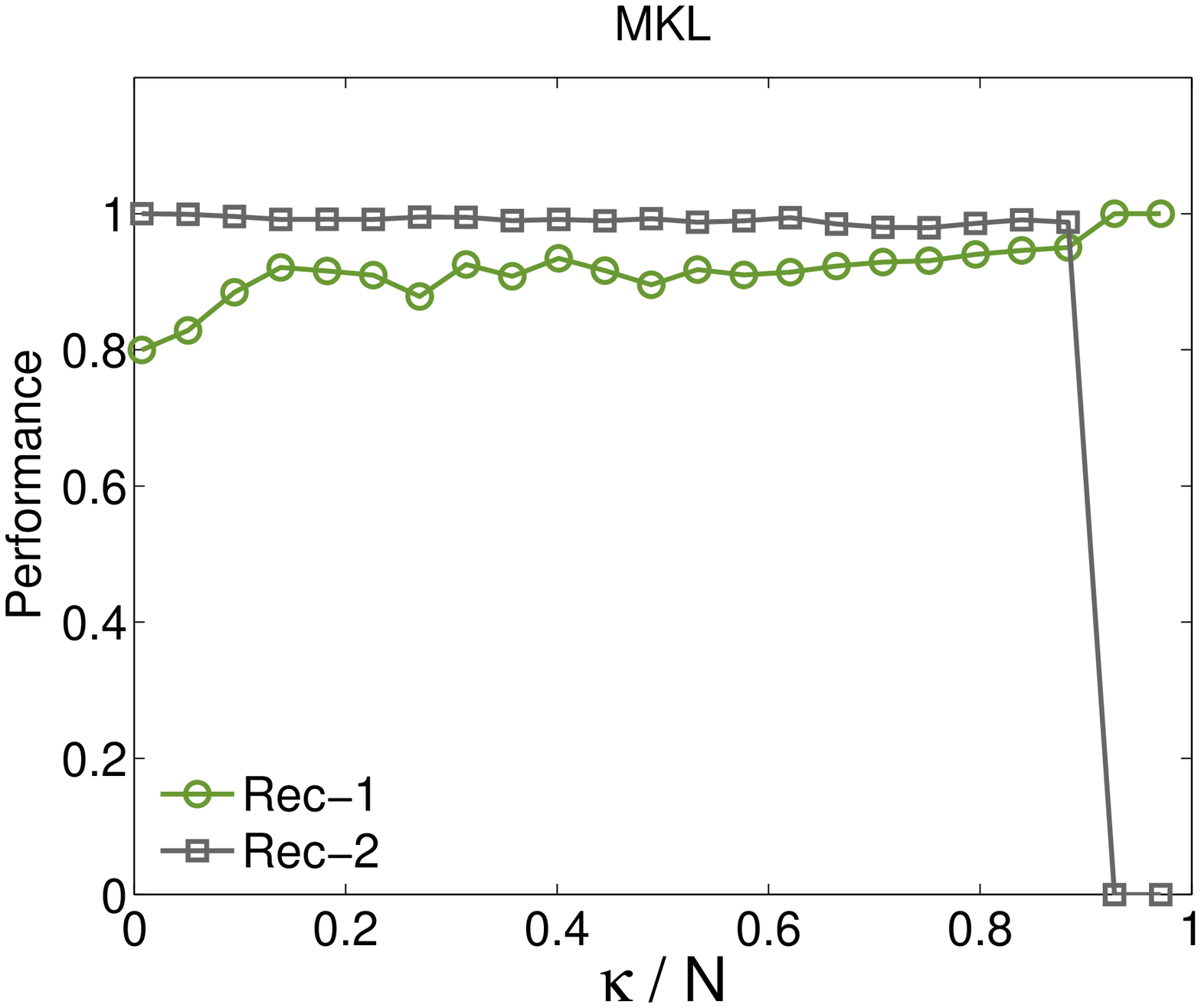}}
\subfloat[MKL for the IEEE 118-bus.]{\includegraphics[width=1.75in, height=1.7in]{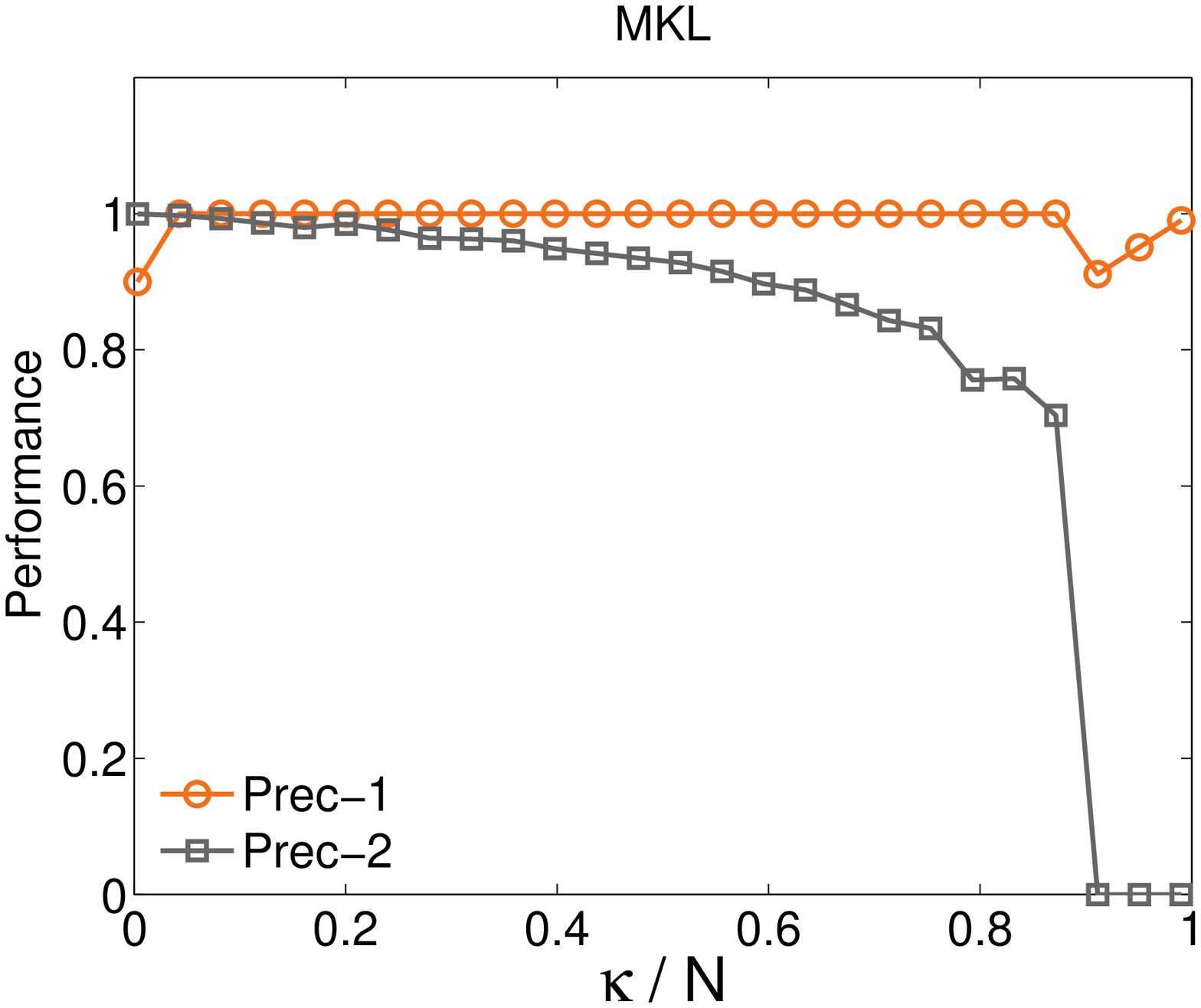}}
\subfloat[MKL for the IEEE 118-bus.]{\includegraphics[width=1.75in, height=1.7in]{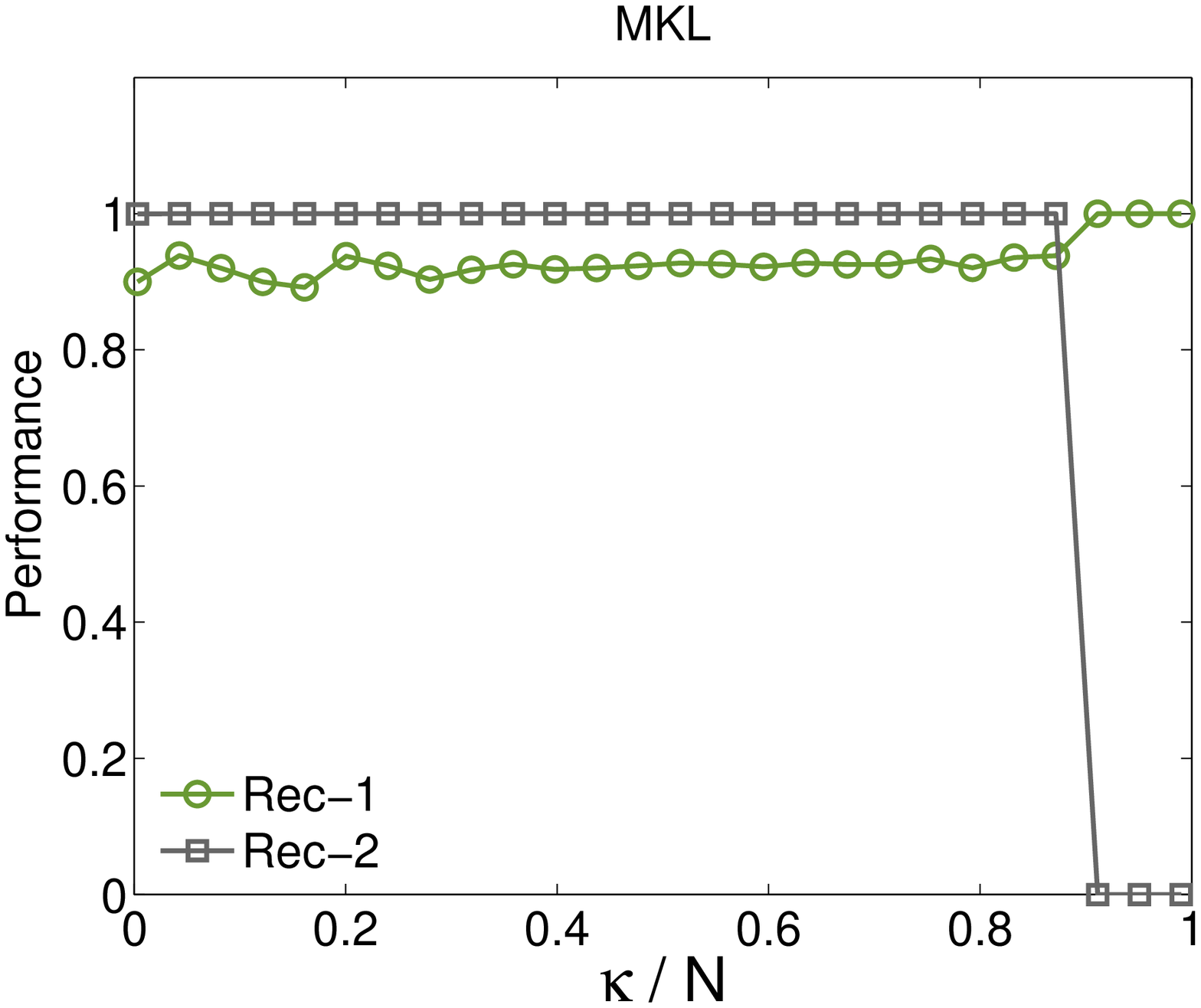}}
\caption{Experiments using Adaboost and MKL. Note that the \textit{fn} values of MKL are greater than the values of Adaboost, and there are no phase transitions of the performance values of MKL compared to the supervised SVM.}
\label{fig:8}
\end{figure*}

\subsection{Results for Decision and Feature Level Fusion Algorithms}

In this section, we analyze Adaboost and MKL. \textit{Decision stumps} are used as weak classifiers in Adaboost \cite{boost_book}. Each decision stump is a single-level two-leaf binary decision tree which is used to construct a set of dichotomies consisting of binary labelings of samples \cite{boost_book}. The number of weak classifiers is selected using leave-one-out cross-validation in the training set. We use MKL with a linear and a Gaussian kernel with the default parameters suggested in the Simple MKL implementation \cite{simplemkl}. The results  given in Fig.~\ref{fig:8} show that Recall values of MKL for Class-1 are less than the values of  Adaboost. In addition, Precision values of MKL decrease faster than the values of Adaboost as $\frac{\kappa}{N}$ increases for Class-2. Therefore, the \textit{fn} values of MKL are greater than the values of Adaboost, or in other words, the number of attacked measurements misclassified as secure by MKL is greater than that of Adaboost. This phenomenon is observed in the results for semi-supervised and supervised SVM given in the previous sections. However, there are no phase transitions of the performance values of MKL compared to the supervised SVM.

\begin{figure*}[ht!]
\centering
\subfloat[OP for the IEEE 57-bus.]{\includegraphics[width=1.75in, height=1.5in]{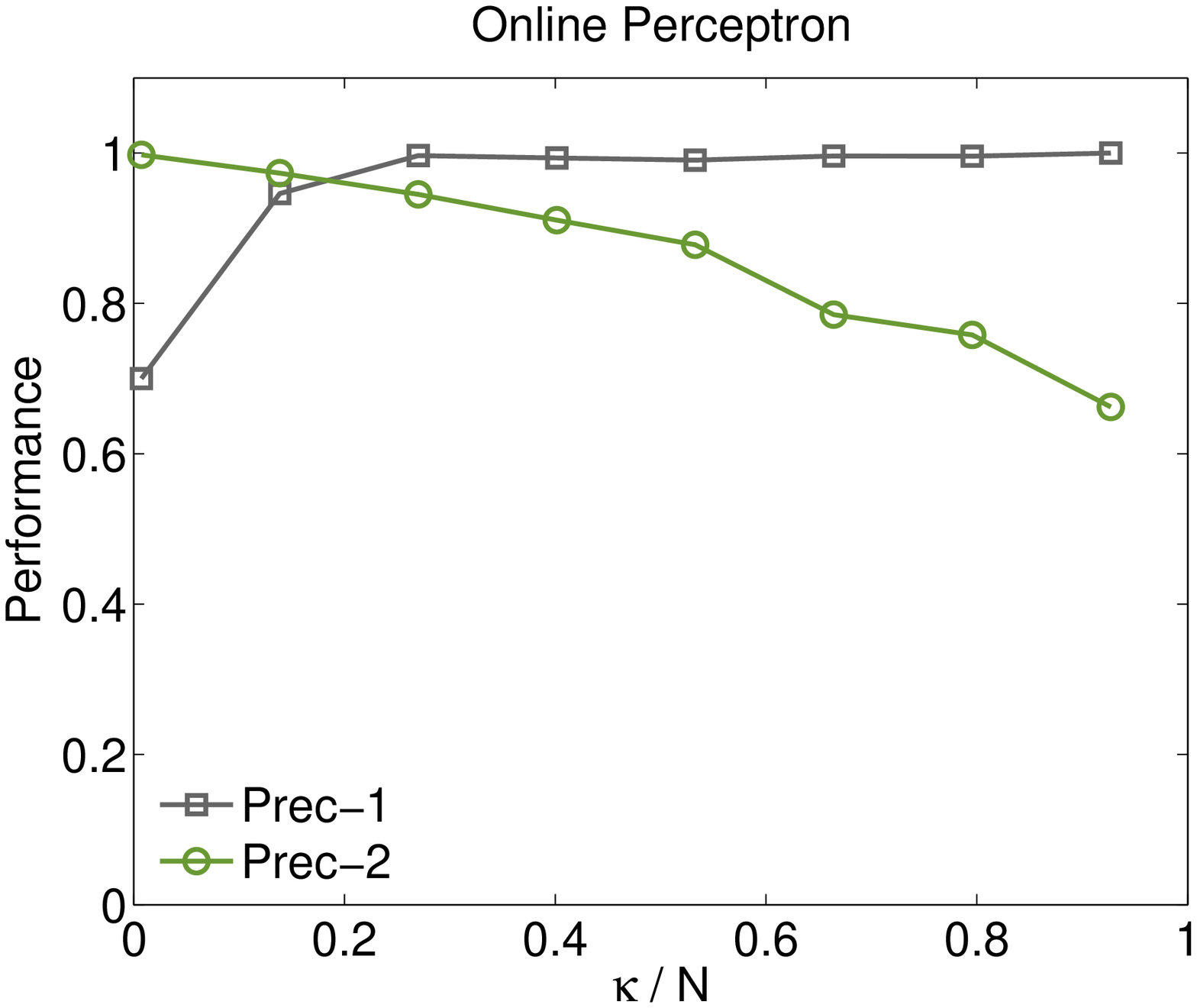}}
\subfloat[OP for the IEEE 57-bus.]{\includegraphics[width=1.75in, height=1.5in]{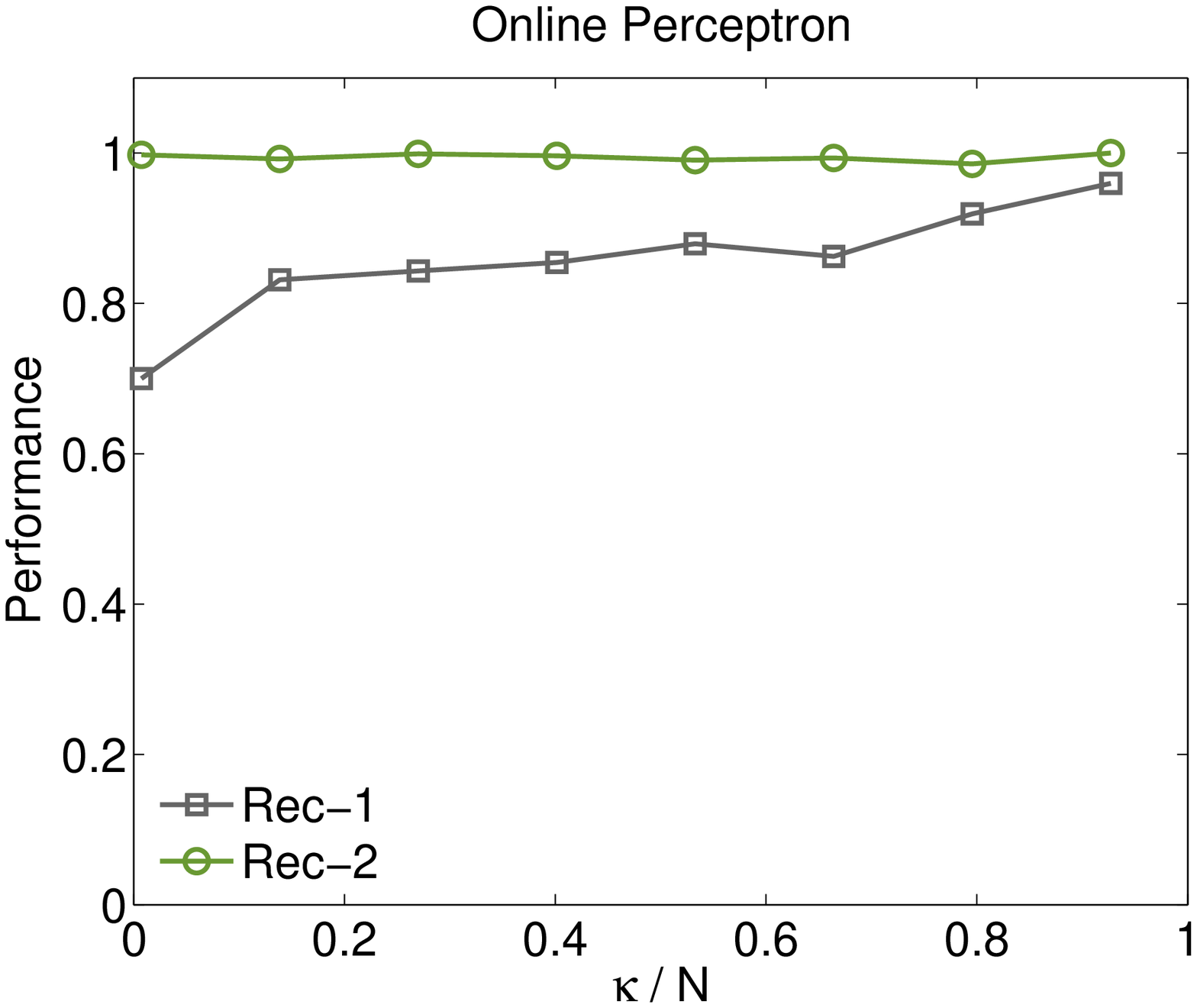}}
\subfloat[OP for the IEEE 118-bus.]{\includegraphics[width=1.75in, height=1.5in]{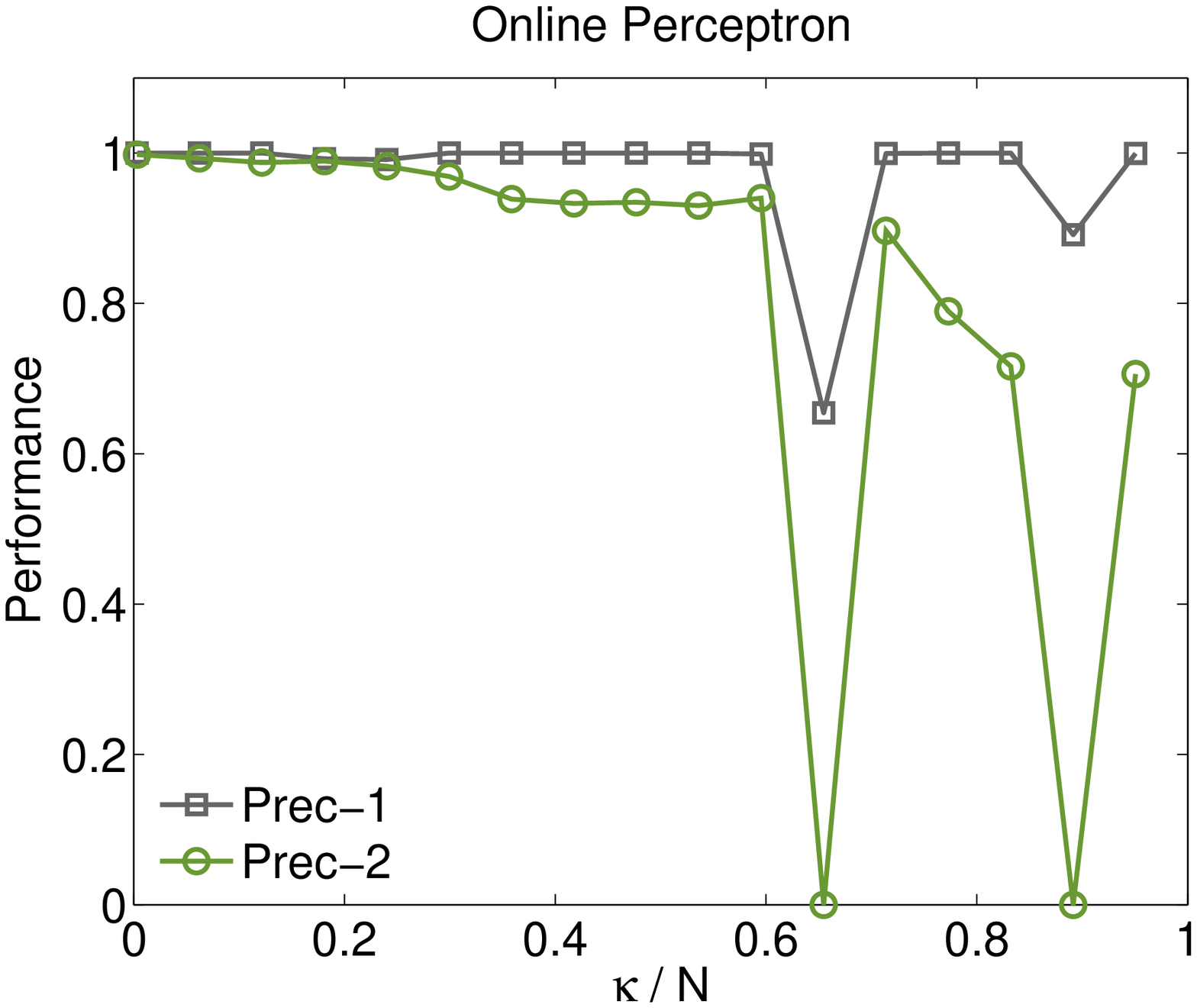}}
\subfloat[OP for the IEEE 118-bus.]{\includegraphics[width=1.75in, height=1.5in]{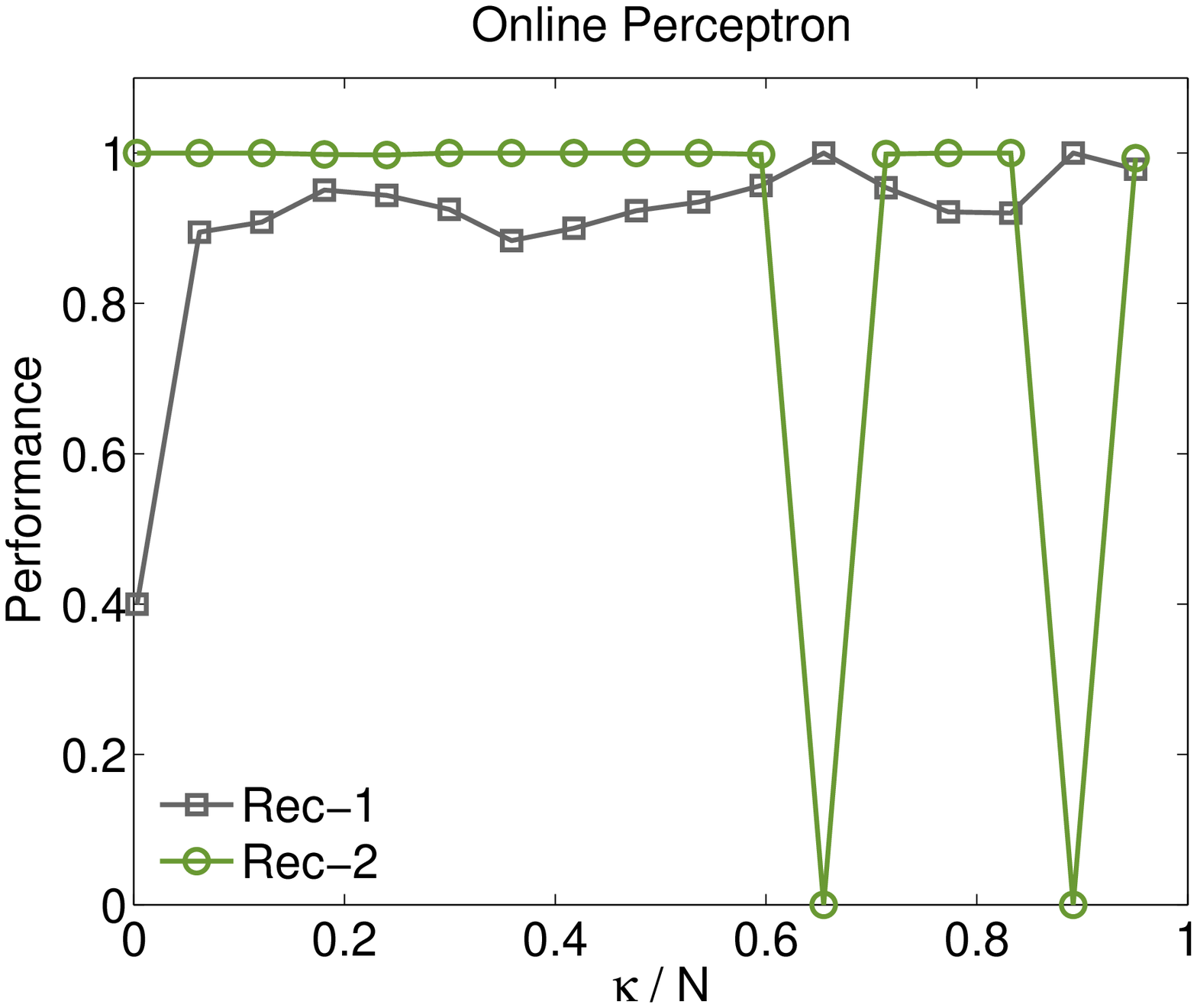}} \\
\subfloat[OPWM for the IEEE 57-bus.]{\includegraphics[width=1.8in, height=1.5in]{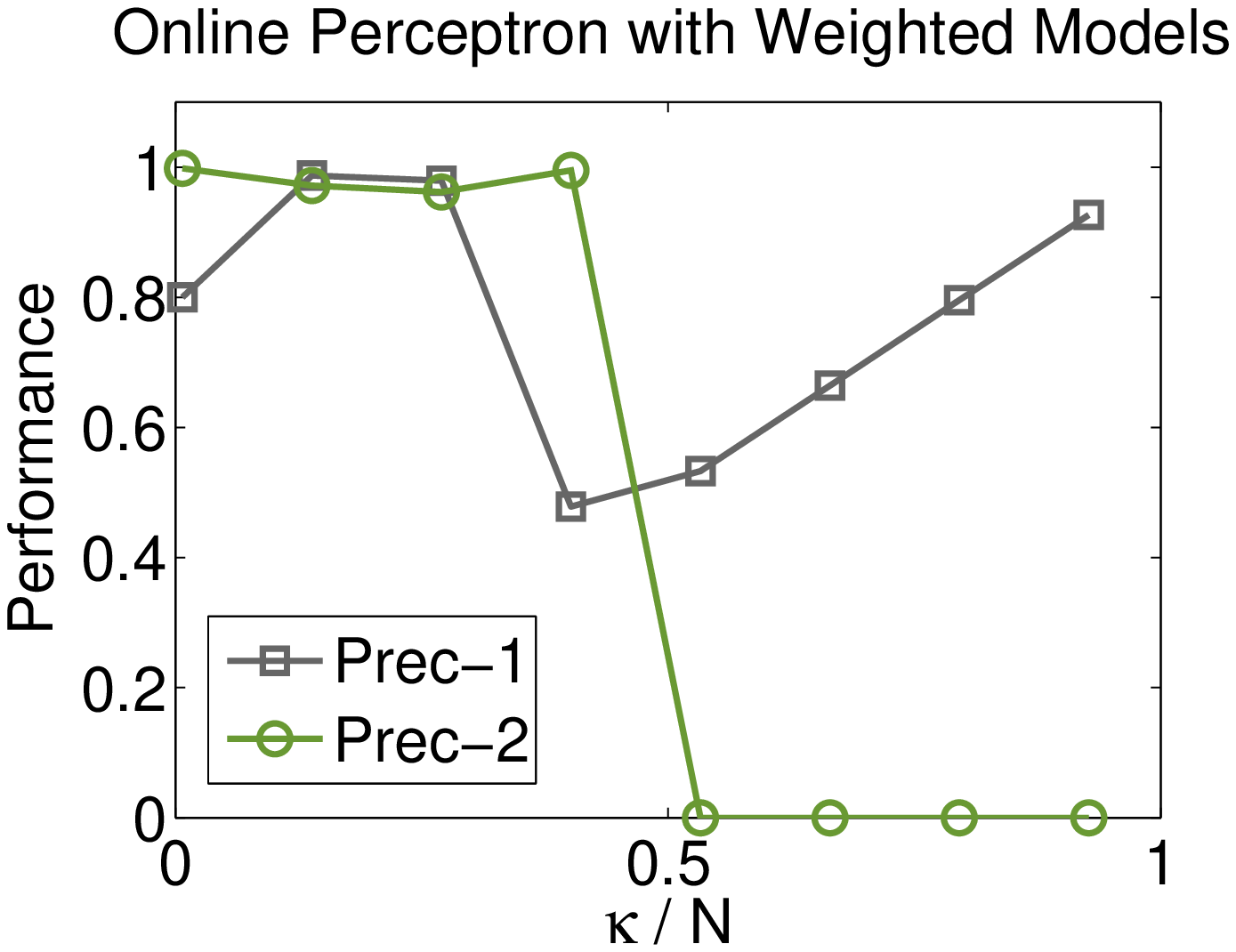}}
\subfloat[OPWM for the IEEE 57-bus.]{\includegraphics[width=1.8in, height=1.5in]{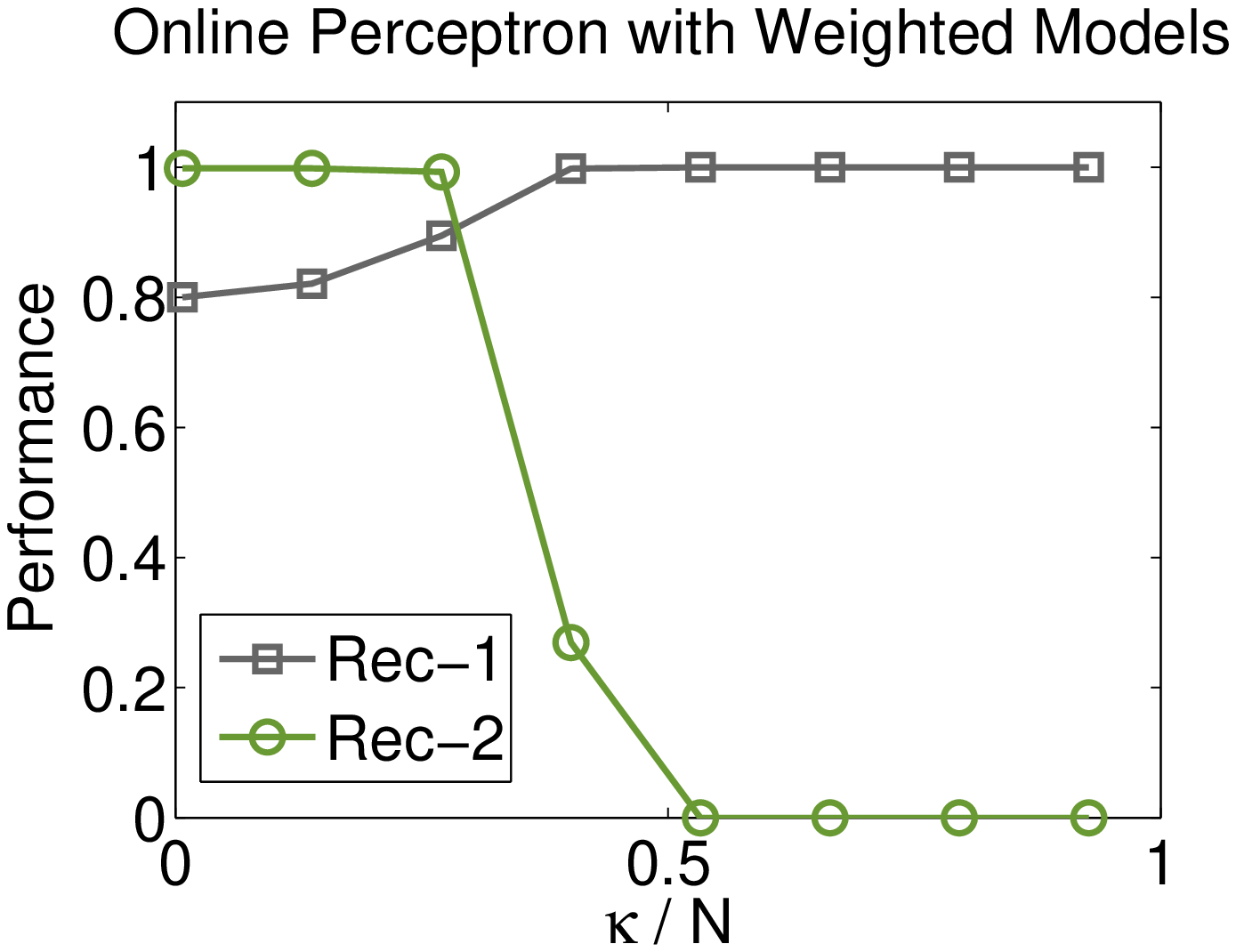}}
\subfloat[OPWM for the IEEE 118-bus.]{\includegraphics[width=1.8in, height=1.5in]{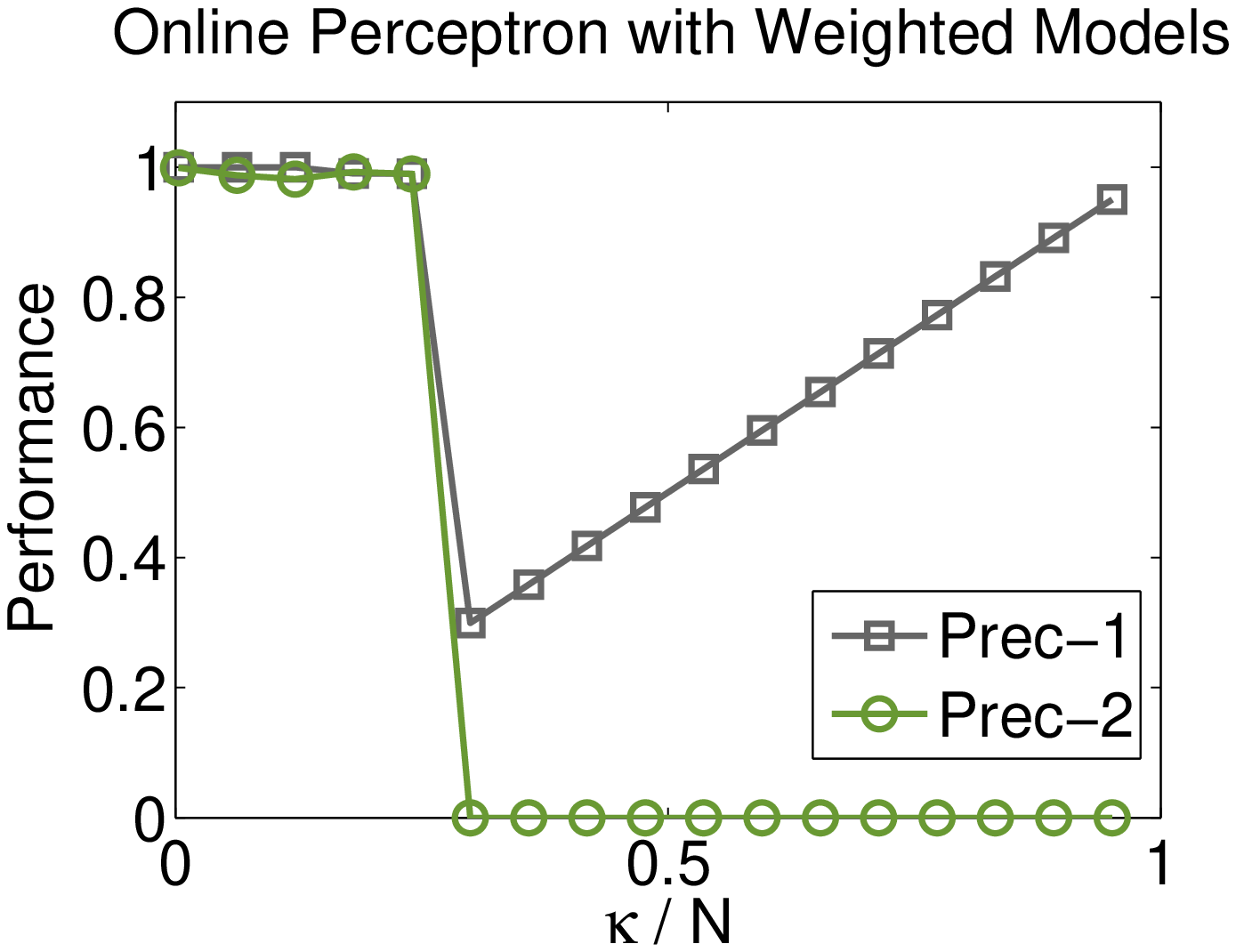}} 
\subfloat[OPWM for the IEEE 118-bus.]{\includegraphics[width=1.8in, height=1.5in]{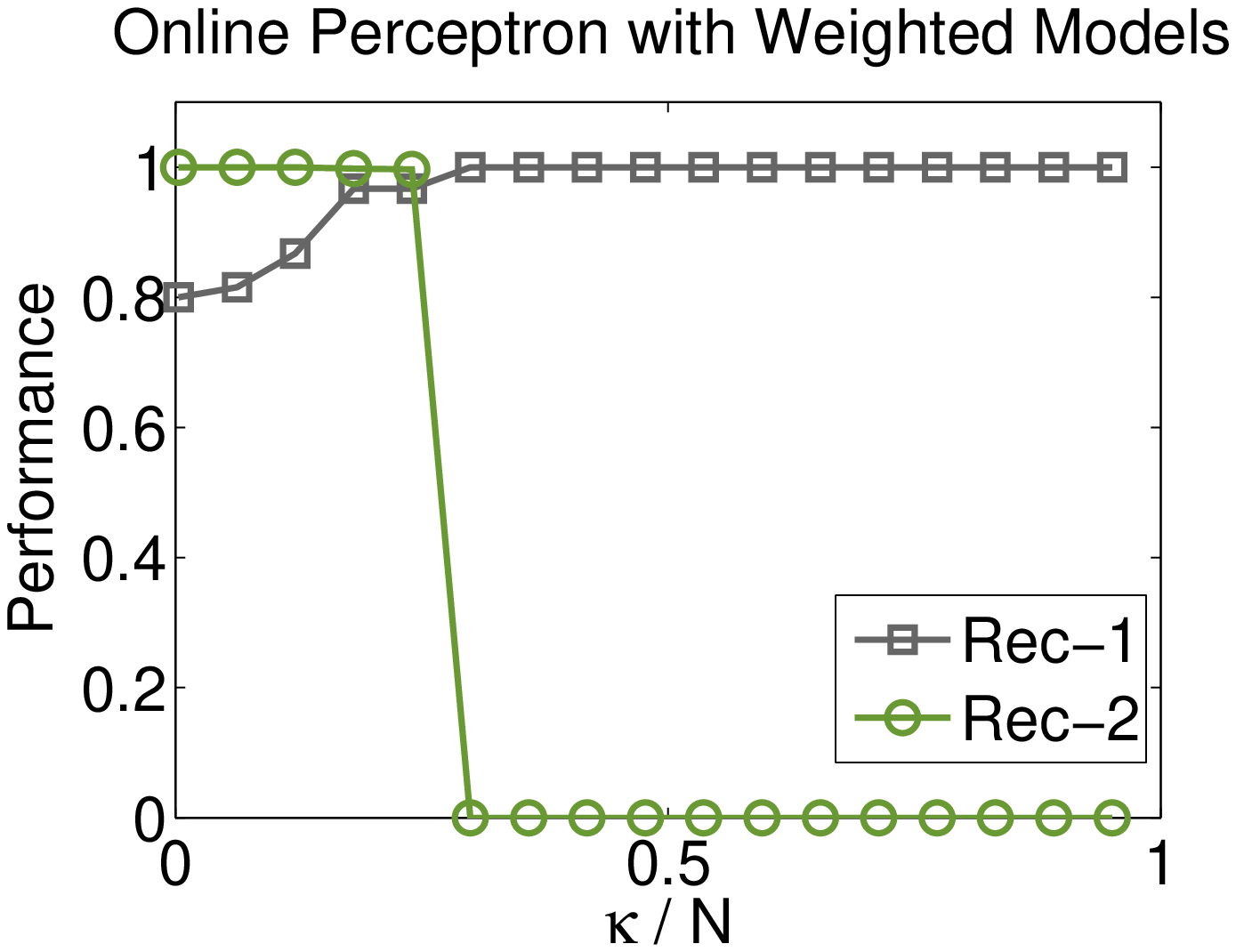}} \\

\subfloat[Online SVM for the IEEE 57-bus.]{\includegraphics[width=1.8in, height=1.5in]{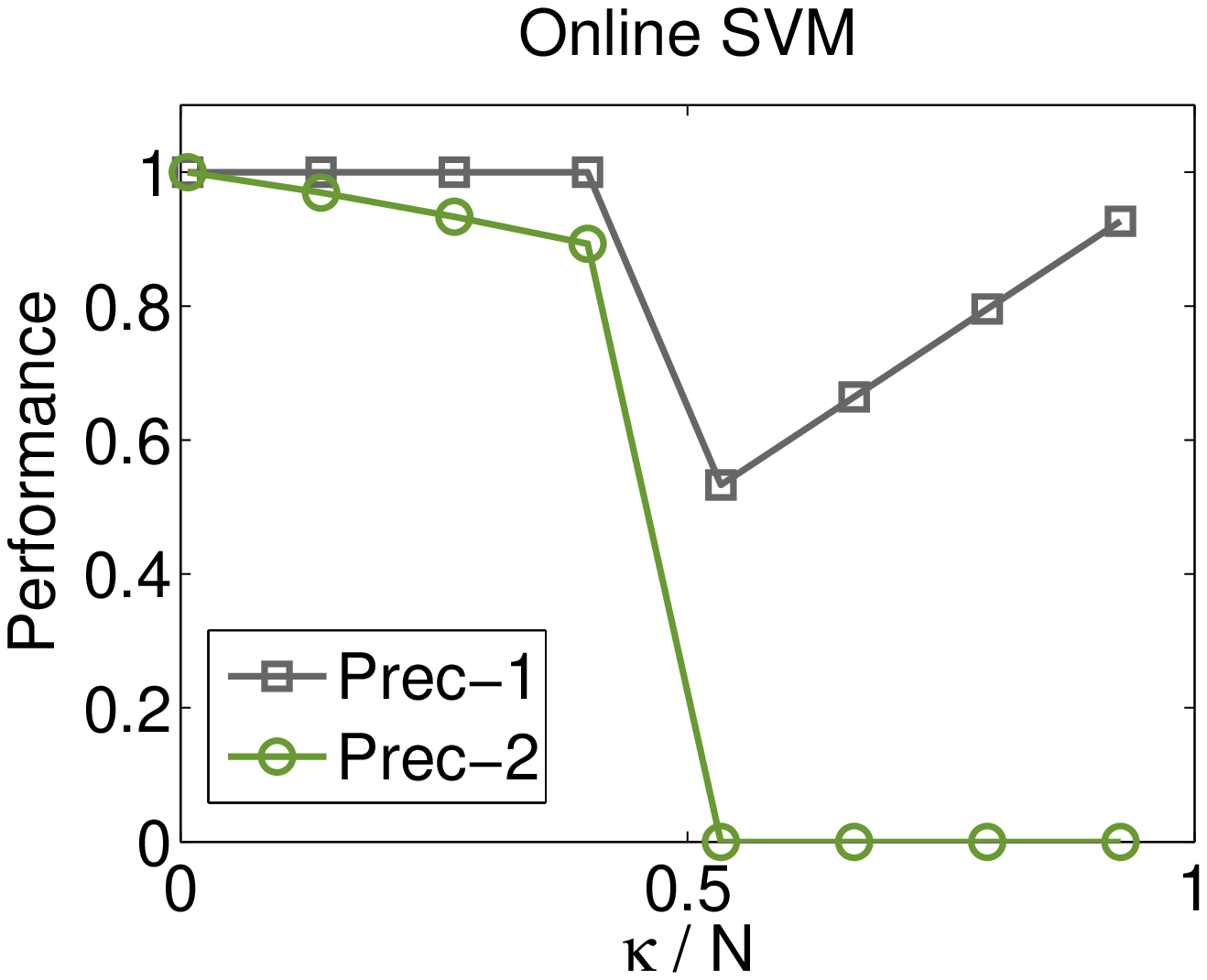}}
\subfloat[Online SVM for the IEEE 57-bus.]{\includegraphics[width=1.8in, height=1.5in]{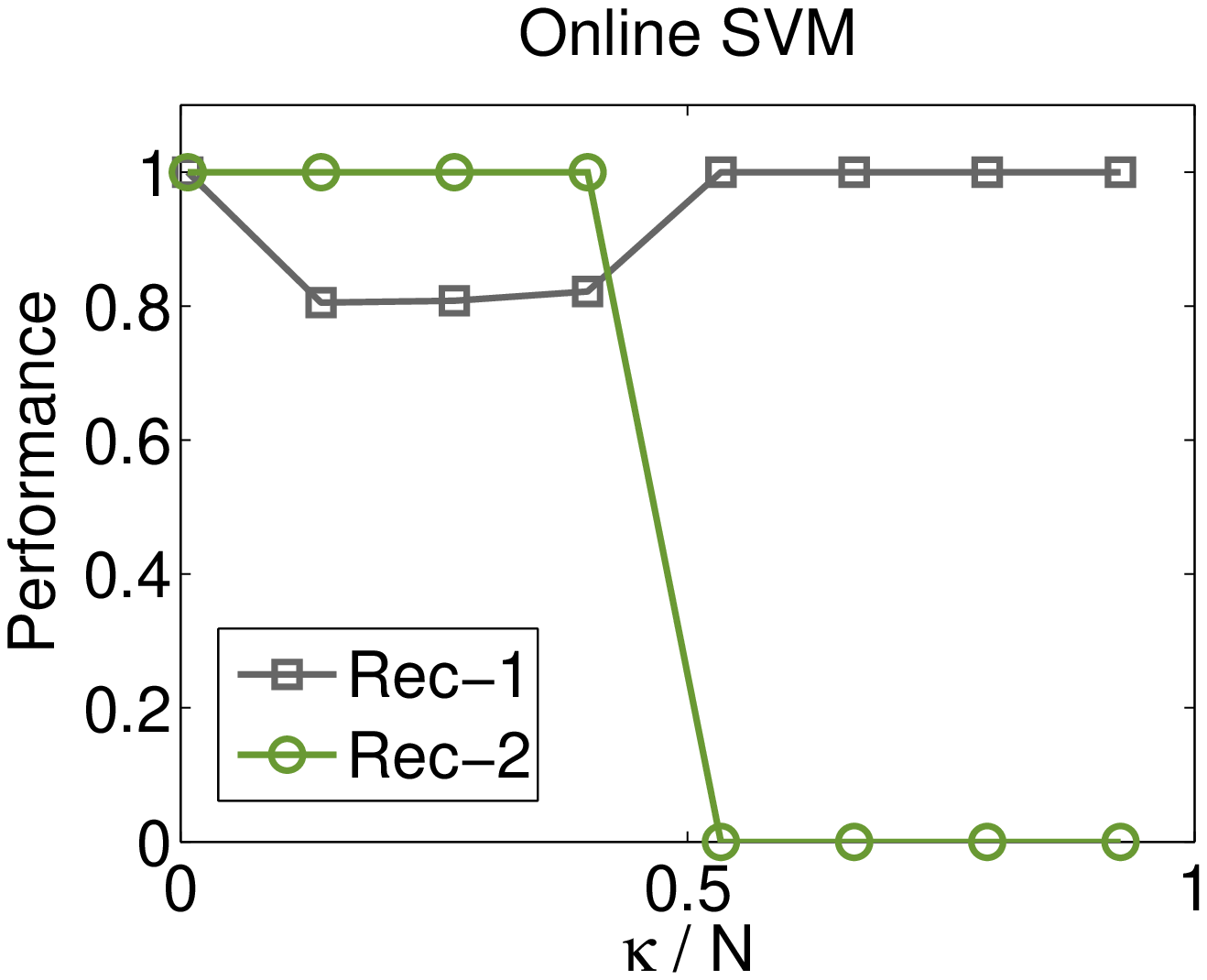}} 
\subfloat[Online SVM for the IEEE 118-bus.]{\includegraphics[width=1.8in, height=1.5in]{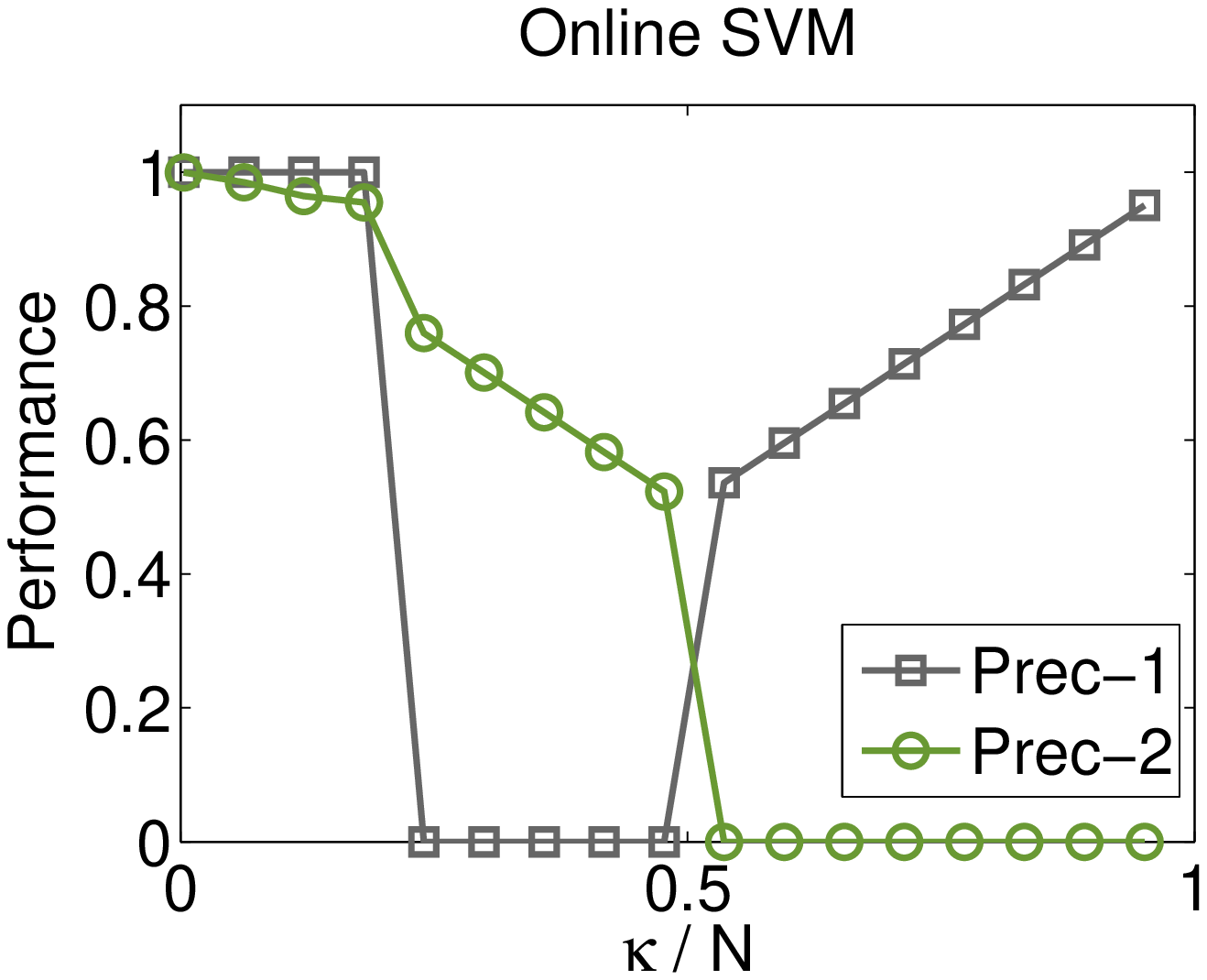}}
\subfloat[Online SVM for the IEEE 118-bus.]{\includegraphics[width=1.8in, height=1.5in]{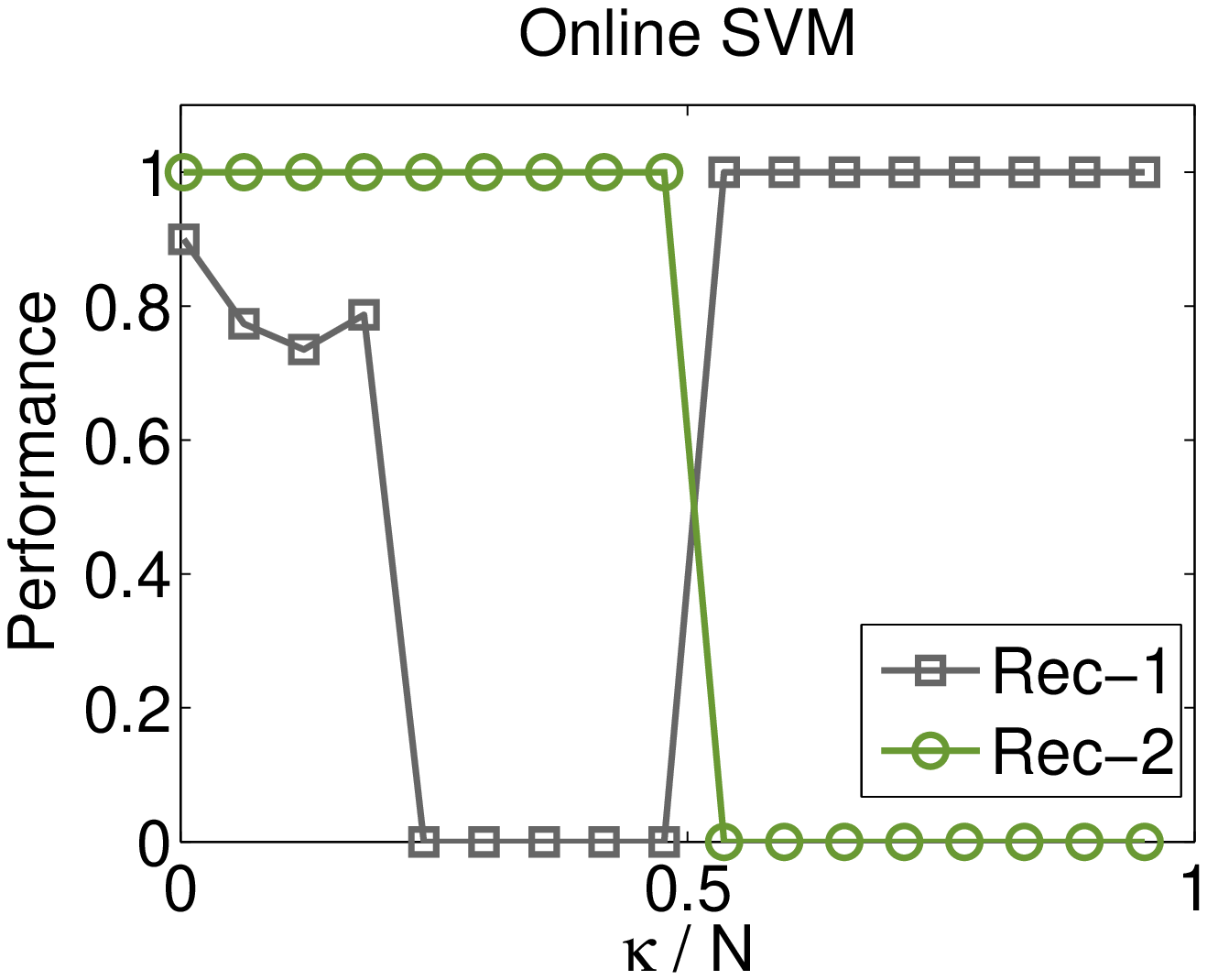}} \\
\subfloat[Online SLR for the IEEE 57-bus.]{\includegraphics[width=1.75in, height=1.5in]{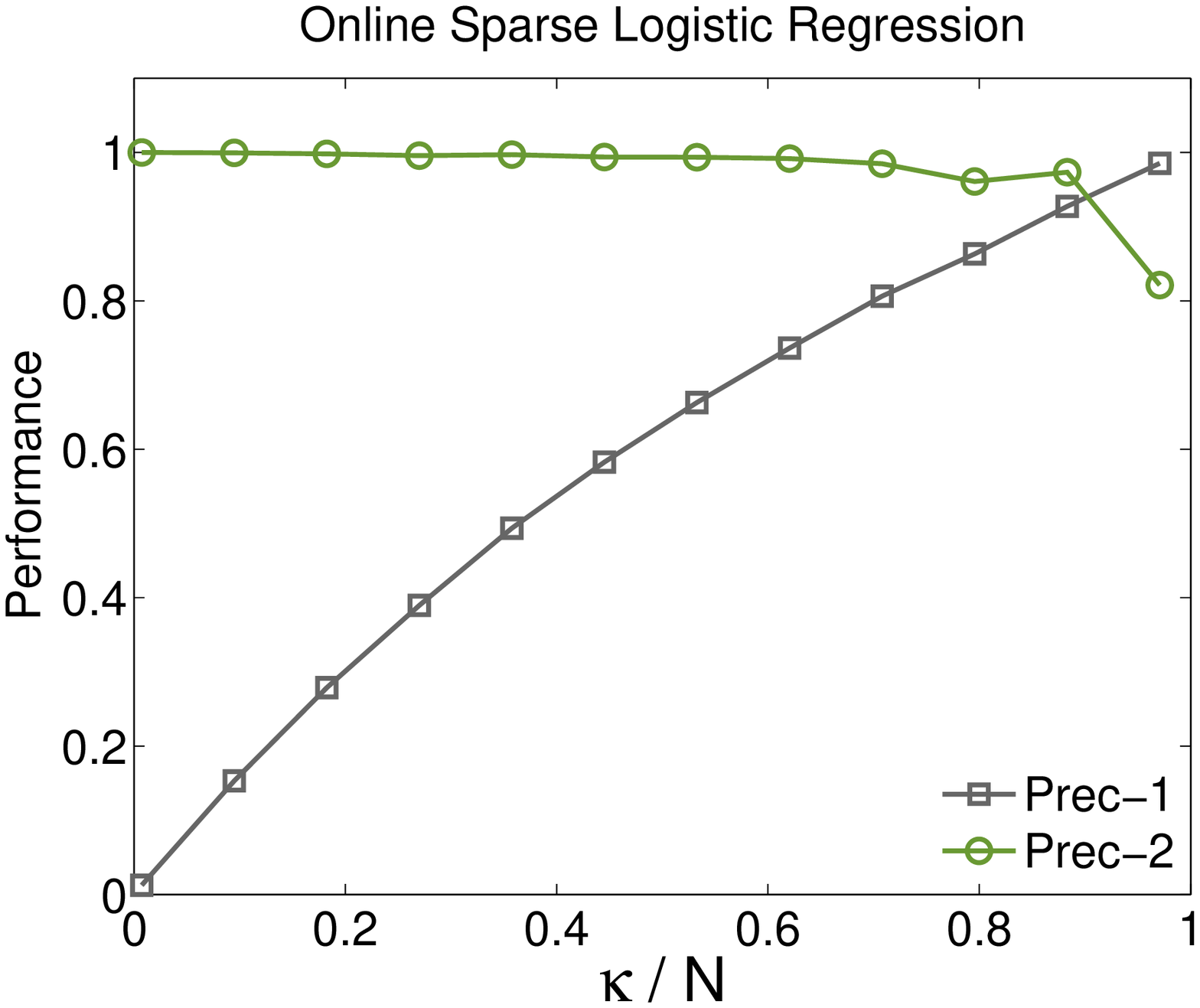}}
\subfloat[Online SLR for the IEEE 57-bus.]{\includegraphics[width=1.75in, height=1.5in]{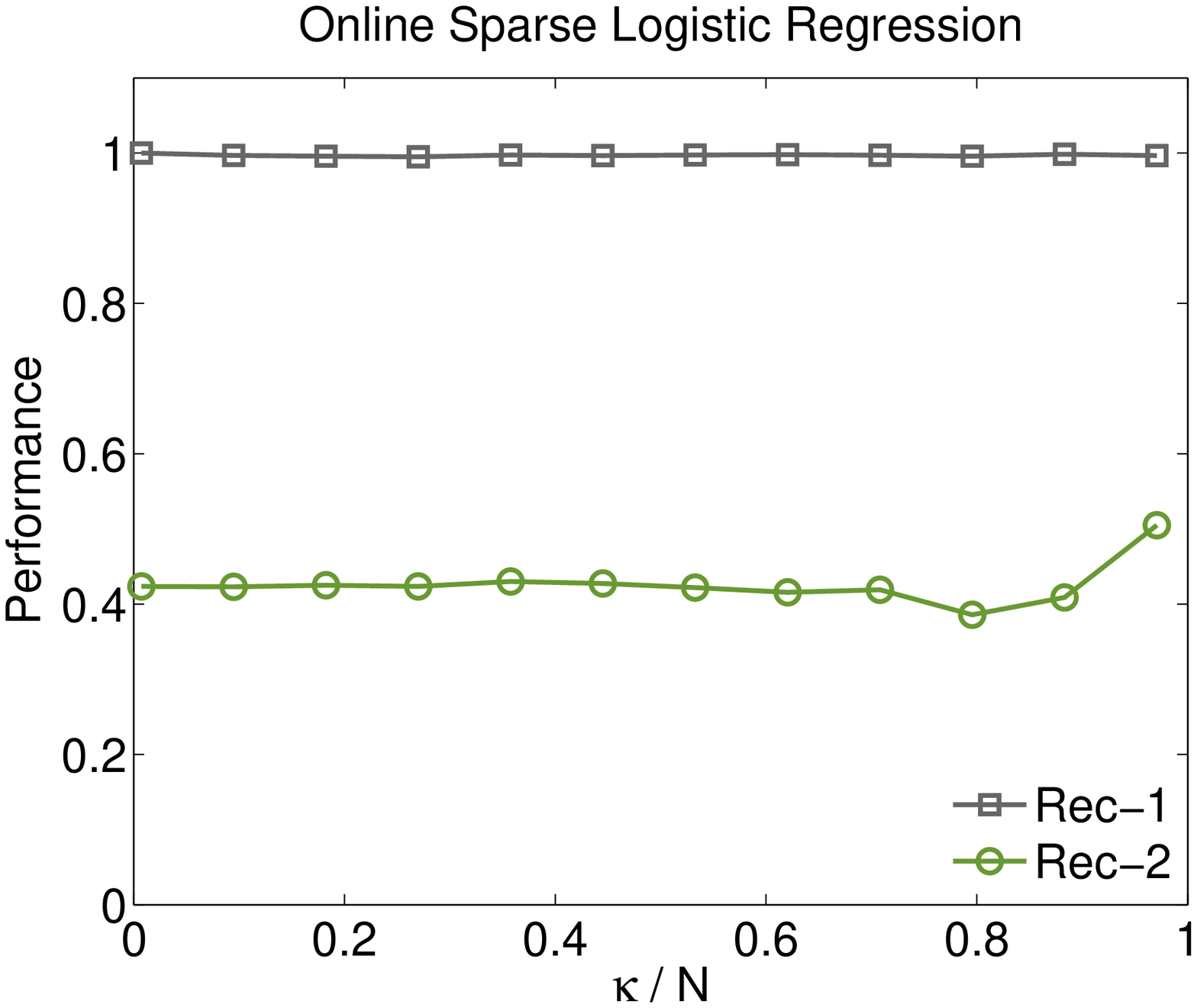}}
\subfloat[Online SLR for the IEEE 118-bus.]{\includegraphics[width=1.75in, height=1.5in]{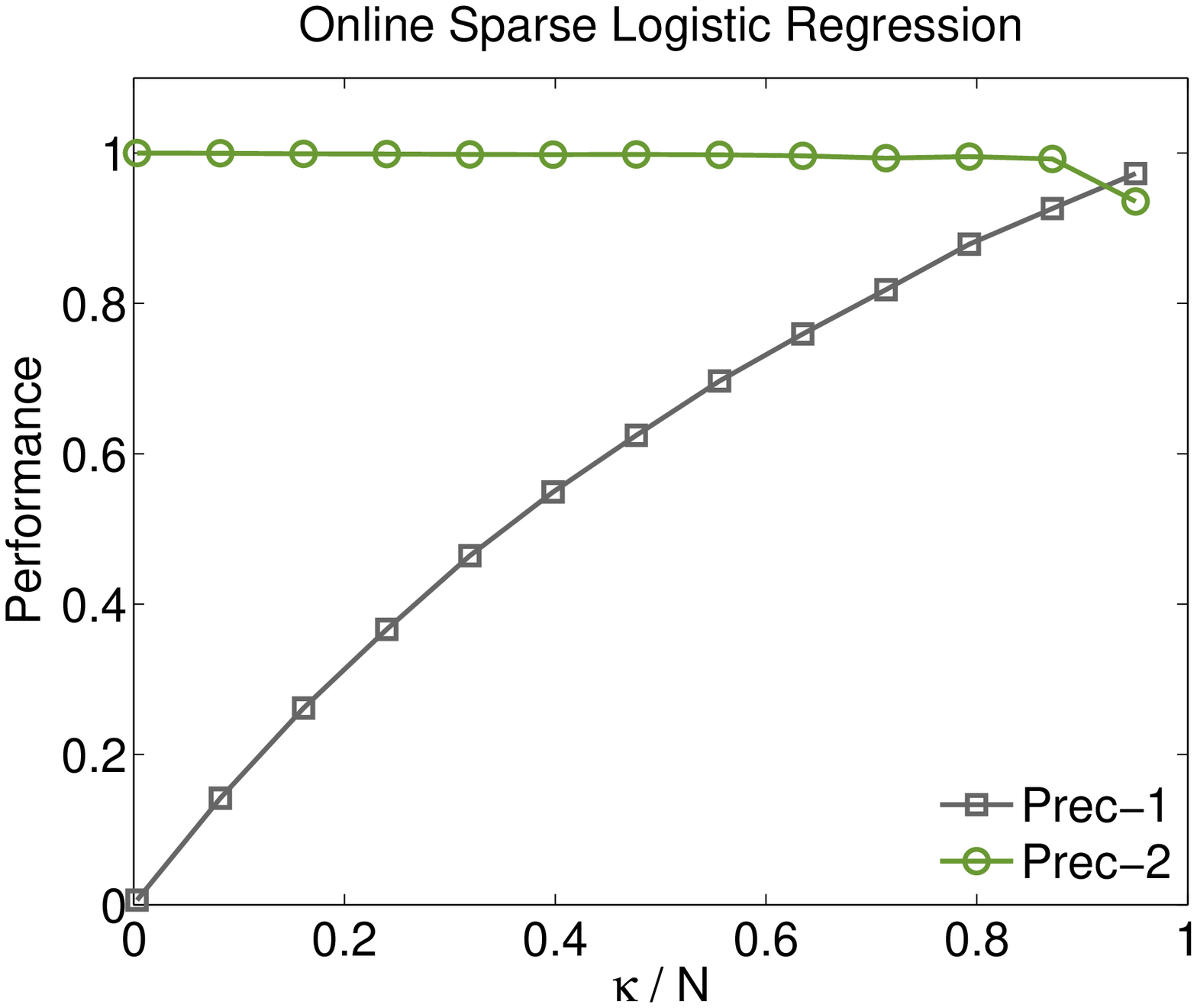}}
\subfloat[Online SLR for the IEEE 118-bus.]{\includegraphics[width=1.75in, height=1.5in]{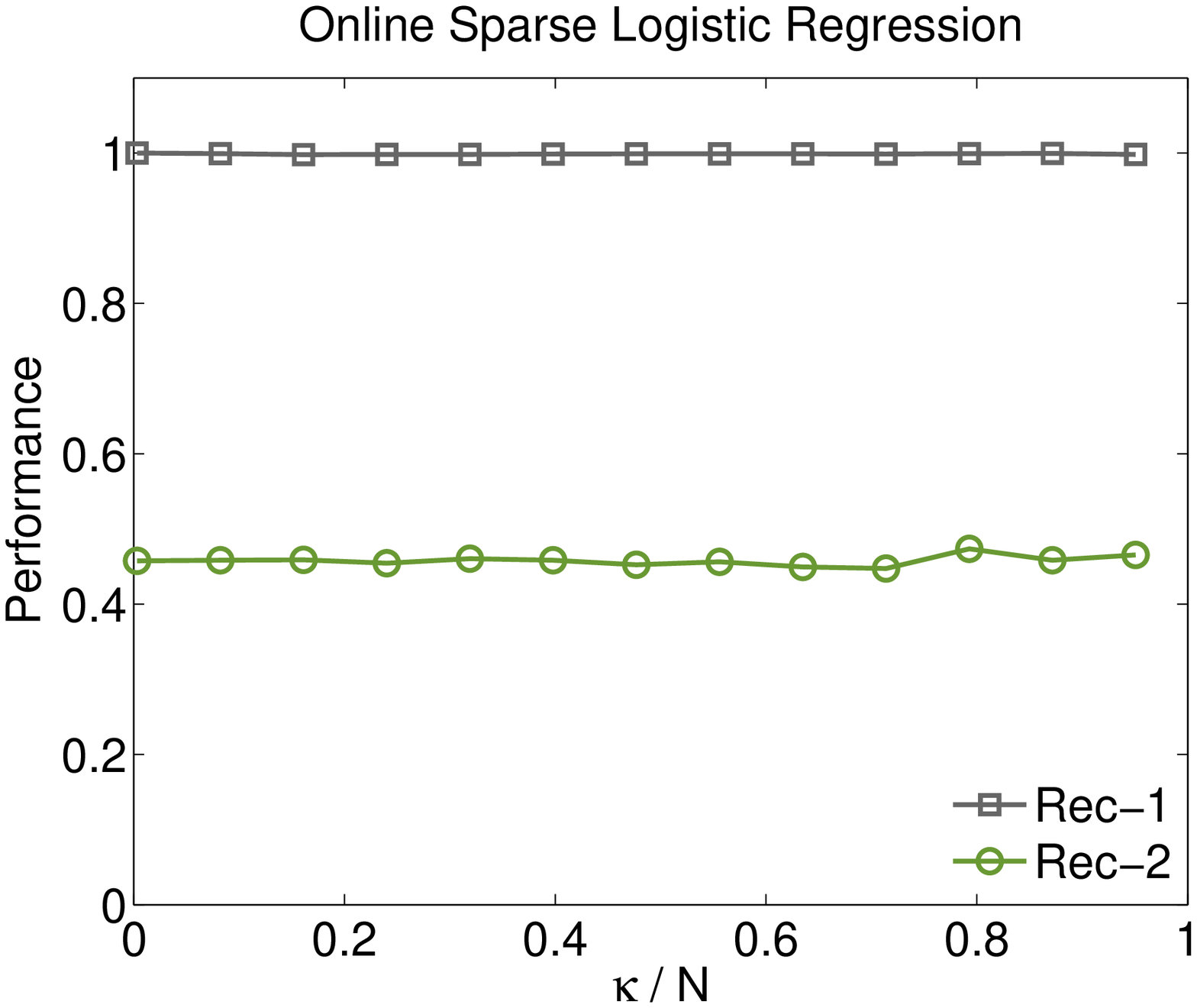}}
\caption{Experiments using the Online Perceptron (OP), Online Perceptron with Weighted Models (OPWM), Online SVM and SLR. Recall values of the OP are less than that of the OPWM for Class-1. Multiple phase transitions of performance values of the Online SVM are observed in the IEEE 118-bus system.}
\label{fig:910}
\end{figure*}

\subsection{Results for Online Learning Algorithms}

We consider four online learning algorithms, namely Online Perceptron (OP), Online Perceptron with Weighted Models (OPWM), Online SVM and Online SLR. Note that these algorithms are the online versions of the batch learning algorithms given in Section~\ref{sec:supervised} and developed considering the online algorithm design approach given in Section~\ref{sec:online}. The details of the implementations of the OP, OPWM, Online SVM and SLR are given in \cite{online,online_l1} and \cite{dogma}. 

When the OP is used, only the model $\mathbf{w}(t)$ computed using the last observed measurement at time $t$ is considered for the classification of the test samples. On the other hand, we consider an average of the models ${\mathbf{w}_{ave}(t)= \frac{1}{T} \sum \limits ^T _{t=1} \mathbf{w}(t)}$ which is computed by minimizing margin errors in the OPWM. Results are given for the OP in Fig.~\ref{fig:910}. In the weighted models, we observe phase transitions of the performance values for Class-2 in Fig.~\ref{fig:910}.e-Fig.~\ref{fig:910}.h. However, the phase transitions occur before the critical values, and the values of the phase transition points decrease as the system size increases. Additionally, we do not observe sharp phase transitions in the OP. 

In the OP, if the label of a measurement $\mathbf{s}$ is not correctly labeled, then the measurement vector is added to a set of supporting measurements $\mathbb{S}$ that are used to update the hypotheses in the training process. However, the hypotheses are updated in the OPWM if a measurement $\mathbf{s'}$ is not correctly labeled, and the vectors of $\mathbf{s'}$ and $ \mathbf{s} \in \mathbb{S}$ are linearly independent. Since the smallest number of linearly dependent measurements increases as $\frac{\kappa}{N}$ increases \cite{mi_jsac,donoho_2003}, the size of $\mathcal{S}$ decreases and the bias is decreased towards Class-1. Therefore, false negative ($fn$) values decrease and false positive ($fp$) values increase \cite{murphy2012machine}. As a result, we observe that Recall values of the OP are less than that of the OPWM for Class-1. The results of the Online SVM and Online SLR are provided in Fig.~\ref{fig:910} for different IEEE test systems. We observe phase transitions of performance values in the Online SVM similar to the batch supervised SVM.

\begin{figure}[h!]
\centering
\subfloat[Observable attacks.]{\includegraphics[width=1.7in, height=1.5in]{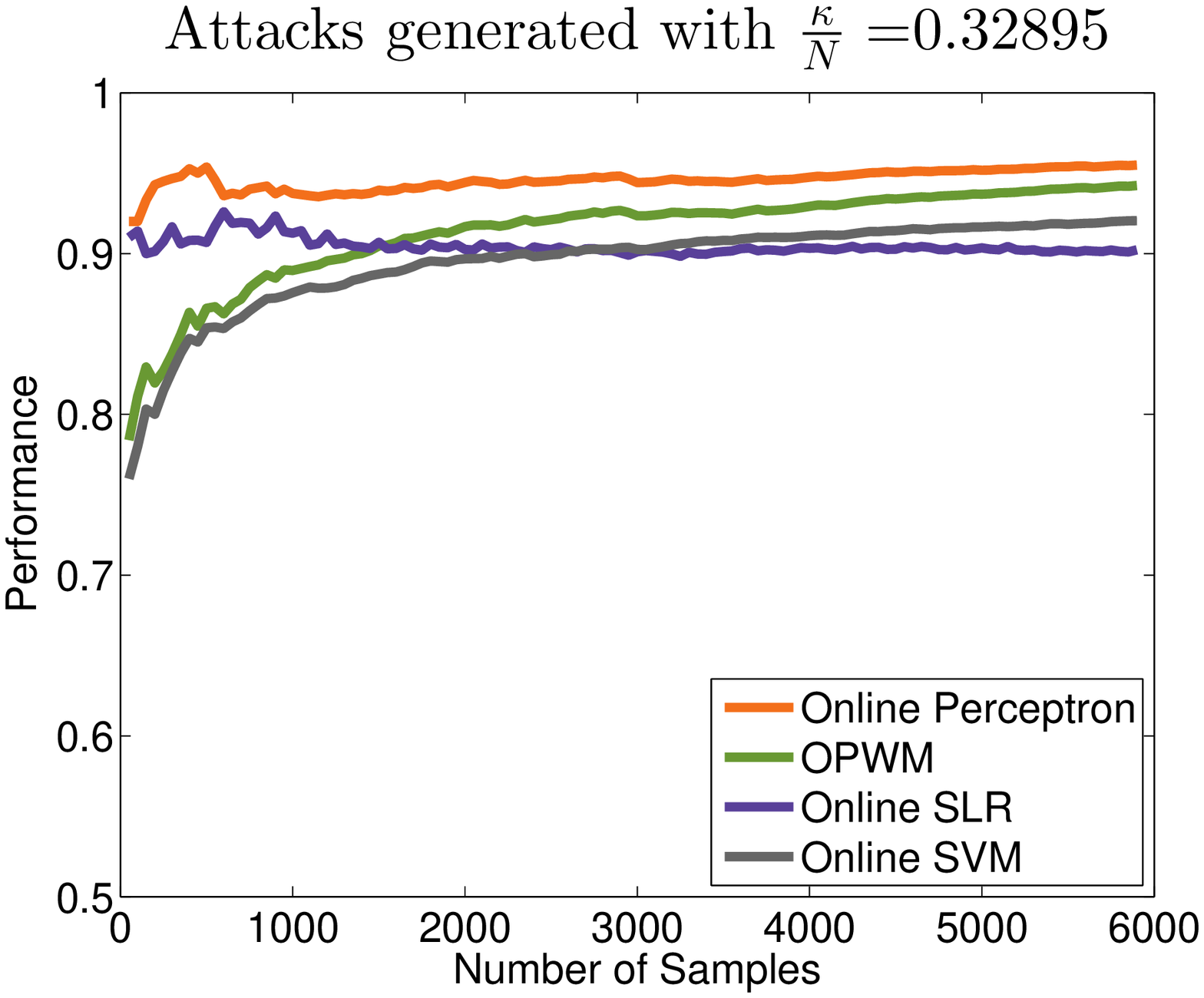}}
\subfloat[Unobservable attacks.]{\includegraphics[width=1.7in, height=1.5in]{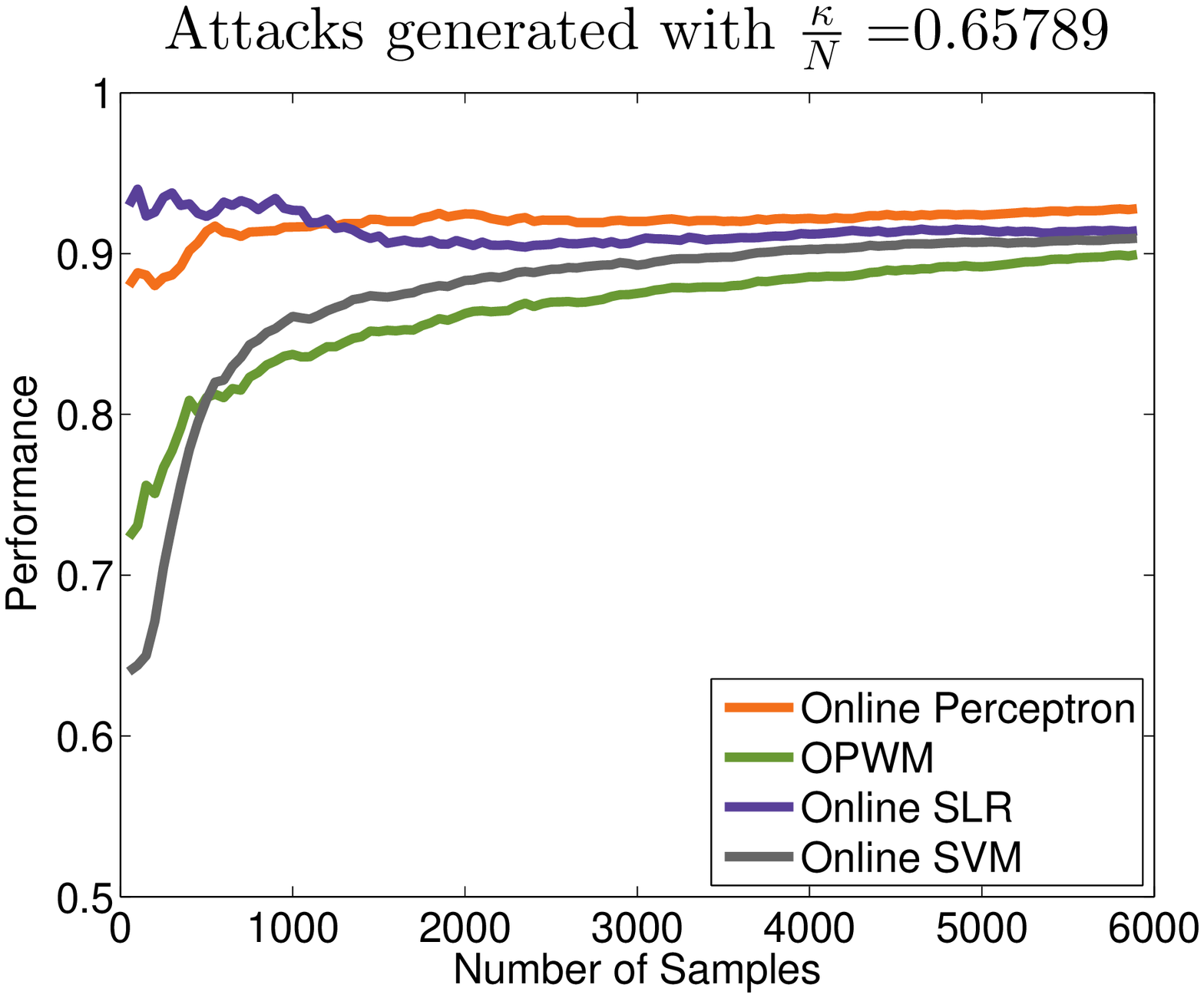}}
\caption{Learning curves of online learning algorithms.}
\label{fig:11}
\end{figure}

Learning curves of online learning algorithms are given in Fig.~\ref{fig:11} for both observable attacks generated with $\frac{\kappa}{N}=0.33$ and unobservable attacks generated with $\frac{\kappa}{N}=0.66$. Since the cost function of each online learning algorithm is different, the learning performance is measured and depicted using accuracy ($Acc$) defined in \eqref{eq:perf}. In the results, performance values of the Online SVM and OPWM increase as the number of samples increases, since the algorithms employ margin learning approaches which provide better learning rates as the number of training samples increases \cite{online,dogma}.

Briefly, we suggest using Online SLR for the scenarios in which the precision of the classification of secure variables is important to avoid false alarms. On the other hand, if the classification of attacked variables with high Precision and Recall values is an important task, we suggest using the Online Perceptron.   

\section{Summary and Conclusion}
\label{sec:conc}
The attack detection problem has been reformulated as a machine learning problem and the performance of supervised, semi-supervised, classifier and feature space fusion and online learning algorithms have been analyzed for different attack scenarios. 

In a supervised binary classification problem, the attacked and secure measurements are labeled in two separate classes. 
In the experiments, we have observed that state of the art machine learning algorithms perform better than the well-known attack detection algorithms which employ a state vector estimation approach for the detection of both observable and unobservable attacks. 

We have observed that the perceptron is less sensitive and the $k$-NN is more sensitive to the system size than the other algorithms. In addition, the imbalanced  data problem affects the performance of the $k$-NN. Therefore, $k$-NN may perform better in small sized systems and worse in large sized systems when compared to other algorithms. The SVM performs better than the other algorithms in large-scale systems. In the performance tests of the SVM, we observe a phase transition at $\kappa^*$, which is the minimum number of measurements that are required to be accessible by the attackers in order to construct unobservable attacks.  Moreover, a large value of $\kappa$ does not necessary imply high impact of data injection attacks. For example, if the attack vector $\mathbf{a}$ has small values in all elements, then the impact of $\mathbf{a}$ may still be limited. More important, if $\mathbf{a}$ is a vector with small values compared to the noise, then even machine learning-based approaches may fail. 

We observe two challenges of SVMs in their application to attack detection problems in smart grid. First, the  performance of the SVM is affected by the selection of kernel types. For instance, we observe that the linear and Gaussian kernel SVM perform similarly in the IEEE 9-bus system. However, for the IEEE 57-bus system the Gaussian kernel SVM outperforms its linear counterparts. Moreover, the values of the phase transition points of the performance of the Gaussian kernel SVM coincide with the theoretically computed $\kappa^*$ values. This implies that the feature vectors in $\mathcal{F}$, which are computed using Gaussian kernels, are linearly separable for higher values of $\kappa$. Interestingly, the transition points miss $\kappa^*$ in the IEEE 118-bus system, which means that alternative kernels are needed for this system. Second, the SVM is sensitive to the sparsity of the systems. In order to solve this problem, sparse SVM  \cite{sparse_svm} and kernel machines \cite{sparse_kernel} can be employed. In this paper, we approached this problem using the SLR. However, obtaining an \textit{optimal} regularization parameter, $\hat{\lambda}$, is computationally challenging \cite{admm}.

In order to use information extracted from test data in the computation of the learning models, semi-supervised methods have been employed in the proposed approach. In semi-supervised learning algorithms, we have used test data together with training data in an optimization algorithm used to compute the learning model. The numerical results show that the semi-supervised learning methods are more robust to the degree of sparsity of the data than the supervised learning methods. 

We have employed Adaboost and MKL as decision and feature level fusion algorithms. Experimental results show that fusion methods provide learning models that are more robust to changes in the system size and data sparsity than the other methods. On the other hand, computational complexities of most of the classifier and feature fusion methods are higher than that of the single classifier and feature extraction methods. 

Finally, we have analyzed online learning methods for real-time attack detection problems. Since a sequence of training samples or just a single sample is processed at each time, the computational complexity of most of the online algorithms is less than the batch learning algorithms. In the experiments, we have observed that classification performance of online learning algorithms are comparable to that of the batch algorithms.

In future work, we plan to first apply the proposed approach and the methods to an attack classification problem for deciding which of several possible attack types have occurred given that an attack have been detected. Then, we plan to consider the relationship between measurement noise and bias-variance properties of learning models for the development of attack detection and classification algorithms. Additionally, we plan to expand our analyses for varying number of clusters $G$ and cluster sizes $N_g$, $\forall g=1,2,\ldots,G$, by relaxing the assumptions made in this work for attack detection in smart grid systems, e.g. when the samples are not independent and identically distributed and obtained from non-stationary distributions, in other words, concept drift \cite{k_shift} and dataset shift \cite{datashift} occur. 

\bibliographystyle{IEEEtran}
\bibliography{sg4}

\end{document}